\documentclass[3p,times,authoryear]{article} 
\usepackage{iclr2021_conference,times}
\usepackage{graphicx}
\usepackage[tableposition=below]{caption}
\usepackage{longtable}
\usepackage{amsmath,amsfonts,bm}

\def\eqref#1{equation~\ref{#1}}

\def\1{\bm{1}}

\DeclareMathAlphabet{\mathsfit}{\encodingdefault}{\sfdefault}{m}{sl}
\SetMathAlphabet{\mathsfit}{bold}{\encodingdefault}{\sfdefault}{bx}{n}

\usepackage{enumitem}
\usepackage{hyperref}
\hypersetup{colorlinks,allcolors=black}

\usepackage{url}
\usepackage{tabularx}
\usepackage{adjustbox}
\usepackage{array}
\usepackage{float}
\usepackage{natbib}
\setcitestyle{round,authoryear,comma}
\usepackage{geometry}

\usepackage{array,ragged2e}
\newcolumntype{M}[1]{>{\Centering\arraybackslash}m{#1}}

\makeatletter
\renewcommand\paragraph{\@startsection{paragraph}{4}{0ex}%
   {1mm}%
   {1mm}%
   {\normalfont\normalsize\bfseries}}
\newcommand\mymaketitle{%
  \begin{titlepage}
    \null\vfil\vskip 40\p@
    \begin{center}
      {\@title}
      \vskip 2.5em
      {\large \lineskip .75em \@author \par}
      \vskip 1.5em
      {\large \@date \par}
    \end{center}\par
    \@thanks
    \vfil\null
  \end{titlepage}
}
\makeatother

\usepackage{fancyhdr}

\title{A Comprehensive Survey on Applications of \\ Transformers for Deep Learning Tasks}

\author{Saidul Islam$^1$, Hanae Elmekki$^1$, Ahmed Elsebai$^1$, Jamal Bentahar$^{1,2,*}$, Najat Drawel $^1$, Gaith Rjoub$^{3,1}$, Witold Pedrycz$^{4,5,6,7}$ \\
$^1$Concordia Institute for Information Systems Engineering, 
Concordia University, Montreal, Canada \\
$^2$Department of Electrical Engineering and Computer Science,
Khalifa University, Abu Dhabi, UAE\\
$^3$King Hussein School of Computing Sciences, Princess Sumaya University for Technology, Jordan\\
$^4$Department of Electrical and Computer Engineering, University
of Alberta, Edmonton, Canada\\
$^5$Systems Research Institute, Polish Academy of Sciences, Warsaw, Poland\\
$^6$Department of Computer Engineering, Istinye University, Sariyer/Istanbul,
Turkiye\\
$^7$Department of Electrical and Computer Engineering, King Abdulaziz University, Jeddah, Saudi Arabia\\\\
\textbf{$^*$Corresponding Author's Email:}  jamal.bentahar@concordia.ca\\
\textbf{Contributing Authors' Emails:} saidul.islam@concordia.ca;
hanae.elmekki@mail.concordia.ca;\\ ahmed.elsebai@outlook.com;
n\_drawe@encs.concordia.ca; g.rjoub@psut.edu.jo; wpedrycz@ualberta.ca\\\\
The authors contributed equally to this work.
}

\usepackage{fancyhdr}
\renewcommand{\headrulewidth}{0pt}
\renewcommand{\footrulewidth}{0pt}
\fancypagestyle{first}{%
\fancyhf{} 

\lhead{This work has been submitted to the Expert Systems With Applications journal (Elsevier) for possible publication} 
\setlength{\headsep}{0.5in}
\renewcommand{\headrulewidth}{0pt}
\renewcommand{\footrulewidth}{0pt}}
\begin{document}

  \begin{center}
\maketitle
\thispagestyle{first}
\end{center}
\begin{abstract}

Transformer is a deep neural network  that employs a self-attention mechanism to comprehend the contextual relationships within sequential data. Unlike conventional neural networks or updated versions of Recurrent Neural Networks (RNNs) such as Long Short-Term Memory (LSTM), transformer models excel in handling long dependencies between input sequence elements and enable parallel processing. 
As a result, transformer-based models have attracted substantial interest among researchers in the field of artificial intelligence. This can be attributed to their immense potential and remarkable achievements, not only in Natural Language Processing (NLP) tasks but also in a wide range of domains, including computer vision, audio and speech processing, healthcare, and the Internet of Things (IoT).
Although  several  survey papers have been published highlighting the transformer's contributions in specific fields, architectural differences, or performance evaluations, there is still a significant absence of a comprehensive survey paper encompassing its major applications across various domains. Therefore, we undertook the task of filling this gap by conducting an extensive survey of proposed transformer models from 2017 to 2022. Our survey encompasses the identification of the top five application domains for transformer-based models, namely: NLP, Computer Vision, Multi-Modality, Audio and Speech Processing, and Signal Processing. 
We analyze the impact of highly influential transformer-based models in these domains and subsequently classify them based on their respective tasks using a proposed taxonomy. Our aim is to shed light on the existing potential and future possibilities of transformers for enthusiastic researchers, thus contributing to the broader understanding of this groundbreaking technology.

\textit{Keywords}: Self-attention; Transformer; Deep learning, Recurrent networks; Long short-term memory-LSTM; Multi-modality.

\end{abstract}

\section{Introduction}
Deep Neural Networks (DNNs) have emerged as the predominant infrastructure and state-of-the-art solution for the majority of learning-based machine intelligence tasks in the field of artificial intelligence. Although various types of DNNs are utilized for specific tasks, the multilayer perceptron (MLP) represents the classic form of neural network which is characterized by multiple linear layers and nonlinear activation functions \citep{MLP}. 
For instance, in computer vision, convolutional neural networks incorporate convolutional layers to process images, while recurrent neural networks employ recurrent cells to process sequential data, particularly in Natural Language Processing (NLP) \citep{CNN, RNN}. 
Despite the  wide use of recurrent neural networks, they exhibit certain limitations. One of the major issues with conventional networks is that they have short-term dependencies associated with exploding and vanishing gradients. In contrast, to achieve good results in NLP, long-term dependencies must be captured. Additionally, recurrent neural networks are slow to train due to their sequential data processing and computational approach \citep{RNN-limitation}. To address these issues, the long-short-term memory (LSTM) version of recurrent networks was developed, which improves the gradient descent problem of recurrent neural networks and increases the memory range of NLP tasks \citep{LSTM}. However, LSTMs still struggle with the problem of sequential processing, which hinders the extraction of the actual meaning of the context.  To tackle this challenge, bidirectional LSTMs were introduced, which process natural language from  both directions, i.e., left to right and right to left, and then concatenate the outcomes to obtain the context's actual meaning. Nevertheless, this technique still results in a slight loss of the true meaning of the context \citep{BiLSTM, BiLSTM-Limitation}.

Transformers are a type of deep neural network (DNNs) that offer a solution to the limitations of sequence-to-sequence (seq-2-seq) architectures, including short-term dependency of sequence inputs and the sequential processing of input, which hinders parallel training of networks. Transformers leverage the multi-head self-attention mechanism to extract features, and they exhibit great potential for application in NLP. Unlike traditional recurrence methods, transformers utilize attention to learn from an entire segment of a sequence, using encoding and decoding blocks. One key advantage of transformers over LSTM and recurrent neural networks is their ability to capture the true meaning of the context, owing to their attention mechanism. Moreover, transformers are faster since they can work in parallel, unlike recurrent networks, and can be calculated using Graphic Processing Units (GPUs), allowing for faster computation of tasks with large inputs \citep{attentionmechanism,r1,intro_ref3}. The advantages of the transformer model have inspired deep learning researchers to explore its potential for various tasks in different fields of application \citep{eswa/WangTrans23}, leading to numerous research papers and the development of transformer-based models for a range of tasks in the field of artificial intelligence \citep{intro_ref1,intro_ref2,intro_ref4}.

In the research community, the importance of survey papers in providing a productive analysis, comparison, and contribution of progressive topics is widely recognized. Numerous survey papers on the topic of transformers can be found in the literature. Most of them are addressing specific fields of application  \citep{DBLP:journals/csur/KhanNHZKS22, transformer_summarization, shamshad2023transformers}, compare the performance of different model\citep{DBLP:journals/csur/TayDBM23,practical_light_transformer, selva2023video}, or conduct architecture-based analysis \citep{DBLP:journals/aiopen/LinWLQ22}. Nevertheless, a well-defined structure that comprehensively focuses on the top application fields and systematically analyzes the contribution of transformer-based models in the execution of various deep learning tasks within those fields is still widely needed.

Indeed, conducting a survey on transformer applications would serve as a valuable reference source for enthusiastic deep-learning researchers seeking to gain a better understanding of the contributions of transformer models in diverse fields. Such a survey would enable the identification and discussion of potential models, their characteristics, and working methodology, thus promoting the refinement of existing transformer models and the discovery of novel transformer models or applications. To address the absence of such a survey, this paper presents a comprehensive analysis of all transformer-based models, and identifies the top five application fields, namely NLP, Computer Vision, Multi-Modality, Audio \& Speech, and Signal Processing), and proposes a taxonomy of transformer models, with significant models being classified and analyzed based on their task execution within these fields. Furthermore, the top-performing and significant models are analyzed within the application fields, and based on this analysis, we discuss the future prospects and challenges of transformer models.

\medskip

\subsection{Contributions and Motivations} 

Although several survey articles on the topic of transformers already exist in the literature, our motivations for conducting this survey stem from two essential observations. First, most of these studies have focused on transformer architecture,  model efficiency, and specific artificial intelligence fields, such as NLP, computer vision, multi-modality, audio \& speech, and signal processing.  They have often neglected other crucial aspects, such as the transformer-based model's execution in deep learning tasks across multiple application domains.  We aim in this survey to cover all major fields of application and present significant models for different task executions. The second motivation is the absence of a comprehensive and methodical analysis encompassing various prevalent application domains, and their corresponding utilization of transformer-based models, in relation to diverse deep learning tasks within distinct fields of application. We propose a high-level classification framework for transformer models, which is based on their most prominent fields of application. The prominent models are categorized and evaluated based on their task performance within the respective fields. 
In this survey, we highlight the application domains of transformers that have received comparatively greater or lesser attention from researchers. To the best of our knowledge, this is the first review paper that presents a high-level classification scheme for the transformer-based models and provides a collection of criteria that aim to achieve two objectives: (1) assessing the effectiveness of transformer models in various applications; and (2) assisting researchers interested in exploring and extending the capabilities of transformer-based models to new domains. Moreover, the paper provides valuable insights into potential future applications and highlights unresolved challenges within this field.

The remainder of the paper is organized as follows. Preliminary concepts important for the rest of the paper are explained in Section \ref{sec:pre}. A detailed description of the systematic methodology used to search for relevant research articles is provided in Section \ref{sec:met}. Section \ref{sec:related} presents related review papers and discusses similarities and differences with the current survey paper, which helps us identify the unique characteristics and the value added of our survey. Section \ref{sec:trans} identifies the the transformer models  proposed so far across different
fields of application. A Classification of the selected scientific articles on section \ref{application}. Section \ref{future_work} outlines potential directions for future work. Finally, Section \ref{conclusion} concludes the paper and summarizes the key findings and contributions of the study.

\section{Preliminaries}\label{sec:pre}
Before delving into the literature of transformers, let us describe some concepts that
will be used throughout this article.

\subsection{Transformer Architecture}


The transformer model was first proposed in 2017 for a machine translation task, and since then, numerous models have been developed based on the inspiration of the original transformer model to address a variety of tasks across different fields. While some models have utilized the vanilla transformer architecture as is, others have leveraged only the encoder or decoder module of the transformer model. As a result, the task and performance of transformer-based models can vary depending on the specific architecture employed. Nonetheless, a key and widely used component of transformer models is self-attention, which is essential to their functionality. All transformer-based models employ the self-attention mechanism and multi-head attention, which typically forms the primary learning layer of the architecture. Given the significance of self-attention, the role of the attention mechanism is crucial in transformer models \citep{r1}

\subsubsection{Attention Mechanism}

The attention mechanism has garnered significant recognition since its introduction in the 1990s, owing to its ability to concentrate on critical pieces of information. In image processing, certain regions of images were found to be more pertinent than others. Consequently, the attention mechanism was introduced as a novel approach in computer vision tasks, aiming to emphasize important parts based on their contextual relevance in the application. This technique yielded significant outcomes when implemented in computer vision, thereby promoting its widespread adoption in various other fields such as language processing.

\begin{figure}
    \begin{center}
    \includegraphics[width=.58\linewidth,height=.44\textwidth]{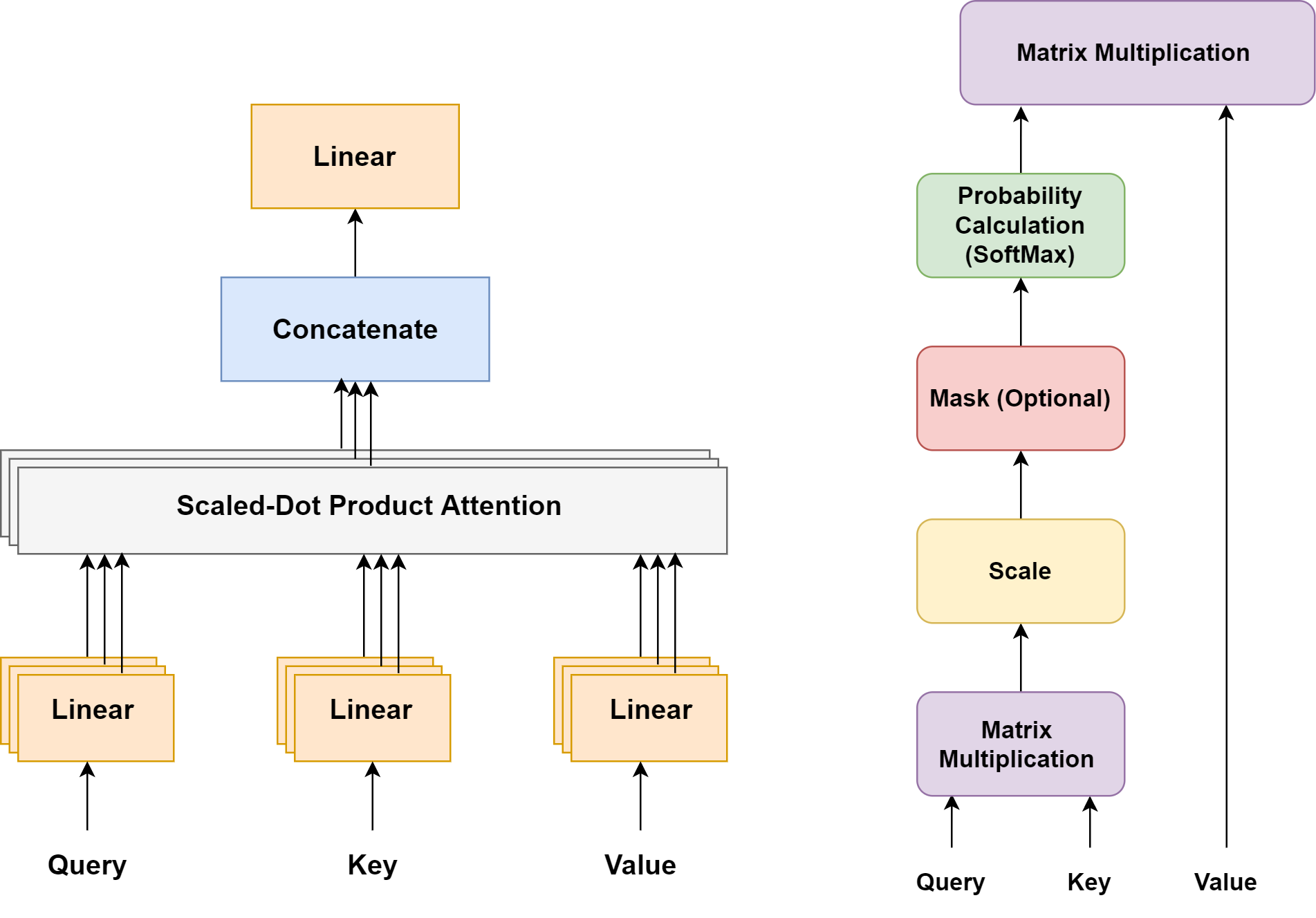}
    \caption{Multi-head attention \& scaled dot product attention \citep{r1}}
    \label{fig:attention}
    \end{center}
\end{figure}

In 2017, a novel attention-based neural network, named ``Transformer'', was introduced to address the limitations of other neural networks (such as A recurrent neural network (RNN)) in encoding long-range dependencies in sequences, particularly in language translation tasks \citep{r1}. The incorporation of a self-attention mechanism in the transformer model improved the performance of the attention mechanism by better capturing local features and reducing the reliance on external information. In the original transformer architecture, the attention technique is implemented through the ``Scaled Dot Product Attention", which is based on three primary parameter matrices: Query (Q), Key (K), and Value (V). Each of these matrices carries an encoded representation of each input in the sequence \citep{r1}. The SoftMax function is applied to obtain the final output of the attention process, which is a probability score computed from the combination of the weights of the three matrices (see Figure \ref{fig:attention}).
Mathematically, the scaled dot product attention function is computed as follows:
\[
Attention(Q,K,V) = softmax\left( \frac{QK^T}{\sqrt{dk}}\right)V
\]
The matrices $Q$ and $K$ represent the Query and Key vectors respectively, both having a dimension of $dk$, while the matrix $V$ represents the values vectors.

\subsubsection {Multi-head attention}

The application of the scaled dot-product attention function in parallel within the multi-head Attention module is essential for extracting the maximum dependencies among different segments in the input sequence. Each head denoted by $k$ performs the attention mechanism based on its own learnable weights $W^{kQ}$, $W^{kK}$, and $W^{kv}$. The attention outputs calculated by each head are subsequently concatenated and linearly transformed into a single matrix with the expected dimension \citep{r1}.

\begin{gather*} 
    headk = Attention(QW^{kQ},KW^{kK},VW^{kV})
\\  MultiHead(Q,K,V) = Concat(head1, head2, .... headH)W^0
\end{gather*} 


The utilization of multi-head attention facilitates the neural network in learning and capturing diverse characteristics of the input sequential data. Consequently, this enhances the representation of the input contexts, as it merges information from distinct features of the attention mechanism within a specific range, which could be either short or long. This approach allows the attention mechanism to jointly function, which results in better network performance  \citep{r1}.

\subsection{Architecture of the Transformer Model}

\begin{figure}[t]
    \begin{center}
    \includegraphics[width=.53\linewidth,height=.64\textwidth]{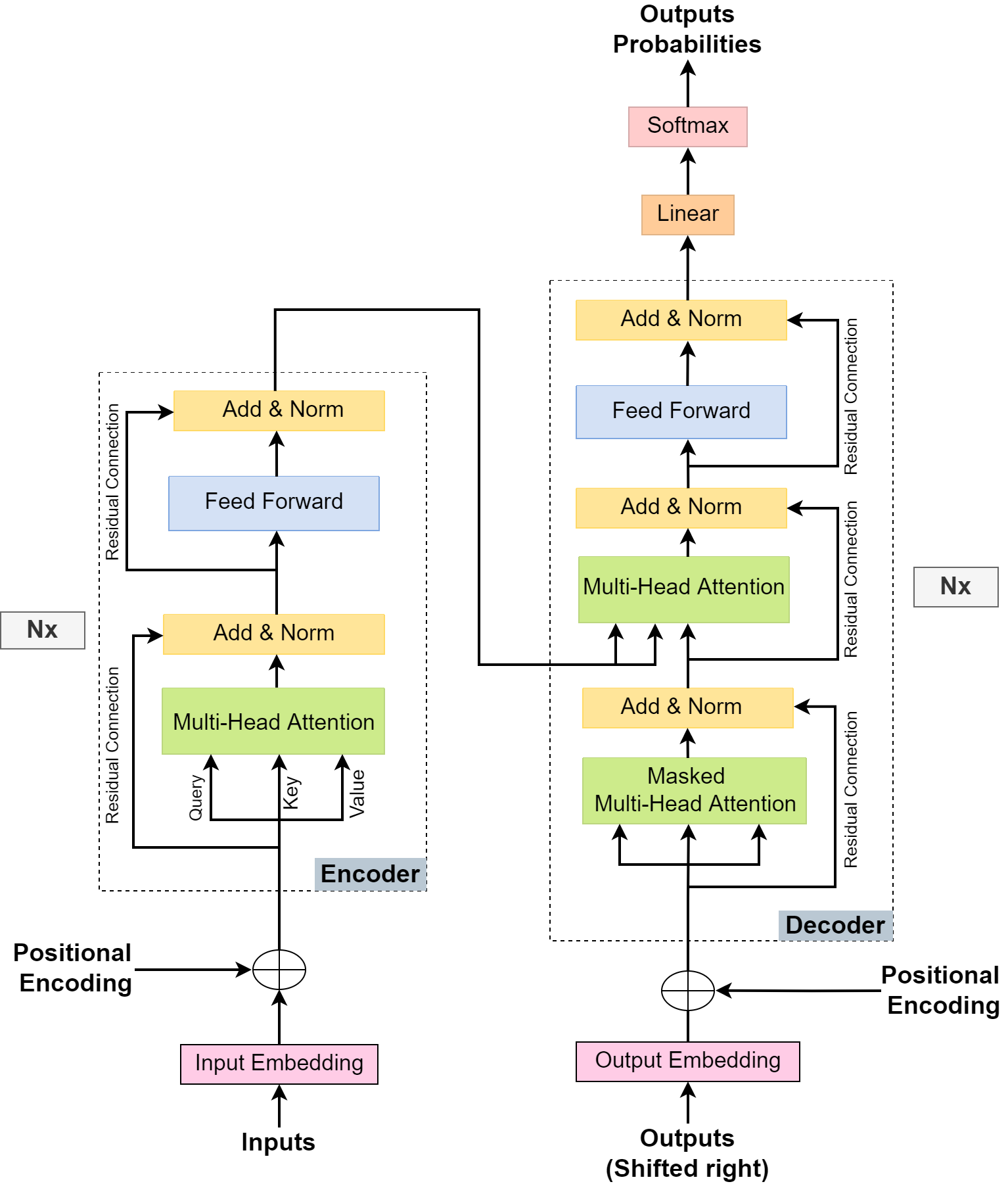}
    \caption{Transformer architecture  \citep{r1}}
    \label{fig:my_label}
    \vspace{-5pt} 
    \end{center}
\end{figure}

The transformer model was primarily developed based on the attention mechanism \citep{r1}, with the aim of processing sequential data. Its outstanding performance, especially in achieving state-of-the-art benchmarks for NLP translation models, has led to the widespread use of transformers. As depicted in Figure \ref{fig:my_label}, the overall architecture of the transformer model for sentence translation tasks involves the use of attention mechanisms. However, for different applications, the transformer architecture may be subject to variation, depending on specific requirements.

The initial transformer architecture was developed based on the auto-regressive sequence transduction model, comprising two primary modules, namely Encoder and Decoder. These modules are executed multiple times, as required by the task at hand. Each module comprises several layers that integrate the attention mechanism. Particularly, the attention mechanism is executed in parallel multiple times within the transformer architecture, which explains the presence of multiple ``Attention Heads" \citep{r1}.

\subsubsection {Encoder module} 

The stacked module within the transformer architecture comprises two fundamental layers, namely the Feed-Forward Layer and Multi-Head Attention Layer. In addition, it incorporates Residual connections around both layers, as well as two Add and Norm layers, which play a pivotal role \citep{r1}. In the case of text translation, the Encoder module receives an embedding input that is generated based on the input's meaning and position information via the Embedding and Position Encoding layers. From the embedding input, three parameter matrices are created, namely the Query ($Q$), Key ($K$), and Value ($V$) matrices, along with positional information, which are passed through the ``Multi-Head Attention" layer. Following this step, the Feed-Forward layer addresses the issue of rank collapse that can arise during the computation process. Additionally, a normalization layer is applied to each step, which reduces the dependencies between layers by normalizing the weights used in gradient computation within each layer. To address the issue of vanishing gradients, the Residual Connection is applied to every output of both the attention and feed-forward layers, as illustrated in Figure \ref{fig:my_label}.

\subsubsection {Decoder module} 

The Decoder module in the transformer architecture is similar to the Encoder module, with the inclusion of additional layers such as Masked Multi-Head Attention. In addition to the Feed-Forward, Multi-Head Attention, Residual connection, and Add and Norm layers, the Decoder also contains Masked Multi-Head Attention layers. These layers use the scaled dot product and Mask Operations to exclude future predictions and consider only previous outputs. The Attention mechanism is applied twice in the Decoder: one for computing attention between elements of the targeted output and another for finding attention between the encoding inputs and targeted output. Each attention vector is then passed through the feed-forward unit to make the output more comprehensible to the layers. The generated decoding result is then caught by Linear and SoftMax layers at the top of the Decoder to compute the final output of the transformer architecture. This process is repeated multiple times until the last token of a sentence is found \citep{r1}, as illustrated in Figure \ref{fig:my_label}.

\section{Research Methodology\label{sec:met}} 

\begin{figure}[htbp]
  \centering
  \begin{adjustbox}{width=1.05\textwidth,center}
    \includegraphics{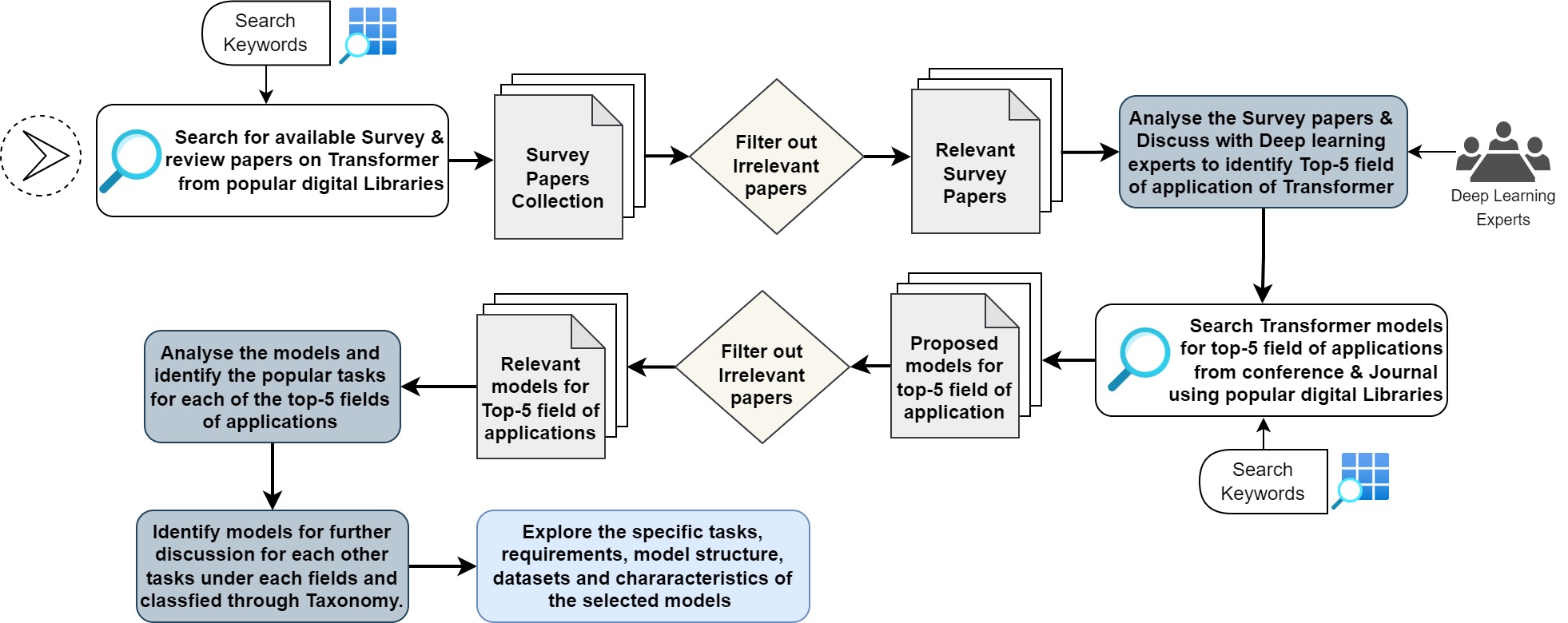}
  \end{adjustbox}
  \caption{Methodology of the survey}
  \label{fig:methodology}
\end{figure}

In this survey, we collect and analyze the most recent surveys on transformers  that have been published in refereed journals and conferences with the aim of studying their contributions and limitations. To gather the relevant papers, we employed a two-fold strategy:
 (1) searching using several established search engines and selected papers based on the keywords ``survey", ``review",  ``Transformer", ``attention", ``self-attention", ``artificial intelligence", and ``deep learning; and (2) evaluating the selected papers and eliminated those that were deemed irrelevant for our study.  A detailed organization of our survey is depicted
in Figure \ref{fig:methodology}.

Indeed, by means of a comprehensive examination of survey papers and expert discussions on deep learning, we have identified  the top five  domains of application for transformer-based models, these are: (i) NLP, (ii) computer vision, (iii) multi-modality, (iv) audio/speech, and (v) signal processing. Subsequently, we performed a systematic search for journal and conference papers that presented transformer-based models in each of the aforementioned fields of application, utilizing the keywords presented in Table \ref{tab: methodology_table}. Our search yielded a substantial number of papers for each field, which we thoroughly reviewed and evaluated. We selected papers that proposed novel transformer-based or transformer-inspired models for deep learning tasks, while disregarding others. Through our examination of this extensive collection of models, we have identified prevalent deep-learning tasks associated with each field of application.

As we have examined more than 600 transformer models during this process, it has become exceedingly difficult to classify such a large number of models and conduct thorough analyses of each task within every field of application. Therefore, we have opted to perform a more comprehensive analysis of a number of transformer models for each task within every field of application. These models were selected based on specific criteria and an in-depth analysis was carried out accordingly. The selected models are as follows:

\begin{enumerate}
  \item   The transformer-based models that have been proposed to execute a deep learning task for the first time and opened up new path for research in the field of transformer applications.
  \item The models that have proposed alternative or novel approaches to implementing the transformer's attention mechanism, as compared to the vanilla architecture, such as introducing a new attention mechanism or enhancing the position encoding module.
  \item The transformer models have had a significant impact in the field, with higher citation rates, and have been widely accepted by the scientific community. Models that have also contributed to breakthroughs in the advancement of transformer applications.
  \item The models and their variants have been proposed for the purpose of applying the transformer technology to real-world applications, with the aim of achieving superior performance results in comparison to other deep learning methods.
  \item The transformer-based models generated a significant buzz within the theoretical and applied artificial intelligence  community.
\end{enumerate}



\medskip
\setlength\LTleft{-.90cm}
\setlength\LTright{\LTleft}
\begin{longtable}{| M{2.5cm} | M{4.5cm}| M{2cm} |M{2.8cm}|M{2cm}|M{2cm}|}

  \hline
\textbf{Fields of Application} & \textbf{Keywords for Paper Search} & \multicolumn{2}{c|}{\textbf{Tasks Of Application}} & \multicolumn{2}{c|}{\textbf{Number of papers}} \\ \cline{5-6} 
 &  & \multicolumn{2}{c|}{} & \multicolumn{1}{l|}{\textbf{\begin{tabular}[c]{@{}l@{}}Relevant models\\ using keywords\end{tabular}}} & \multicolumn{1}{l|}{\textbf{\begin{tabular}[c]{@{}l@{}}Selected models \\ for Taxonomy\end{tabular}}} \\ \hline
 \endfirsthead
 
 \multicolumn{6}{c}%
{{\bfseries \tablename\ \thetable{} -- continued from previous page}} \\
 \hline
\textbf{Fields of Application} & \textbf{Keywords for Paper Search} & \multicolumn{2}{c|}{\textbf{Tasks Of Application}} & \multicolumn{2}{c|}{\textbf{Number of papers}} \\ \cline{5-6} 
 &  & \multicolumn{2}{c|}{} & \multicolumn{1}{l|}{\textbf{\begin{tabular}[c]{@{}l@{}}Relevant models\\ using keywords\end{tabular}}} & \multicolumn{1}{l|}{\textbf{\begin{tabular}[c]{@{}l@{}}Selected models \\ for Taxonomy\end{tabular}}} \\ \hline
\endhead

\hline \multicolumn{6}{|r|}{{Continued on next page}} \\ \hline
\endfoot

\endlastfoot

\begin{tabular}[c]{@{}l@{}}Natural Language \\ Processing\end{tabular} & \begin{tabular}[c]{@{}l@{}}“Natural Language Processing”,\\ “NLP”,“Text”,“Text Processing”,\\ “Transformer”, “Attention”,\\ “Self-attention”, “multi-head\\ attention”, “Language model”.\end{tabular} & \multicolumn{2}{c|}{Language Translation} & 257 & 25 \\ \cline{3-4}
 &  & \multicolumn{2}{c|}{\begin{tabular}[c]{@{}c@{}}Text Classification \& Segmentation\end{tabular}} & \multicolumn{1}{l|}{} & \multicolumn{1}{l|}{} \\ \cline{3-4}
 &  & \multicolumn{2}{c|}{Question Answering} & \multicolumn{1}{l|}{} & \multicolumn{1}{l|}{} \\ \cline{3-4}
 &  & \multicolumn{2}{c|}{Text Summarization} & \multicolumn{1}{l|}{} & \multicolumn{1}{l|}{} \\ \cline{3-4}
 &  & \multicolumn{2}{c|}{Text Generation} & \multicolumn{1}{l|}{} & \multicolumn{1}{l|}{} \\ \cline{3-4}
 &  & \multicolumn{2}{c|}{\begin{tabular}[c]{@{}c@{}}Natural Language Reasoning\end{tabular}} & \multicolumn{1}{l|}{} & \multicolumn{1}{l|}{} \\ \cline{3-4}
 &  & \multicolumn{2}{c|}{\begin{tabular}[c]{@{}c@{}}Automated Symbolic \\ Reasoning\end{tabular}} & \multicolumn{1}{l|}{} & \multicolumn{1}{l|}{} \\ 
 \hline
Computer Vision & \begin{tabular}[c]{@{}l@{}}“Transformer”,“Attention”,\\ “Self-attention”,“Image”,\\ “Natural image”,“medical\\ image”,“Biomedical”,\\ “health”,“Image processing”,\\ “Computer vision”,“Vision”.\end{tabular} & \begin{tabular}[c]{@{}c@{}}Natural Image \\ Processing\end{tabular} & \begin{tabular}[c]{@{}c@{}}Image \\ Classification\end{tabular} & 197 & 27 \\ \cline{4-4}
 &  &  & \begin{tabular}[c]{@{}c@{}}Recognition \& \\ Object Detection\end{tabular} &  &  \\ \cline{4-4}
 &  &  & \begin{tabular}[c]{@{}c@{}}Image \\ Segmentation\end{tabular} &  &  \\ \cline{4-4}
 &  &  & \multicolumn{1}{l|}{Image Generation} &  &  \\ \cline{3-4}
 &  & \begin{tabular}[c]{@{}c@{}}Medical Image \\ Processing\end{tabular} & \begin{tabular}[c]{@{}c@{}}Image \\ Segmentation\end{tabular} &  &  \\ \cline{4-4}
 &  &  & \begin{tabular}[c]{@{}c@{}}Image \\ Classification\end{tabular} &  &  \\ \cline{4-4}
 &  &  & \begin{tabular}[c]{@{}c@{}}Image \\ Translation\end{tabular} &  &  \\ 
\hline
Multi-modal & \begin{tabular}[c]{@{}l@{}}“Transformer”,“Attention”,\\ “Self-attention”,“multi-head\\  attention”,“multimodal”,\\ “multi-modality”,“text-image”,\\ “image-text”,“ video-audio-\\ text, “text-audio”,“audio-text”,\\ “vision-language”, \\ “language-vision”.\end{tabular} & \multicolumn{2}{c|}{\begin{tabular}[c]{@{}c@{}} Classification \& \\ Segmentation\end{tabular}} & 94 & 20 \\ \cline{3-4}
 &  & \multicolumn{2}{c|}{Visual Question Answering} &  &  \\ \cline{3-4}
 &  & \multicolumn{2}{c|}{Visual Captioning} &  &  \\ \cline{3-4}
 &  & \multicolumn{2}{c|}{\begin{tabular}[c]{@{}c@{}}Visual Common-sense\\  Reasoning\end{tabular}} &  &  \\ \cline{3-4}
 &  & \multicolumn{2}{c|}{\begin{tabular}[c]{@{}c@{}}Text/Image/Video/Speech \\ Generation\end{tabular}} &  &  \\ \cline{3-4}
 &  & \multicolumn{2}{c|}{Cloud Task Computing} &  &  \\ 
\hline
Audio \& Speech & \begin{tabular}[c]{@{}l@{}}“Transformer”,“Attention”,\\  “Self-attention”,“multi-head\\  attention”,“audio”,“Speech”,\\  “audio processing”,“speech\\ processing”,\end{tabular} & \multicolumn{2}{l|}{Audio \& Speech Recognition} & 70 & 16 \\ \cline{3-4}
 &  & \multicolumn{2}{l|}{Audio \& Speech Separation} &  &  \\ \cline{3-4}
 &  & \multicolumn{2}{l|}{Audio \& Speech Classification} &  &  \\ 
\hline
Signal Processing & \begin{tabular}[c]{@{}l@{}}“Transformer”, “Attention”, \\ “Self-attention”, “multi-head \\ attention”, “signal”, “signal \\ processing” , “wireless”, \\ “wireless signal”, “wireless \\ network”, “biosignal”, “medical \\ signal”.\end{tabular} & \multicolumn{2}{c|}{\begin{tabular}[c]{@{}c@{}}Wireless network Signal \\ processing\end{tabular}} & 23 & 11 \\ \cline{3-4}
 &  & \multicolumn{2}{c|}{Medical Signal Processing} &  &  \\ 
 \hline
\caption{\label{tab: methodology_table}Transformer models' field of application, used keywords for paper search, popular deep learning tasks, number of relevant papers by search, and number of selected models for taxonomy and further discussion.}
\end{longtable}

In the field of application, we have classified the selected models based on their task execution and developed a taxonomy of transformer applications. Our analysis involved a comprehensive examination of the models, including their structures, characteristics, operational methods, and datasets, among others. Based on this investigation, we provide an in-depth discussion of the potential future applications of transformers. To conduct this research, we consulted various prominent research repositories, such as “AAAI”, ``ACM", “ACL”, “CVPR”, “ICML”, “ICLR”, “ICCV”, “NeurlIPS”, “LREC”, ``IEEE", “PMLR”, "National Library of medicine",``SCOPUS", ``MDPI", ``ScienceDirect”, and ``Cornell University-arxiv library". Table \ref{tab: methodology_table}  depicts the category of the selected models.

\section{Related Work \label{sec:related}}
Transformers have been subjected to a number of surveys in recent years due to their effectiveness and a broad range of applications. We recorded more than $50$ survey papers about transformers in various digital libraries and examined them. After carefully considering these surveys, we selected $17$ significant survey papers for further in-depth analysis of their works. During this process, we considered the surveys published in reputed conferences and journals with high number of citations, whereas we discarded the papers that have not been published yet. We extensively analyzed this set of $17$ papers, delving into their content and investigating their respective fields of work and applications. We prioritized examining the resemblances and disparities between the existing surveys and our own paper. Our investigation revealed that numerous surveys primarily concentrated on the architecture and efficiency of transformers, while others solely focused on applications in NLP and computer vision. However, only a few explored the utilization of transformers in multi-modal combinations involving text and image data. These findings, along with supporting details, are presented in Table \ref{tab:related works}.
\par
Several review papers centered their attention on conducting architecture and performance-based analyses of transformers. Among them, the survey paper titled ``A Survey of Transformers" stands out as it offers a comprehensive examination of different X-formers and introduces a taxonomy based on architecture, pre-training, and application \citep{DBLP:journals/aiopen/LinWLQ22}. Another survey paper on transformers  was entitled “Efficient Transformers: A survey” to compare the computational power and memory efficiency of X-formers \citep{DBLP:journals/csur/TayDBM23}. Moreover, another paper focused on light and fast transformers while it explored different efficient alternatives to the standard transformers \citep{practical_light_transformer}.

Within the field of NLP, there exists a survey paper titled ``Visualizing Transformers for NLP: A Brief Survey" \citep{DBLP:conf/iv/BrasoveanuA20}. This particular study centers its attention on exploring the different aspects of transformers that can be effectively examined and understood through the application of visual analytics techniques. On a related note, another survey paper delves into the realm of pre-trained transformer-based models for NLP \citep{DBLP:journals/corr/abs-2108-05542}. This study extensively discusses pretraining methods and tasks employed in these models. Moreover, it introduces a taxonomy that effectively categorizes the wide range of transformer-based Pre-Trained Language Models (T-PTLMs) available in the literature.

Moreover, the paper entitled  ``Survey on Automatic Text Summarization and Transformer Models Applicability" focused on using transformers for text summarization tasks and proposed a transformer model that solves the issue of the long sequence input   \citep{transformer_summarization}. On the other hand, another survey worked on applying a bidirectional transformer encoder (BERT) in multi-layer as a word-embedding tool   \citep{bert_survey_word_embedding}. Furthermore, the application of transformers to detect different levels of emotions from text-based data has been explored in \citep{transformer_emotion_detection} under the tile ``Transformer models for text-based emotion detection: a review of BERT-based approaches". Another paper explored the use of the transformer language model in different information systems \citep{deep_transfer_survey}. It focused on using transformers as text miners to extract useful data from large organizations' data.
\par
Due to huge improvements in image processing tasks and amazing applications on computer vision with the help of transformer models in recent years, these models gained popularity among computer vision researchers. For instance, “Transformers in Vision: A Survey” provided a comprehensive overview of the existing transformer models in the field of computer vision and classified the models based on popular recognition tasks \citep{DBLP:journals/csur/KhanNHZKS22}.  A meticulous survey was undertaken to comprehensively analyze the merits and drawbacks of the leading ``Vision Transformers". This study placed significant emphasis on scrutinizing the training and testing datasets associated with these top-performing models, offering valuable insights into their performance and suitability for various applications \citep{DBLP:journals/pami/00020C0GLTXXXYZ23}.

Another survey paper compared transformer models developed for image and video data based on their performance in classification tasks \citep{selva2023video}.  Recent advances in computer vision and multi-modality have been  emphasized in another survey paper \citep{xu2022transformers} comparing the performance of different transformer models and providing some information regarding their pre-training.  Furthermore, an existing survey describes in detail several transformer models that have been developed for medical images; however, it does not provide information regarding medical signals \citep{li2023transforming}. Another paper gives an overview of transformer models developed in the medical field; however, it only concerns medical images, not medical signals \citep{shamshad2023transformers}.
\par
Multi-modality is getting very popular in deep learning tasks that helped to decipher several surveys on transformer focusing multi-modal domain. A paper worked on categorizing transformer vision-language models based on tasks and summarizing their co-responding advantages and disadvantages. Moreover, this survey paper covered video-language pre-trained models and categorized the models into single-stream and multi-stream structures, and the performance of the models is also compared in this survey  \citep{DBLP:journals/aiopen/RuanJ22}. Another survey, ``Perspectives and Prospects on Transformer Architecture for Cross-Modal Tasks with Language and Vision", explored transformers in multi-modal  visual-linguistic tasks \citep{pp_cross}. Other than NLP, computer vision and multi-modality, transformers are getting significant attention from researchers to apply to other fields such as time series and reasoning tasks. 

\medskip
\setlength\LTleft{-.90cm}
\setlength\LTright{\LTleft}
\begin{longtable}{| M{1.8cm} | M{1.8cm}| M{4.8cm} |M{9.3cm} |}

 \hline
\textbf{Approach} & \textbf{Fields of Application} & \textbf{Similarities} & \textbf{Differences} \\
 \hline
 \endfirsthead
 
 \multicolumn{4}{c}%
{{\bfseries \tablename\ \thetable{} -- continued from previous page}} \\
 \hline
\textbf{Approach} & \textbf{Fields of Application} & \textbf{Similarities} & \textbf{Differences} \\
 \hline
\endhead

\hline \multicolumn{4}{|r|}{{Continued on next page}} \\ \hline
\endfoot

\endlastfoot

Q Fournier et al. \citep{practical_light_transformer}  & Performance /Architecture & 

\begin{itemize}[leftmargin=0.3cm]
\item A classification of the transformers is suggested, and this 
classification is based on attention mechanism modification or architecture modification \end{itemize} & 

\begin{itemize}[leftmargin=0.3cm]
\item This pare surveyed the different alternatives of the standard transformers that are more efficient in terms of time and memory complexities, and these alternatives are categorized by either modifying the attention mechanism or the network architecture. Their classification is based on the change in architecture or change in attention mechanism, while our classification is driven by application areas.
\end{itemize}\\
 \hline
 
T. Lin et al. \citep{DBLP:journals/aiopen/LinWLQ22}  & Performance /Architecture & 
\begin{itemize}[leftmargin=0.3cm]
\item Proposed taxonomy of X-formers covering several fields 
\end{itemize}
& 

\begin{itemize}[leftmargin=0.3cm]
 \item This existing survey compared X-formers from architectural modification, pre-training, and a very small range of application perspectives, while our survey deeply focuses on popular tasks under each field of application.

\item The wireless/medical signal processing and cloud computing tasks application were missing in this exciting survey, while our survey covers these tasks and applications.
\end{itemize}\\
 \hline

 Y. Tay et al. \citep{DBLP:journals/csur/TayDBM23} & Performance /Architecture & 
\begin{itemize}[leftmargin=0.3cm]
\item Proposed a taxonomy considering the primary use case of transformer models in language and vision domains .
\end{itemize}
& 
\begin{itemize}[leftmargin=0.3cm]
\item This existing survey compared the computational power and memory efficiency of transformer models, whereas our survey focuses  on deep learning tasks and applications.

\item  This exciting survey focused on language and vision domain only, while we cover other top five fields of transformer applications: NLP, computer vision, multi-modality, audio/speech, and signal processing.
\end{itemize}\\
 \hline

A. M. P. Braşoveanu et al. \citep{DBLP:conf/iv/BrasoveanuA20} & Natural language Processing-NLP & 
 \begin{itemize}[leftmargin=0.3cm]
 \item Explain transformer architecture and explain its features.
 \end{itemize}
 & 
 \begin{itemize}[leftmargin=0.3cm]
 \item Our survey describes the transformer model and the significant models' working processing for a range of tasks. However, this existing paper focused on visualization techniques used to explain the most recent transformer architectures and explored two large tool classes to explain the inner workings of Transformers.
\item we covered five fields of transformer applications: NLP, computer vision, multi-modality, audio/speech, and signal processing and this exciting survey focused on the models for NLP only.
\end{itemize}\\
 \hline

W Guan et al. \citep{transformer_summarization}& Natural language Processing-NLP & 

\begin{itemize}[leftmargin=0.3cm]
\item Survey an application area of transformers, which is text summarization, which is one of the application areas covered in our survey  
\end{itemize}
& 
\begin{itemize}[leftmargin=0.3cm]
\item The authors propose a transformer-based summarizer that solves the issues of standard transformers that cannot take a long text as an input. They survey different use cases of applying transformers to different text summarization tasks and they only cover text summarization. no proposed transformers have been built in our survey.
\end{itemize}\\
 \hline
 
 R Kumar \citep{bert_survey_word_embedding}& Natural language Processing-NLP & 
 \begin{itemize}[leftmargin=0.3cm]
 \item Discussion of different NLP downstream tasks that BERT performs. BERT is covered in our survey as well as the different NLP tasks  
 \end{itemize}
 & 
 \begin{itemize}[leftmargin=0.3cm]
 \item Survey different techniques on using BERT as a word-embedder against traditional word-embedding techniques. Their survey is only focused as using transformers as a tool for embedding text
 \end{itemize}\\
 \hline
 
 F Acheampong et al. \citep{transformer_emotion_detection}& Natural language Processing-NLP & 
 
  \begin{itemize}[leftmargin=0.3cm]
 \item Survey different transformer architectures that accomplish the emotion detection task. We do the same, the application of different transformers to the same type if task 
 \end{itemize}
 & 
 \begin{itemize}[leftmargin=0.3cm]
 \item Survey the application of transformer architecture to a single application area but in too much detail, which is emotion detection from text-based data, a form of sentiment analysis but the goal is to extract fine-grained emotion from the data. The task of sentiment analysis is covered in our survey, but we didn't cover especially the task of detecting emotions on different levels and not just as a binary classification task as usually done in sentiment analysis
 \end{itemize}
 \\
 \hline
 
 R Gruetzemacher et al. \citep{deep_transfer_survey}& Natural language Processing-NLP & 
 \begin{itemize}[leftmargin=0.3cm]
 \item Survey the progress of transformers in the text-mining application area. We do cover in our survey the progress of transformers on a wide variety of tasks 
 \end{itemize}
 & 
 \begin{itemize}[leftmargin=0.3cm]
 \item Tackle the different transformers on how they can be used as text miners for organizations that have huge amounts of unstructured data against traditional NLP text-mining techniques
 \end{itemize}\\
 \hline

 J. Selva et al. \citep{selva2023video} & Computer Vision & 
\begin{itemize}[leftmargin=0.3cm]
\item This paper is an overview of transformers developed for modeling images and video data  
\end{itemize}
& 
\begin{itemize}[leftmargin=0.3cm]
\item This survey focuses solely on image and video data. Models are compared based on their performance in video classification, it does not cover any other applications. The paper proposes a taxonomy of various transformer models based on their recurrence properties, memory capacities, and architectural design
\end{itemize}\\
 \hline
 
 K. S. Kalyan et al. \citep{subramanyam2021ammu} & Natural language Processing-Medical & 
 \begin{itemize}[leftmargin=0.3cm]
 \item This paper provides an overview of the developed transformer-based BPLMs for a wide range of NLP tasks, including Natural language inference, Entity extraction, Relation extraction, Semantic textual similarity, Text classification, Question answering, and Text summarization 
 \end{itemize}
 & 
  \begin{itemize}[leftmargin=0.3cm]
 \item This survey addresses only transformer-based biomedical pre-trained language models, 
which restricts its scope to the specific field of biomedical natural language processing.
The taxonomy does not distinguish models based on the type of application they are used for, but rather based on the dataset of pre-training, the embedding type, and other criteria such as the targeted language
  \end{itemize}\\
 \hline

  K. Han et al. \citep{DBLP:journals/pami/00020C0GLTXXXYZ23} & Computer Vision & 
 \begin{itemize}[leftmargin=0.3cm]
 \item Categorized vision transformer models based on different tasks 
 \end{itemize}
 & 
  \begin{itemize}[leftmargin=0.3cm]
 \item This existing paper analyzed transformer models' advantages and disadvantages, and efficient transformer methods for the backbone network, while our survey categorizes transformer models based on tasks and summarize downstream tasks and commonly used dataset.

\item While our survey paper classified computer vision tasks into two segments: natural image processing \& medical image processing and then focused on popular computer vision like visual question answering, classification, segmentation, question answering, and so on, then this existing paper focused on high/mid-level vision, low-level vision, and video processing computer vision tasks.

\item This survey focused on computer vision tasks only, while we covered 
other four fields of applications-NLP, Multi-modal, Audio/speech, and signal processing besides computer vision
\end{itemize}\\
 \hline

Y. Xu  et al. \citep{xu2022transformers} & Computer Vision & 
\begin{itemize}[leftmargin=0.3cm]
\item The survey covers the fields of computer vision and multimodal in a similar fashion to our survey  
\end{itemize}
& 
\begin{itemize}[leftmargin=0.3cm]
\item This survey focuses primarily on recent advancements in computer vision by comparing the performance of different transformer models. Specifically, this study discusses four areas of research: advances in the design of the ViT models for image classification, high-level vision tasks (such as object detection and semantic segmentation), low-level vision tasks (such as super-resolution, and image generation), and multimodal learning (such as visual question answering (VQA), image captioning)
\end{itemize}\\
 \hline

J Li  et al. \citep{li2023transforming} & Computer Vision &

\begin{itemize}[leftmargin=0.3cm]
\item Comparative analysis of transformer models is presented in this paper for several tasks involved in medical vision. Several criteria are considered when comparing papers, including the type of dataset, the type of input data, and the architecture of the model 
\end{itemize}
& 
\begin{itemize}[leftmargin=0.3cm]
\item This paper describes in detail several transformer models that have been developed for medical images; however, it does not provide information regarding medical signals

\end{itemize}\\
 \hline
 
F Shamshad  et al. \citep{shamshad2023transformers} & Computer Vision-medical & 
\begin{itemize}[leftmargin=0.3cm]
\item A review of a number of transformer models with a focus on some tasks related to medical images and different image modalities, and a description of the datasets used for these tasks 
\end{itemize}
& 
\begin{itemize}[leftmargin=0.3cm]
\item This paper compares deep learning models starting with CNNs and moving up to vision transformers. In this paper, medical image modalities and several medical computer vision tasks are discussed to compare papers through the specification of datasets used and also provide an overview of models' performance.
In this paper, the comparison is based solely on medical images; medical signals are not considered
\end{itemize}\\
 \hline

 Salman Khan et al. \citep{DBLP:journals/csur/KhanNHZKS22} & Computer Vision & 
\begin{itemize}[leftmargin=0.3cm]
\item A overview of existing transformer computer vision models 
and classified the models based on popular tasks 
\end{itemize}
& 
 \begin{itemize}[leftmargin=0.3cm]
\item While this existing survey paper compared the popular techniques 
in terms of architectural design and experimental value, while 
our survey worked based on popular tasks and applications.

\item In the computer vision section, we put a special focus on Medical image tasks besides natural image processing.

\item This survey focused on computer vision tasks only, while we covered  other four fields of applications, namely NLP, Multi-modal, Audio/speech, and signal processing besides computer vision
\end{itemize}
\\
 \hline

L. Ruan et al. \citep{DBLP:journals/aiopen/RuanJ22}  & Multi-modal(NLP-CV) & 
 \begin{itemize}[leftmargin=0.3cm]
\item Categorize transformer vision-language models based on tasks and summarize downstream tasks and commonly used video dataset 
\end{itemize}
& 
 \begin{itemize}[leftmargin=0.3cm]
     \item  This existing survey focused on multi-modal(NLP-CV) tasks only, while we covered 
other four fields of applications-NLP, Computer vision, Audio/speech, and signal processing besides multi-modal
 \end{itemize}\\
 \hline
 
A Shin et al. \citep{pp_cross} & Multi-modal (Performance /Architecture) & 
 \begin{itemize}[leftmargin=0.3cm]
 \item They survey transformers for multi-modal tasks, which we do also include in our different application tasks 
 \end{itemize}
 & 
 \begin{itemize}[leftmargin=0.3cm]
 \item Cover only one application area in detail, which is multimodal visual-linguistic tasks
 \end{itemize}\\
 \hline
\caption{\label{tab:related works}Comparative summary between our survey and the existing surveys}

\end{longtable}

After having a thorough search and analysis of these survey papers, we realized that still, a survey on transformers is missing which focused on a most common field of application together and discussed the contribution of transformer-based models in the execution of different deep learning tasks in regarding fields of application. In this paper, first, we surveyed all transformer-based models out there based on our best possible search, identified top-$5$ fields of application and the proportion of transformers models’ contribution in the progression of top fields of application: Natural Language Processing (NLP),  Computer Vision (CV), Multi-modal, Audio and Speech, and Signal processing. Moreover, we proposed a taxonomy of transformer models based on these top five fields of application whereas top performed, and significant models are being classified and analyzed based on their task’s execution under the regarding fields. Through this survey, different aspects of Transformer-based models’ tasks and applications become more explicit in different fields, and it also depicted the fields of transformer applications that got higher and less attention by the researchers so far. Based on this analysis, we discussed future prospects and possibilities of transformers application in different fields of application. One of the objectives of this survey is to make a combined source of reference for a better understanding of the contribution of transformer models in different fields and the characteristics and execution methods of the models which kept significant contributions to improving the performance of the different tasks in their fields. Besides, this paper would be a resource to perceive future possibilities and scope of transformer-based models’ application for enthusiastic researchers who wants to extend and work for the new application of transformers.

\section{TRANSFORMER APPLICATIONS \label{sec:trans}} 

Since 2017, the transformer model has emerged as a highly attractive research model in the field of deep learning. Originally developed for processing long-range textual sentences. However, its scope has expanded to a variety of applications beyond NLP tasks. In fact, after a series of successes in NLP, researchers turned their attention to computer vision, exploring the potential of transformer models' global attention capability, while Convolutional Neural Networks (CNNs) were adept at tracking local features. 
The transformer model has also been tested and applied in various other fields and for various tasks. To gain a deeper understanding of transformer applications, we conducted a comprehensive search of various research libraries and reviewed the transformer models available from $2017$ to the present day. Our search yielded approximately more than $650$ transformer-based models that are being applied in various fields.

We identified the major fields in which transformer models are being used, including NLP, CV, Multi-modal applications, Audio and Speech processing, and Signal Processing. Our analysis provides an overview of the transformer models available in each field, their applications, and their impact on their respective industries.

\begin{figure}[htbp]
    \begin{center}
    \centering
    \includegraphics[width=.88\linewidth,height=.46\textwidth]{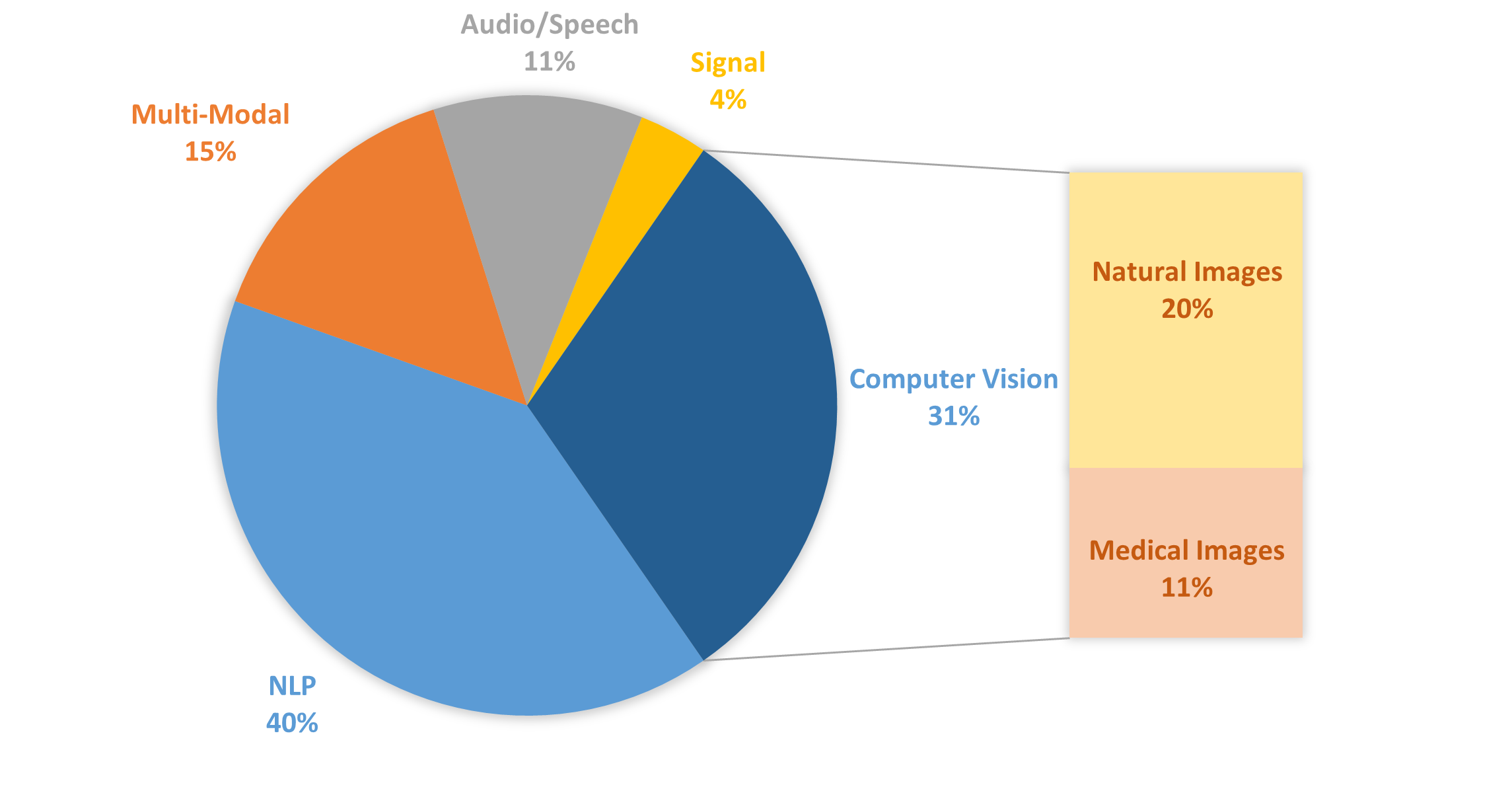}
    \caption{Proportion of transformer application in Top-5 fields}
    \label{fig:pie_chart}
    \vspace{-5pt} 
    \end{center}
\end{figure}


Figure  \ref{fig:pie_chart} shows  the percentage breakdown of the transformer models proposed so far across different application fields. Our analysis revealed approximately $250$ transformer-based models for NLP, representing around $40\%$ of the total transformer models collected. Moreover, we accounted for approximately $200$ models for computer vision. Due to the different processing of natural and medical images, and the extensive growth of both fields, we segmented computer vision into two categories: (i) Natural Image Processing and (ii) Medical Image Processing. As per this categorization, Natural Image Processing accounted for $20\%$ of transformer-based models, medical image processing accounted for $11\%$, and combinedly they accounted for $31\%$ of transformer-based models. Additionally, our analysis identified approximately $90$ transformer models for multi-modal applications, representing $15\%$ of the total, and around $70$ models for audio and speech processing, representing $11\%$ of total transformer models. Finally, only $4\%$ of transformer models were recorded for signal processing.

Our analysis provides a clear understanding of the proportion of attention received by transformer applications in each field, facilitating the identification of further research areas and tasks where transformer models can be implemented.

\section{APPLICATION-BASED CLASSIFICATION TAXONOMY OF TRANSFORMERS} \label{application}
\begin{figure}[!htbp]
  \centering
  \begin{adjustbox}{width=1.2\textwidth,center}
   \includegraphics{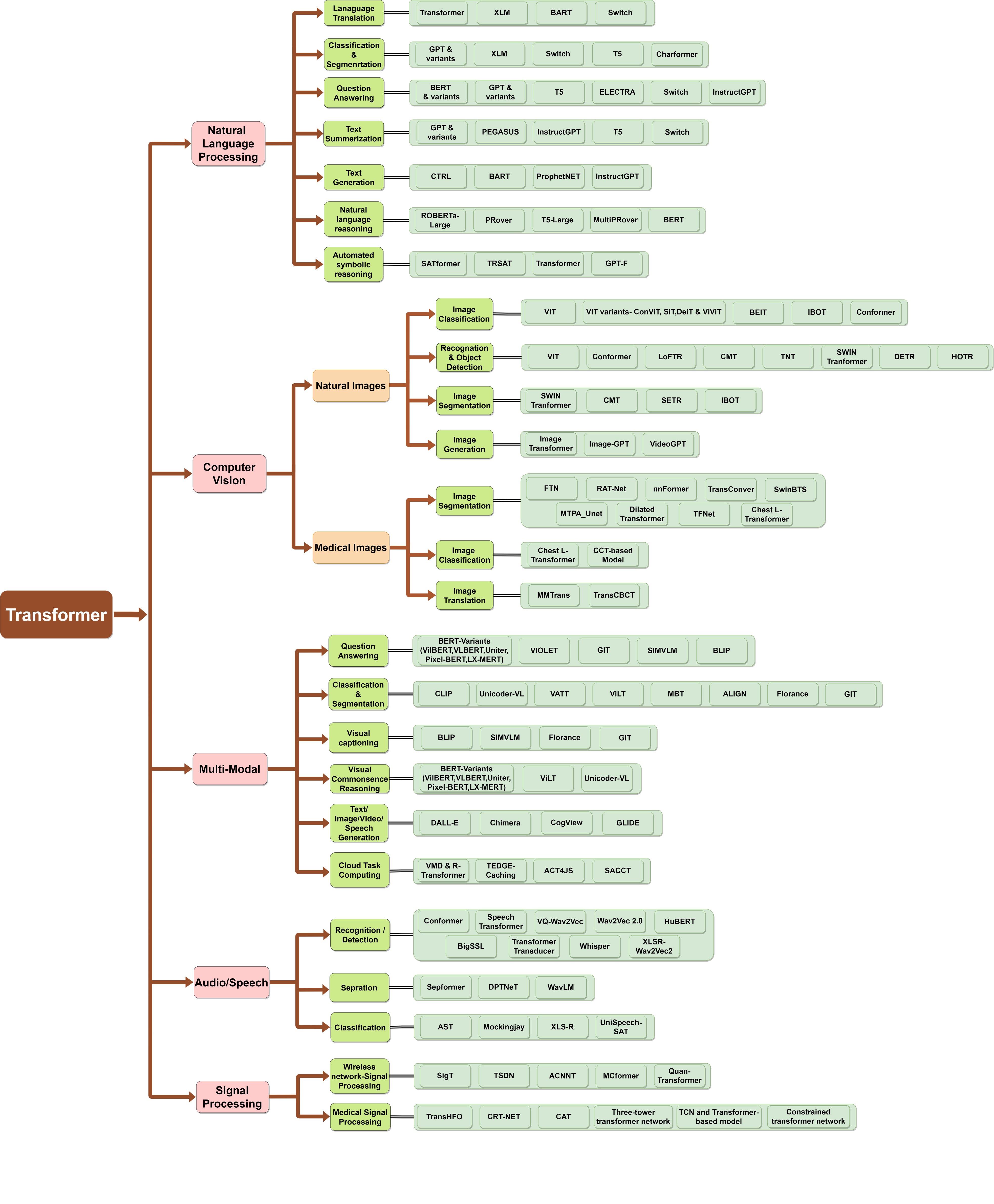}
  \end{adjustbox}
  \vspace{-50pt} 
  \caption{Application-based taxonomy of transformer models}
  \label{fig:taxonomy}
\end{figure}

As a result of conducting a thorough comprehensive analysis of all selected articles following the methodology explained in Section \ref{sec:met}, 
we noticed that the existing categorizations did not fully capture the wide range of transformer-based models and their diverse applications across different fields. Hence, in this study, we aimed to propose a more comprehensive taxonomy of transformers that would reflect their practical applications. 
To achieve this, we carefully reviewed a large number of transformer models and classified them based on their tasks within their respective fields of application. Our analysis identified several highly impactful and significant transformer-based models that have been successfully applied in a variety of fields. We then organized these models into five different application areas: Natural Language Processing (NLP), Computer Vision, Multi-modality, Audio and Speech, and Signal Processing.
The proposed taxonomy in Figure \ref{fig:taxonomy} provides a more nuanced and comprehensive framework for understanding the diverse applications of transformers. We believe that this taxonomy would be beneficial for researchers and practitioners working on transformer-based models, as it would help them to identify the most relevant models and techniques for their specific applications.

\subsection{Natural Language Processing (NLP)} \label{Sec:LT}

Transformers have become a vital tool in NLP, and various NLP tasks have largely benefited from these models. Our proposed taxonomy focuses on NLP and organizes transformer models into seven popular NLP tasks, including Translation, Summarization, Classification and Segmentation, Question Answering, Text Generation, Natural Language Reasoning, and Automated Symbolic Reasoning.
To ensure a comprehensive analysis, we only considered transformer models that have significantly impacted the NLP field and improved its performance. Our analysis included an in-depth discussion of each NLP task, along with essential information about each model presented in Table \ref{tab:language Translation}. We also highlighted the significance and working methods of each model.
This taxonomy provides a valuable framework for understanding the different transformer models used in NLP and their practical applications. It can help researchers and practitioners select the most appropriate transformer model for their specific NLP task. 

\subsubsection{Language Translation} \label{sec.6.1.1}

Language translation is a fundamental task in NLP, aimed at converting input text from one language to another. Its primary objective is to produce an output text that accurately reflects the meaning of the source text in the desired language. For example, given an English sentence as input text, the task aims to produce its equivalent in French or any other desired language.
The original transformer model was developed explicitly for the purpose of language translation, highlighting the significance of this task in the NLP field. Table \ref{tab:language Translation}  identifies the transformer-based models that have demonstrated significant performance in the Language Translation task. These models play a vital role in facilitating effective communication and collaboration across different languages, enabling more efficient information exchange and knowledge sharing.
Overall, the language translation task represents a crucial area of research in NLP, with significant implications for diverse applications, including business, science, education, and social interactions. The transformer-based models presented in the table offer promising solutions for advancing the state-of-the-art in this field, paving the way for new and innovative approaches to language translation \citep{NLP,language_translation1,language_translation2}.

\medskip
\setlength\LTleft{-0.87cm}
\setlength\LTright{\LTleft}
\begin{longtable}{| M{2cm} | M{2.5cm}| M{.7cm} |M{1.8cm} |M{1.6cm} |M{3cm} |M{4.4cm} |}

 \hline
\textbf{Transformer Models} & \textbf{Task Accomplished} & \textbf{Year} & \textbf{Architecture (Encoder/ Decoder)} & \textbf{Pre-trained (Yes/NO)} & \textbf{Pre-training Dataset} & \textbf{Dataset (Fine-tuning, Training, Testing)} \\
 \hline
 \endfirsthead
 
 \multicolumn{7}{c}%
{{\bfseries \tablename\ \thetable{} -- continued from previous page}} \\
 \hline
\textbf{Transformer Models} & \textbf{Task Accomplished} & \textbf{Year} & \textbf{Architecture (Encoder/ Decoder)} & \textbf{Pre-trained (Yes/NO)} & \textbf{Pre-training Dataset} & \textbf{Dataset(Fine-tuning, Training, Testing)} \\
 \hline
\endhead

\hline \multicolumn{7}{|r|}{{Continued on next page}} \\ \hline
\endfoot

\endlastfoot

Transformer-2017 \citep{r1} & Language Translation & 2017 & Encoder \& Decoder & No & NA & WMT 2014 English-German,WMT 2014 English-French \\
 \hline
XLM \citep{XLM} & Translation and Classification for multiple language & 2019 & Encoder \& Decoder & Yes & WMT’16, WMT’14 English-French, WMT’16 (English-German, English-Romanian, Romanian-English) & Wikipedia of 16 XNLI languages(English, French, Spanish, Russian, Arabic, Chinese, Hindi, German, Greek, Bulgarian, Turkish, Vietnamese, Thai, Urdu, Swahili, Japanese) \\
 \hline
BART \citep{BART} & Language Translation, Sentence Reconstruction, Comprehension, text Generation & 2019 & Encoder \& Decoder & Yes & Corrupting documents, 1M steps on a combination of books and Wikipedia data, news, stories, and web text (Training) & SQuAD, MNLI, ELI5, XSum, ConvAI2, CNN/DM, CNN/DailyMail, WMT16 Romanian-English, augmented with back-translation data from Sennrich et al. (2016). \\
 \hline
Switch Transformer \citep{Switch_Transformer} & Language understanding task- Translation, question answering, Classification, and so on. & 2021 & Encoder \& Decoder & Yes & C4(Colossal Clean Crawled Corpus) & GLUE and SuperGLUE benchmarks, CNNDM, BBC XSum, and SQuAD data sets, ARC Reasoning Challenge,3 closed-book question answering data sets (Natural Questions, Web Questions, and Trivia QA), Winogrande Schema Challenge, Adversarial NLI Benchmark\\
 \hline
\caption{\label{tab:language Translation}Transformer models for NLP - language translation task}
\end{longtable}

\begin{itemize}[leftmargin=.1in]
\item 
\textbf{Transformer: }
In 2017, Vaswani et al. \citep{r1} introduced the first transformer model, which has since revolutionized the field of NLP. The transformer model was designed specifically for language translation and is known as the Vanilla transformer model. Unlike its predecessors, the transformer model incorporates both an encoder and a decoder module, employing multi-head attention and masked-multi-head attention mechanisms. The encoder module is responsible for analyzing the contextual information of the input language, while the decoder module generates the output in the target language, using the output of the encoder and masked multi-head attention.
The transformer model's success is largely attributed to its ability to perform parallel computations, which allows it to process words simultaneously with positional information. This feature makes it highly efficient in processing large volumes of text and enables it to handle long-range dependencies, which are crucial in language translation.

\item \textbf{XLM: }
It is a cross-lingual language pretraining model developed to support multiple languages. The model is built using two methods: a supervised method and an unsupervised method. The unsupervised method utilizes Masked Language Modeling (MLM) and Casual Language Modeling (CLM) techniques and has shown remarkable effectiveness in translation tasks. On the other hand, the supervised method has further improved the translation tasks \citep{XLM}. This combination of supervised and unsupervised learning has made the XLM model a powerful tool for cross-lingual applications, making it possible to perform natural language processing tasks in multiple languages. The effectiveness of the XLM model in translation tasks has made it a popular choice among researchers in the field of natural language processing.

\item \textbf{BART:}
BART (Bidirectional and Auto-Regressive Transformers) is an advanced pre-trained model primarily aimed at cleaning up the corrupt text. It features two pre-training stages: the first stage corrupts the text with noise, while the second stage focuses on recovering the original text from the noisy version. BART employs a transformer translation model that integrates both the encoder and decoder modules, allowing it to perform various tasks such as text generation, translation, and comprehension with impressive accuracy \citep{BART}. Its bi-directional approach enables it to learn from the past and future tokens, while its auto-regressive properties make it suitable for generating output tokens sequentially. These features make BART an incredibly versatile model for various natural language processing tasks.

\item \textbf{Switch Transformer:} 
The Switch transformer model is a recent development in the field of NLP that has gained attention for its ability to perform various tasks with high accuracy. It incorporates two key components: a permutation-based routing mechanism and a gating mechanism. 
The permutation-based routing mechanism allows the model to learn a routing strategy that selects which parts of the input sequence to attend to. This enables the model to handle variable-length inputs, as it can dynamically determine which parts of the sequence to attend to for each input.
The gating mechanism allows the model to perform both Classification and Segmentation tasks. The gating mechanism is designed to learn how to combine information from different parts of the input sequence in order to make predictions. This allows the model to perform Classification tasks by predicting a label for the entire input sequence, or Segmentation tasks by predicting labels for each part of the input sequence.
The Switch transformer is a highly versatile model that can effectively perform both Classification and Segmentation tasks \citep{BogatinovskiTDK22}. Its detailed description can be found in the dedicated Classification \& Segmentation section below \citep{Switch_Transformer}.

\end{itemize}

\subsubsection{Classification \& Segmentation} \label{sec.6.1.2}

Text classification and segmentation are fundamental tasks in natural language processing (NLP) that enable the automatic organization and analysis of large volumes of textual data. Text classification involves assigning tags or labels to text based on its contents, such as sentiment, topic, or intent, among others. This process helps to categorize textual documents from different sources and can be useful in a variety of applications, such as recommendation systems, information retrieval, and content filtering.
On the other hand, text segmentation involves dividing the text into meaningful units, such as sentences, words, or topics, to facilitate further analysis or processing. This task is crucial for various NLP applications, including language understanding, summarization, and question answering, among others \citep{NLP, NLP_classficiation, NLP_segmentation}.

Transformer-based models have been shown to achieve state-of-the-art performance in text classification and segmentation tasks. These models are characterized by their ability to capture long-range dependencies and contextual information in text, making them well-suited for complex NLP tasks. Table \ref{tab:classification & segmentation} highlights some of the most prominent transformer-based models that have demonstrated significant performance in text classification and segmentation tasks.

\medskip
\setlength\LTleft{-0.87cm}
\setlength\LTright{\LTleft}
\begin{longtable}{| M{2cm} | M{2.5cm}| M{.7cm} |M{1.8cm} |M{1.6cm} |M{3cm} |M{4.4cm} |}

  \hline
\textbf{Transformer Models} & \textbf{Task Accomplished} & \textbf{Year} & \textbf{Architecture (Encoder/ Decoder)} & \textbf{Pre-trained (Yes/NO)} & \textbf{Pre-training Dataset} & \textbf{Dataset (Fine-tuning, Training, Testing)} \\
 \hline
 \endfirsthead
 
 \multicolumn{7}{c}%
{{\bfseries \tablename\ \thetable{} -- continued from previous page}} \\
 \hline
\textbf{Transformer Models} & \textbf{Task Accomplished} & \textbf{Year} & \textbf{Architecture (Encoder/ Decoder)} & \textbf{Pre-trained (Yes/NO)} & \textbf{Pre-training Dataset} & \textbf{Dataset (Fine-tuning, Training, Testing)} \\
 \hline
\endhead

\hline \multicolumn{7}{|r|}{{Continued on next page}} \\ \hline
\endfoot

\endlastfoot

GPT \& variants \citep{GPT_1,GPT_2,GPT_3} & Text classification, Question answering, textual entailment, semantic similarity & 2018 & Decoder & Yes & Book corpus & SNLI, MNLI, QNLI, SciTail, RTE, RACE, CNN, SQuaD, MRPC, QQP, STS-B, SST2 \& CoLA \\
 \hline
XLM \citep{XLM} & Translation and Classification for multiple language & 2019 & Encoder \& Decoder & Yes & WMT’16, WMT’14 English-French, WMT’16 (English-German, English-Romanian, Romanian-English) & Wikipedia of 16 XNLI languages(English, French, Spanish, Russian, Arabic, Chinese, Hindi, German, Greek, Bulgarian, Turkish, Vietnamese, Thai, Urdu, Swahili, Japanese) \\
 \hline
T5 \citep{T5} & Text summarization, Question answering, text classification & 2020 & Encoder \& Decoder & Yes & C4 (Colossal Clean Crawled Corpus) & GLUE and SuperGLUE benchmarks, CNN/Daily Mail abstractive summarization, SQuAD question answering, WMT English to German, French, and Romanian translation \\
 \hline
Charformer \citep{Charformer} & Classification task, toxicity detection, and so on. & 2022 & Encoder \& Decoder & Yes & The same datasets used in T5 model-C4(Colossal Clean
Crawled Corpus) & GLUE IMDb, AGNews,(Maas et al., 2011), (Zhang et al., 2015), Civil Comments, Wikipedia Comments, TyDiQA-GoldP, XQuAD, MLQA, XNLI, PAWS-X.. \\
 \hline
Switch Transformer \citep{Switch_Transformer} & Language understanding task- Translation, question answering, Classification, and so on. & 2021 & Encoder \& Decoder & Yes & C4(Colossal Clean Crawled Corpus) & GLUE and SuperGLUE benchmarks, CNNDM, BBC XSum and SQuAD data sets, ARC Reasoning Challenge,3 closed-book question answering data sets (Natural Questions, Web Questions, and Trivia QA), Winogrande Schema Challenge, Adversarial NLI Benchmark\\
 \hline
\caption{\label{tab:classification & segmentation}Transformer models for NLP - language classification \& segmentation tasks}
\end{longtable}

\begin{itemize}[leftmargin=.1in]

\item \textbf{Charformer: }
It is a transformer-based model that introduces Gradient-based subword tokenization (GBST), a lightweight approach to learning latent subwords directly from characters at the byte level. The model has both English and multi-lingual variants and has demonstrated outstanding performance on language understanding tasks, such as the classification of long text documents \citep{Charformer}.

\item \textbf{Switch Transformer: }
The use of pre-trained models such as BERT and GPT, trained on large datasets, has gained popularity in the field of natural language processing. However, there are concerns about the economic and environmental costs of training such models. To address these concerns, the Switch transformer was introduced, which offers a larger model size without a significant increase in computational cost. The Switch transformer replaces the feed-forward neural network (FFN) with a switch layer that contains multiple FFNs, resulting in a model with trillions of parameters. Despite the increase in model size, the computational cost of the Switch transformer remains comparable to that of other models. In fact, the Switch transformer has been evaluated on 11 different tasks and has shown significant improvement in tasks such as translation, question-answering, classification, and summarization \citep{Switch_Transformer}.\\

\textbf{GPT \& Variants, XLM, T5: }
These models are versatile and capable of performing a range of NLP tasks, including but not limited to classification, segmentation, question answering, and language translation. Section \ref{Question_Answering} will provide detailed description.

\end{itemize}
\subsubsection{Question Answering} \label{Question_Answering}

Question Answering  is a classical NLP task. It involves matching a text query to the most relevant answer in the form of text, based on the relevance of the text to the query. This task is challenging, as finding a concise and accurate answer to a given query can be difficult \citep{NLP,NLP_question_answering}. Recent research has focused on this task, leading to the development of several transformer-based models that leverage deep learning techniques to improve the accuracy and efficiency of this task. A detailed overview of these models is provided in Table \ref{tab:Question Answering}.

\medskip
\setlength\LTleft{-0.87cm}
\setlength\LTright{\LTleft}
\begin{longtable}{| M{2cm} | M{2.5cm}| M{.7cm} |M{1.8cm} |M{1.6cm} |M{3cm} |M{4.4cm} |}

 \hline
\textbf{Transformer Models} & \textbf{Task Accomplished} & \textbf{Year} & \textbf{Architecture (Encoder/ Decoder)} & \textbf{Pre-trained (Yes/NO)} & \textbf{Pre-training Dataset} & \textbf{Dataset (Fine-tuning, Training, Testing)} \\
 \hline
 \endfirsthead
 
 \multicolumn{7}{c}%
{{\bfseries \tablename\ \thetable{} -- continued from previous page}} \\
 \hline
\textbf{Transformer Models} & \textbf{Task Accomplished} & \textbf{Year} & \textbf{Architecture (Encoder/ Decoder)} & \textbf{Pre-trained (Yes/NO)} & \textbf{Pre-training Dataset} & \textbf{Dataset(Fine-tuning, Training, Testing)} \\
 \hline
\endhead

\hline \multicolumn{7}{|r|}{{Continued on next page}} \\ \hline
\endfoot

\endlastfoot

BERT \citep{BERT} & Question answering, Sentence Prediction, language understanding & 2018 & Encoder & Yes & Book Corpus, English Wikipedia & SQuAD v1.1, SQuAD v2.0, SWAG, QNLI, MNLI \\
 \hline
ELECTRA \citep{Electra} & Language understanding tasks- Question answering and so on & 2020 & Encoder & Yes & Wikipedia, BooksCorpus, ClueWeb, CommonCrawl, Gigaword & SQuAD 1.1, SQuAD 2.0, GLUE \\
\hline
 GPT \& variants \citep{GPT_1, GPT_2,GPT_3} & Text classification, Question answering, textual entailment, semantic similarity & 2018 & Decoder & Yes & Book corpus & SNLI, MNLI, QNLI, SciTail, RTE, RACE, CNN, SQuaD, MRPC, QQP, STS-B, SST2 \& CoLA \\
\hline
Switch Transformer \citep{Switch_Transformer} & Language understanding task- Translation, question answering, Classification, and so on. & 2021 & Encoder \& Decoder & Yes & C4(Colossal Clean Crawled Corpus) & GLUE and SuperGLUE benchmarks, CNNDM, BBC XSum, and SQuAD data sets, ARC Reasoning Challenge,3 closed-book question answering data sets (Natural Questions, Web Questions, and Trivia QA), Winogrande Schema Challenge, Adversarial NLI Benchmark\\
\hline
T5 \citep{T5} & Text summarization, Question answering, text classification & 2020 & Encoder \& Decoder & Yes & C4 (Colossal Clean Crawled Corpus) & GLUE and SuperGLUE benchmarks, CNN/Daily Mail abstractive summarization, SQuAD question answering, WMT English to German, French, and Romanian translation \\
 \hline
InstructGPT \citep{InstructGPT} & Text Generation, Question Answering, summarization, and so on. & 2022 & Decoder & Yes & Based on the pre-training model GPT-3 & SFT dataset, RM dataset, PPO dataset, a dataset of prompts and completions Winogender, CrowS-Pairs, Real Toxicity Prompts, TruthfulQA, DROP, QuAC, SquadV2, Hellaswag, SST, RTE and WSC, WMT 15 Fr ! En, CNN/Daily Mail Summarization, Reddit TLDR Summarization datasets.\\
 \hline
\caption{\label{tab:Question Answering}Transformer models for NLP - question answering task}
\end{longtable}

\begin{itemize}[leftmargin=.1in]

\item \textbf{BERT \& BERT variants: } 
BERT is an acronym that stands for Bidirectional Encoder Representations of transformers. It was introduced by the Google AI team and is embedded within the encoder module of the transformer. BERT employs a bidirectional approach, allowing it to pre-train a transformer on unannotated text by considering the context of each word. As a result, BERT has achieved remarkable performance on various natural language processing (NLP) tasks \citep{BERT}.
\\
A variety of BERT-based models have been developed with different characteristics. For instance, some are optimized for fast computation, while others produce superior results with a small number of parameters. Some are also tailored to specific tasks, such as \textbf{RoBERTa}, which is designed for masked language modeling and next sentence prediction \citep{Robert}. \textbf{FlueBERT} is another model that can be used for tasks such as text classification, paraphrasing, natural language inference, parsing, and word sense disambiguation \citep{FlueBERT}. Additionally, \textbf{DistilBERT} is suitable for question answering and other specific tasks. These models have significantly improved pre-trained transformer models \citep{DistilBERT}.

\item \textbf{GPT \& GPT variants:}
Generative Pre-Trained Transformer (GPT) models are built exclusively on the decoder block of transformers, which significantly improves the progress of transformers in natural language processing. GPT adopts a semi-supervised approach to language comprehension, which involves unsupervised pre-training and supervised fine-tuning methods \citep{GPT_1}. In 2019, following the success of the GPT model, a massively pre-trained transformer-based model called GPT-2 with 1.5 billion parameters was introduced, which significantly improved the pre-trained version of transformers \citep{GPT_2}. Subsequently, in 2020, the largest pre-trained version of GPT with 175 billion parameters, called GPT-3, was released. This model is 10 times larger than the previous non-sparse language model. One of the most notable achievements of GPT-3 is that it exhibits strong performance across a range of tasks without the need for gradient updates or fine-tuning, which is a requirement for pre-training models like BERT \citep{GPT_3}.

\item \textbf{Electra:}
An acronym for “Efficiently Learning an Encoder that Classifies Token Replacements Accurately”, utilizes a distinct pre-training method compared to other pre-trained models. Electra deploys a "Masked Language Modeling" approach that masks certain words and trains the model to predict them. Additionally, Electra incorporates a "Discriminator" network that aids in comprehending language without the need to memorize the training data. This unique approach enables Electra to generate superior text and surpass the performance of BERT \citep{Electra}.\\

\textbf{InstructGPT, T5 and Switch Transformer: } 
While the InstructGPT model can generate text apart from question-answering tasks, the T5 is significant in test summarization tasks and Switch transformer models can perform classification and segmentation tasks as well. More descriptions of these models have been provided in Sections \ref{sec.6.1.1} and \ref{sec.6.1.2}.

\end{itemize}

\subsubsection{Text Summarization}
Text summarization is a natural language processing task that involves breaking down lengthy texts into shorter versions while retaining essential and valuable information and preserving the meaning of the text. Text summarization is particularly useful in comprehending lengthy textual documents, and it also helps to reduce computational resources and time \citep{NLP,NLP_summerization}. transformer-based models have shown exceptional performance in text summarization tasks. The  transformer-based models in text summarization are listed in Table \ref{tab:Summarization}.

\medskip
\setlength\LTleft{-0.87cm}
\setlength\LTright{\LTleft}
\begin{longtable}{| M{2cm} | M{2.5cm}| M{.7cm} |M{1.8cm} |M{1.6cm} |M{3cm} |M{4.4cm} |}

 \hline
\textbf{Transformer Models} & \textbf{Task Accomplished} & \textbf{Year} & \textbf{Architecture (Encoder/ Decoder)} & \textbf{Pre-trained (Yes/NO)} & \textbf{Pre-training Dataset} & \textbf{Dataset (Fine-tuning, Training, Testing)} \\
 \hline
 \endfirsthead
 
 \multicolumn{7}{c}%
{{\bfseries \tablename\ \thetable{} -- continued from previous page}} \\
 \hline
\textbf{Transformer Models} & \textbf{Task Accomplished} & \textbf{Year} & \textbf{Architecture (Encoder/ Decoder)} & \textbf{Pre-trained (Yes/NO)} & \textbf{Pre-training Dataset} & \textbf{Dataset (Fine-tuning, Training, Testing)} \\
 \hline
\endhead

\hline \multicolumn{7}{|r|}{{Continued on next page}} \\ \hline
\endfoot

\endlastfoot
 \hline
 GPT \& variants \citep{GPT_1,GPT_2,GPT_3} & Text classification, Question answering, textual entailment, semantic similarity & 2018 & Decoder & Yes & Book corpus & SNLI, MNLI, QNLI, SciTail, RTE, RACE, CNN, SQuaD, MRPC, QQP, STS-B, SST2 \& CoLA \\
 \hline
PEGASUS \citep{Pegasus} & Text summarization & 2020 & Encoder \& Decoder & Yes & C4, HugeNews & XSum, CNN/DailyMail, NEWSROOM, Multi-News, Gigaword, arXiv, PubMed, BIGPATENT, WikiHow, Reddit TIFU, AESLC,BillSum \\
\hline
Switch Transformer \citep{Switch_Transformer} & Language understanding task- Translation, question answering, Classification, and so on. & 2021 & Encoder \& Decoder & Yes & C4(Colossal Clean Crawled Corpus) & GLUE and SuperGLUE benchmarks, CNNDM, BBC XSum, and SQuAD data sets, ARC Reasoning Challenge,3 closed-book question answering data sets (Natural Questions, Web Questions, and Trivia QA), Winogrande Schema Challenge, Adversarial NLI Benchmark\\
\hline
T5 \citep{T5} & Text summarization, Question answering, text classification & 2020 & Encoder \& Decoder & Yes & C4 (Colossal Clean Crawled Corpus) & GLUE and SuperGLUE benchmarks, CNN/Daily Mail abstractive summarization, SQuAD question answering, WMT English to German, French, and Romanian translation \\
\hline
InstructGPT \citep{InstructGPT} & Text Generation, Question Answering, summarization, and so on. & 2022 & Decoder & Yes & Based on the pre-training model GPT-3 & SFT dataset, RM dataset, PPO dataset, a dataset of prompts and completions Winogender, CrowS-Pairs, Real Toxicity Prompts, TruthfulQA, DROP, QuAC, SquadV2, Hellaswag, SST, RTE, and WSC, WMT 15 Fr ! En, CNN/Daily Mail Summarization, Reddit TLDR Summarization datasets.\\
 \hline
\caption{\label{tab:Summarization}Transformer models for NLP - text summarization task}
\end{longtable}

\begin{itemize}[leftmargin=.1in]

\item \textbf{PEGASUS: } 
It is an exemplary model for generative text summarization that employs both the encoder and decoder modules of the transformer. While models based on masked language modeling only mask a small portion of text, PEGASUS masks entire multiple sentences, selecting the masked sentences based on their significance and importance, and generating them as the output. The model has exhibited significant performance on unknown summarization datasets  \citep{Pegasus}.

\item \textbf{T5: }
The T5 transformer model, which stands for Text-To-Text Transfer Transformer, introduced a dataset named ``Colossal Clean Crawled Corpus (C4)" that  improved the performance in various downstream NLP tasks. T5 is a multi-task model that can be trained to perform a range of NLP tasks using the same set of parameters. Following pre-training, the model can be fine-tuned for different tasks and achieves comparable performance to several task-specific models \citep{T5}.\\

\textbf{GPT \& variants, InstructGPT and Switch Transformer: }
These models have been discussed in earlier sections. Moreover, apart from text summarization, certain models such as GPT and its variants can perform question-answering tasks, InstructGPT can generate text, and Charformer models are capable of classification and segmentation tasks as well.

\end{itemize}
\subsubsection{Text Generation}

The task of text generation has gained immense popularity in the field of NLP due to its usefulness in generating long-form documentation, among other applications. Text generation models attempt to derive meaning from trained text data and create a connection between the text that has been previously outputted. These models typically operate on the basis of this connection \citep{NLP,NLP_generation}. The use of transformer-based models has led to significant advancements in the task of text generation. Please refer to Table \ref{tab:Text Generation}.

\medskip
\setlength\LTleft{-0.87cm}
\setlength\LTright{\LTleft}
\begin{longtable}{| M{2cm} | M{2.5cm}| M{.7cm} |M{1.8cm} |M{1.6cm} |M{3cm} |M{4.5cm} |}

 \hline
\textbf{Transformer Models} & \textbf{Task Accomplished} & \textbf{Year} & \textbf{Architecture (Encoder/ Decoder)} & \textbf{Pre-trained (Yes/NO)} & \textbf{Pre-training Dataset} & \textbf{Dataset (Fine-tuning, Training, Testing)} \\
 \hline
 \endfirsthead
 
 \multicolumn{7}{c}%
{{\bfseries \tablename\ \thetable{} -- continued from previous page}} \\
 \hline
\textbf{Transformer Models} & \textbf{Task Accomplished} & \textbf{Year} & \textbf{Architecture (Encoder/ Decoder)} & \textbf{Pre-trained (Yes/NO)} & \textbf{Pre-training Dataset} & \textbf{Dataset (Fine-tuning, Training, Testing)} \\
 \hline
\endhead

\hline \multicolumn{7}{|r|}{{Continued on next page}} \\ \hline
\endfoot

\endlastfoot
\hline
CTRL \citep{CTRL} & Text Generation & 2019 & Encoder \& Decoder & Yes & Project Gutenberg, subreddits, News Data, Amazon Review, open WebText, WMT Translation date, question-answer pairs, MRQA & Multilingual Wikipedia and Open WebText.\\

\hline
BART \citep{BART} & Language Translation, Sentence Reconstruction, Comprehension, text Generation & 2019 & Encoder \& Decoder & Yes & Corrupting documents, 1M steps on a combination of books and Wikipedia data, news, stories, and web text (Training) & SQuAD, MNLI, ELI5, XSum, ConvAI2, CNN/DM, CNN/DailyMail, WMT16 Romanian-English, augmented with back-translation data from Sennrich et al. (2016). \\
\hline
ProphetNET \citep{ProphetNet} & Text Prediction & 2020 & Encoder \& Decoder & Yes & Bookcorpus, English Wikipedia news, books, stories, and web text & CNN/dailymail, Giga-word Corpus, SQuAD dataset. \\
 \hline
InstructGPT \citep{InstructGPT} & Text Generation, Question Answering, summarization and so on. & 2022 & Decoder & Yes & Based on the pre-training model GPT-3 & SFT dataset, RM dataset, PPO dataset, dataset of prompts and completions Winogender, CrowS-Pairs, Real Toxicity Prompts , TruthfulQA , DROP , QuAC , SquadV2 , Hellaswag , SST , RTE and WSC, WMT 15 Fr ! En, CNN/Daily Mail Summarization, Reddit TLDR Summarization datasets.\\
 \hline
\caption{\label{tab:Text Generation}Transformer models for NLP - text generation task}
\end{longtable}

\begin{itemize}[leftmargin=.1in]
\item\textbf{CTRL:} 
The acronym CTRL denotes the Conditional transformer Language model, which excels in generating realistic text resembling human language, contingent on a given condition. In addition, CTRL can produce text in multiple languages. This model is large-scale, boasting 1.63 billion parameters, and can be fine-tuned for various generative tasks, such as question answering and text summarization \citep{CTRL}.

\item \textbf{ProphetNET:} 
ProphetNET is a sequence-to-sequence model that utilizes future n-gram prediction to facilitate text generation by predicting n-grams ahead. The model adheres to the transformer architecture, comprising encoder and decoder modules. It distinguishes itself by employing an n-stream self-attention mechanism. ProphetNET demonstrates remarkable performance in summarization and is also competent in question generation tasks \citep{ProphetNet}.

\item \textbf{InstructGPT: } 
It was proposed as a solution to the problem of language generative models failing to produce realistic and truthful results. It achieves this by incorporating human feedback during fine-tuning and reinforcement learning from the feedback. The GPT-3 model was fine-tuned for this purpose. As a result, InstructGPT can generate more realistic and natural output that is useful in real-life applications. ChatGPT, which follows a similar methodology as InstructGPT, has gained significant attention in the field of NLP at the end of 2022 \citep{InstructGPT}. \\

\textbf{BART:}
The BART model’s description is mentioned above in the language translation section. This model can execute the language translation task as well.
\end{itemize}
\subsubsection{Natural Language Reasoning}


The pursuit of natural language reasoning is a field of study that is distinct from that of question-answering. Question-answering focuses on finding the answer to a specific query within a given text passage. On the other hand, natural language reasoning involves the application of deductive reasoning to derive a conclusion from the given premises and rules that are represented in natural language. Neural network architectures aim to learn how to utilize these premises and rules to infer new conclusions. Previously, a similar task was traditionally tackled by systems equipped with the knowledge represented in a formal format and rules to be applied for the derivation of new knowledge. However, the use of formal representation has posed a significant challenge to this line of research \citep{chal_lang}. With the advent of transformers and their remarkable performance in numerous NLP tasks, it is now possible to circumvent formal representation and allow transformers to engage in reasoning directly using natural language.  Table \ref{tab:reasoning} highlights some of the
significant transformer models for natural language reasoning tasks.





\medskip
\setlength\LTleft{-0.86cm}
\setlength\LTright{\LTleft}
\begin{longtable}{| M{3cm} | M{2.8cm}| M{1cm} |M{2.5cm} |M{1.6cm} |M{2cm} |M{3cm} |}

 \hline
\textbf{Transformer Models} & \textbf{Task Accomplished} & \textbf{Year} & \textbf{Architecture (Encoder/ Decoder)} & \textbf{Pre-trained (Yes/NO)} & \textbf{Pre-training Dataset} & \textbf{Dataset (Fine-tuning, Training, Testing)} \\
 \hline
 \endfirsthead
 
 \multicolumn{7}{c}%
{{\bfseries \tablename\ \thetable{} -- continued from previous page}} \\
 \hline
\textbf{Transformer Models} & \textbf{Task Accomplished} & \textbf{Year} & \textbf{Architecture (Encoder/ Decoder)} & \textbf{Pre-trained (Yes/NO)} & \textbf{Pre-training Dataset} & \textbf{Dataset (Fine-tuning, Training, Testing)} \\
 \hline
\endhead

\endfoot

\endlastfoot
(Clark et al., 2020) \citep{Roberta-Clark} & Binary Classification & 2020 & RoBERTa (Encoder) & Yes & RACE & RuleTaker \\
\hline
(Richardson et al., 2022) \citep{Roberta-large} & Binary Classification & 2022 & RoBERTa Large (Encoder) & Yes & RACE & Hard-RuleTaker \\
\hline

(Saha et al., 2020) \citep{PRover} & Binary Classification, Sequence Generation & 2020 & PRover [RoBERTa-based](Encoder) & No & NA & RuleTaker \\
\hline

(Sinha et al., 2019) \citep{CLUTRR} & Sequence Generation & 2019 & BERT (Encoder) & No & NA & CLUTRR \\ 
\hline

(Picco et al., 2021) \citep{BERT-Based} & Binary Classification & 2021 & BERT-Based (Encoder) & Yes & RACE & RuleTaker \\
 \hline
\caption{\label{tab:reasoning}Transformer models for natural language reasoning}
\end{longtable}


\begin{itemize}[leftmargin=.1in]
\item \textbf{RoBERTa: } 
In a 2020 study by \cite{Roberta-Clark}, a binary classification task was assigned to the transformer, which aimed to determine whether a given statement can be inferred from a provided set of premises and rules represented in natural language. The architecture utilized for the transformer was RoBERTa-large, which was pre-trained on a dataset of high school exam questions that required reasoning skills. This pre-training enabled the transformer to achieve a high accuracy of $98\%$ on the test dataset. The dataset contained theories that were randomly sampled and constructed using sets of names and attributes. The task required the transformer to classify whether the given statement (Statement) followed from the provided premises and rules (Context) \citep{Roberta-Clark}

\item \textbf{RoBERTa-Large: }
In the work by \cite{Roberta-large}, the authors aimed to address a limitation of the dataset construction approach presented in the work by \cite{Roberta-Clark}. They highlighted that the uniform random sampling of theories, as done in \citep{Roberta-Clark}, does not always result in challenging instances. To overcome this limitation, they proposed a novel methodology for creating more challenging algorithmic reasoning datasets. The key idea of their methodology is to sample hard instances from ordinary SAT propositional formulas and translate them into natural language using a predefined set of English rule languages. By following this approach, they were able to construct a more challenging dataset that is consequential for training robust models and for reliable evaluation.
To demonstrate the effectiveness of their approach, the authors conducted experiments where they tested the models trained using the dataset from \citep{Roberta-Clark} on their newly constructed dataset. The results showed that the models achieved an accuracy of 57.7\% and 59.6\% for T5 and RoBERTa, respectively. These findings highlight that models trained on easy datasets may not be capable of solving challenging instances of the problem. 
\item \textbf{PRover:} 
In a related study, \cite{PRover} proposed a model called PRover, which is an interpretable joint transformer capable of generating a corresponding proof with an accuracy of $87\%$. The task addressed by PRover is the same as that in the study by \cite{Roberta-Clark} and \cite{Roberta-large}, where the aim is to determine whether a given conclusion follows from the provided premises and rules.
The proof generated by PRover is represented as a directed graph, where the nodes represent statements and rules, and the edges indicate which new statements follow from applying rules on the previous statements. Overall, the proposed approach by  \cite{PRover} provides a promising direction towards achieving interpretable and accurate reasoning models. 
\item \textbf{BERT-based:} 
In \citep{BERT-Based}, a BERT-based architecture called ``neural unifier" was proposed to improve the generalization performance of the model on the RuleTaker dataset. The authors aimed to mimic some elements of the backward-chaining reasoning procedure to enhance the model's ability to handle queries that require multiple steps to answer, even when trained on shallow queries only.
The neural unifier consists of two standard BERT transformers, namely the fact-checking unit and the unification unit. The fact-checking unit is trained to classify whether a query of depth 0, represented by the embedding vector q-0, follows from a given knowledge base represented by the embedding vector C. The unification unit takes as input the embedding vector q-n of a depth-n query and the embedding vector of the knowledge base, vector C, and tries to predict an embedding vector q0, thereby performing backward-chaining. 

\item \textbf{BERT: }
 \cite{CLUTRR} introduced a dataset named CLUTRR, which differs from the previously discussed studies in that the rules are not given in the input to be used to infer conclusions. Instead, the BERT transformer model is tasked with both extracting relationships between entities and inferring the rules governing these relationships. For instance, given a knowledge base consisting of statements such as ``Alice is Bob's mother" and ``Jim is Alice's father", the network can infer that ``Jim is Bob's grandfather".
\end{itemize}

\subsubsection{Automated symbolic reasoning}


Automated symbolic reasoning is a subfield of computer science that deals with solving logical problems such as SAT solving and automated theorem proving. These problems are traditionally addressed using search techniques with heuristics. However, recent studies have explored the use of learning-based techniques to improve the efficiency and effectiveness of these methods.
One approach is to learn the selection of efficient heuristics used by traditional algorithms. Alternatively, an end-to-end learning-based solution can be employed for the problem. Both approaches have shown promising results and offer the potential for further advancements in automated symbolic reasoning \citep{sat_learn, sat_learn2}. In this regard, a number of transformer based models have shown significant performance in automated symbolic reasoning tasks. For those models, please look at Table \ref{tab:naturalReasoning}.






\medskip
\setlength\LTleft{-0.86cm}
\setlength\LTright{\LTleft}
\begin{longtable}{| M{3cm} | M{2.8cm}| M{1cm} |M{2.5cm} |M{1.6cm} |M{2cm} |M{3cm} |}

 \hline
\textbf{Transformer Models} & \textbf{Task Accomplished} & \textbf{Year} & \textbf{Architecture (Encoder/ Decoder)} & \textbf{Pre-trained (Yes/NO)} & \textbf{Pre-training Dataset} & \textbf{Dataset (Fine-tuning, Training, Testing)} \\
 \hline
 \endfirsthead
 
 \multicolumn{7}{c}%
{{\bfseries \tablename\ \thetable{} -- continued from previous page}} \\
 \hline
\textbf{Transformer Models} & \textbf{Task Accomplished} & \textbf{Year} & \textbf{Architecture (Encoder/ Decoder)} & \textbf{Pre-trained (Yes/NO)} & \textbf{Pre-training Dataset} & \textbf{Dataset(Fine-tuning, Training, Testing)} \\
 \hline
\endhead

\endfoot

\endlastfoot
(Shi et al., 2022) \citep{SATformer-reasoning} & Binary Classification & 2022 & SATFormer (Encoder / Decoder) & No & NA & Synthetic\\
\hline
(Shi et al., 2021) \citep{TRSAT} & Binary Classification & 2021 & TRSAT (Encoder / Decoder) & No & NA & Synthetic, SATLIB \\
\hline

(Hahn et al., 2021) \citep{Transformer-reasoning} & Sequence Generation & 2021 & Transformer (Encoder / Decoder) & No & NA & Synthetic \\
\hline

(Polu et al, 2020) \citep{GPT-f} & Sequence Generation & 2020 & GPT-f (Decoder) & Yes & CommonCrawl, Github, arXiv, WebMath & set.mm \\ 
\hline
\caption{\label{tab:naturalReasoning}Transformer models for automated symbolic reasoning}
\end{longtable}

\begin{itemize}[leftmargin=.1in]

\item \textbf{SATformer:} 
The SAT-solving problem for boolean formulas was addressed by Shi et al. in 2022 \citep{TRSAT} through the introduction of SATformer, a hierarchical transformer architecture that offers an end-to-end learning-based solution for solving the problem. Traditionally, in the context of SAT-solving, a boolean formula is transformed into its conjunctive normal form (CNF), which serves as an input for the SAT solver. The CNF formula is a conjunction of boolean variables and their negations, known as literals, organized into clauses where each clause is a disjunction of these literals. For example, a CNF formula utilizing boolean variables would be represented as (A OR B) AND (NOT A OR C), where each clause (A OR B) and (NOT A OR C) is made up of literals.

The authors employ a graph neural network (GNN) to obtain the embeddings of the clauses in the CNF formula. SATformer then operates on these clause embeddings to capture the interdependencies among clauses, with the self-attention weight being trained to be high when groups of clauses that could potentially lead to an unsatisfiable formula are attended together, and low otherwise. Through this approach, SATformer efficiently learns the correlations between clauses, resulting in improved SAT prediction capabilities \citep{SATformer-reasoning}.

\item\textbf{TRSAT:} 
Another research endeavor conducted by Shi et al. in 2021 investigated a variant of the boolean SAT problem known as MaxSAT and introduced a transformer model named TRSAT, which serves as an end-to-end learning-based SAT solver \citep{TRSAT}. A comparable problem to the boolean SAT is the satisfiability of a linear temporal formula \citep{Pnueli77}, where a satisfying symbolic trace to the formula is sought after given a linear temporal formula.

\item \textbf{Transformer: } 
In a study conducted by \cite{Transformer-reasoning}, the authors addressed the boolean SAT problem and the temporal satisfiability problem, both of which are more complex than binary classification tasks that were tackled in previous studies. In these problems, the task is to generate a satisfying sequence assignment for a given formula, rather than simply classifying whether the formula is satisfied or not. The authors constructed their datasets by using classical solvers to generate linear temporal formulas with their corresponding satisfying symbolic traces, and boolean formulas with their corresponding satisfying partial assignments. 
The authors employed a standard transformer architecture to solve the sequence-to-sequence task. The Transformer was able to generate satisfying traces, some of which were not observed during training, demonstrating its capability to solve the problem and not merely mimic the behavior of the classical solvers used in the dataset generation process.

\item \textbf{GPT-f:} 
In their work, \cite{GPT-f} presented GPT-F, an automated prover and proof assistant that utilizes a decoder-only transformers architecture similar to GPT-2 and GPT-3. GPT-F was trained on a dataset called set.mm, which contains approximately 38,000 proofs. The largest model investigated by the authors consists of 36 layers and 774 million trainable parameters. This deep learning network has generated novel proofs that have been accepted and incorporated into mathematical proof libraries and communities.
\end{itemize}

\subsection{Computer Vision}
Motivated by the success of transformers in natural language processing, researchers have explored the application of the transformer concept in computer vision tasks. Traditionally, convolutional neural networks (CNNs) have been considered the fundamental component for processing visual data. However, different types of images require different processing techniques, with natural images and medical images being two prime examples. Furthermore, research in computer vision for natural images and medical images is vast and distinct. As a result, transformer models for computer vision can be broadly classified into two categories: (i) those designed for natural image processing, and (ii) those designed for medical image processing.

\subsubsection{Natural Image processing}
In the domain of computer vision, natural image processing is a primary focus as compared to medical image processing, owing to the greater availability of natural image data. Furthermore, computer vision with natural images has wide-ranging applications in various domains. Among the numerous tasks associated with computer vision and natural images, we have identified four of the most common and popular tasks: (i) classification and segmentation, (ii) recognition and feature extraction, (iii) mask modeling prediction, and (iv) image generation. In this context, we have provided a comprehensive discussion of each of these computer vision tasks with natural images. Additionally, we have presented a table that provides crucial information about each transformer-based model and have highlighted their working methods and significance.

\subsubsection{Image Classification}

Image classification is a crucial and popular task in the field of computer vision, which aims to analyze and categorize images based on their features, type, genre, or objects. This task is considered as a primary stage for many other image processing tasks. For example, if we have a set of images of different animals, we can classify them into different animal categories such as cat, dog, horse, etc., based on their characteristics and features \citep{image_classification,image_classification1}. Due to its significance, many transformer-based models have been developed to address image classification tasks. Table \ref{tab:image classification} highlights some of the significant transformer models for image classification tasks and discusses their important features and working methodologies.
\medskip
\medskip
\setlength\LTleft{-0.87cm}
\setlength\LTright{\LTleft}
\begin{longtable}{| M{2cm} | M{2.5cm}| M{.7cm} |M{1.8cm} |M{1.6cm} |M{3cm} |M{4.4cm} |}

 \hline
\textbf{Transformer Models} & \textbf{Task Accomplished} & \textbf{Year} & \textbf{Architecture (Encoder/ Decoder)} & \textbf{Pre-trained (Yes/NO)} & \textbf{Pre-training Dataset} & \textbf{Dataset (Fine-tuning, Training, Testing)} \\
 \hline
\endfirsthead
 


VIT \citep{Vit} & Image classification, image recognition & 2021 & Encoder & Yes & JFT-300M, ILSVRC-2012 ImageNet, ImageNet-21k & ImageNet-RL, CIFAR-10/100, Oxford Flowers-102, Oxford-IIIT Pets, VTAB \\
 \hline
ViT Variants \citep{Convit,Sit,DEIT,ViViT} & Image classification & 2020-2021 & Encoder & Yes & \textbf{ConViT:} ImageNet (Based on DeiT) \ \textbf{SiT:} STL10, CUB200, CIFAR10, CIFAR100, ImageNet-1K, Pascal VOC, MS-COCO, Visual-Genome \ \textbf{DEIT:} ImageNet. \ \textbf{ViViT:} ImageNet, JFT & \textbf{ConViT:} ImageNet,CIFAR100 \ \textbf{SiT:} CIFAR-10,CIFAR-100 , STL-10, CUB200, ImageNet-1K, Pascal VOC, MS-COCO, Visual-Genome. \ \textbf{DEIT:} ImageNet, iNaturalist 2018, iNaturalist 2019, Flowers-102, Stanford Cars, CIFAR-100, CIFAR-10. \textbf{ViViT:} Larger JFT, Kinetics, Epic Kitchens-100, Moments in Time, SSv2. \\
 \hline
BEIT \citep{BEiT} & Image classification \& segmentation & 2021 & Encoder & Yes & ImageNet-1K, ImageNet-22k & ILSVRC-2012 ImageNet, ADE20K, CIFAR-100 \\
 \hline
IBOT \citep{iBOT} & Image classification, segmentation, object detection \& recognition & 2022 & Encoder & Yes & ImageNet-1K, ViT-L/16, ImageNet-22K & COCO, ADE20K \\
 \hline
Conformer \citep{Conformer} & Image recognition \& object detection, Classification & 2021 & Encoder & Not mentioned in paper & N/A & LibriSpeech\\
 \hline
\caption{\label{tab:image classification}Transformer models for natural image processing - image classification}
\end{longtable}

\begin{itemize}[leftmargin=.1in]

\item \textbf{ViT Variants:}
There are several ViT-based models that have been developed for specific tasks. For instance, ConViT is an improved version of ViT that combines CNN and transformer by adding an inductive bias to ViT, resulting in better accuracy for image classification tasks \citep{Convit}. Self-supervised Vision transformer (SiT) allows for the use of the architecture as an autoencoder and seamlessly works with multiple self-supervised tasks \citep{Sit}. Data Efficient Image Transformer (DeiT) is a type of vision transformer designed for image classification tasks that require less data to be trained \citep{DEIT}. There are numerous ViT variants available with certain improvements or designed for specific tasks. For example, Video Vision Transformer (ViViT) is a ViT-based model that classifies videos using both encoder and decoder modules of the transformer, whereas most ViT and ViT-variant models use only the encoder module \citep{ViViT}.

\item \textbf{BEIT:}
Bidirectional Encoder Representation from Image Transformers (BEIT)  \citep{BEiT} is a transformer-based model that draws inspiration from BERT and introduces a new pre-training task called Masked Image Modeling (MIM) for vision Transformers. In MIM, a portion of the image is randomly masked, and the corrupted image is passed through the architecture, which then recovers the original image tokens. BEIT has shown competitive performance on image classification and segmentation tasks, demonstrating its effectiveness for a variety of computer vision applications.

\item \textbf{Conformer:}
In the field of computer vision, Conformer \citep{Conformer} is a model that works similarly to the CMT model. While CNN is responsible for capturing the local features of the image, Transformer works for the global context and long-range dependencies of images. However, the Conformer model proposes a new method called cross-attention, which combines both local and global features to focus on various parts of the image based on the task. The model has shown promising results for classification and object detection/recognition tasks.

\item \textbf{IBOT:}
IBOT represents Image BERT Pre-training with Online Tokennizer which is a self-supervised model. This model studied masked image modeling using an online tokenizer and it learns to distill features using a tokenizer. This online tokenizer helps this model to improve the feature representation capability. Besides, the image classification task, this model shows significant performance in object detection and segmentation tasks.\\

\textbf{ViT } 
The ViT model, which has been discussed in detail in the Recognition and Object Detection section, is also capable of performing recognition and object detection tasks.
\end{itemize}
\subsubsection{Image Recognition \& Object Detection} \label{sec:IR}
Image recognition \& Object detection is often considered as nearly similar and related task in computer vision. It is the capability of detecting or recognizing any object, person, or feature in an image or video. An image or video contains a number of objects \& features; by extracting the features from the image, a model tries to capture the features of an object through training. By understanding these useful features, a model can recognize the specific object from the other available object in the image or video \citep{object_detection1,object_detection2,image_recognition1,image_recognition2}. Here we highlight and discuss the significant transformer models for image/object recognition tasks (see Table  \ref{tab:image recognition}).

\medskip
\setlength\LTleft{-0.87cm}
\setlength\LTright{\LTleft}
\begin{longtable}{| M{2cm} | M{2.5cm}| M{.7cm} |M{1.8cm} |M{1.6cm} |M{3cm} |M{4.4cm} |}

 \hline
\textbf{Transformer Models} & \textbf{Task Accomplished} & \textbf{Year} & \textbf{Architecture (Encoder/ Decoder)} & \textbf{Pre-trained (Yes/NO)} & \textbf{Pre-training Dataset} & \textbf{Dataset (Fine-tuning, Training, Testing)} \\
 \hline
 \endfirsthead
 
 \multicolumn{7}{c}%
{{\bfseries \tablename\ \thetable{} -- continued from previous page}} \\
 \hline
\textbf{Transformer Models} & \textbf{Task Accomplished} & \textbf{Year} & \textbf{Architecture (Encoder/ Decoder)} & \textbf{Pre-trained (Yes/NO)} & \textbf{Pre-training Dataset} & \textbf{Dataset(Fine-tuning, Training, Testing)} \\
 \hline
\endhead

\hline \multicolumn{7}{|r|}{{Continued on next page}} \\ \hline
\endfoot

\endlastfoot
 \hline
VIT \citep{Vit} & Image classification, image recognition & 2021 & Encoder & Yes & JFT-300M, ILSVRC-2012 ImageNet, ImageNet-21k & ImageNet-RL, CIFAR-10/100, Oxford Flowers-102, Oxford-IIIT Pets, VTAB \\
 \hline
Conformer \citep{Conformer} & Image recognition \& object detection, classification & 2021 & Encoder & Not mentioned in paper & N/A & LibriSpeech\\
 \hline
LoFTR \citep{LoFTR} & Image feature matching \& visual localization & 2021 & Encoder \& Decoder & No & NA & MegaDepth, ScanNet HPatches, ScanNet, MegaDepth, VisLoc benchmark (the Aachen-Day-Night, InLoc)\\
 \hline
CMT \citep{CMT} & Image recognition, detection \& segmentation & 2022 & Encoder & No & NA & ImageNet, CIFAR10, CIFAR100, Flowers, Standford cars, Oxford-IIIT pets, COCO val2017 \\
 \hline
Transformer in Transformer-TNT \citep{TnT} & Image recognition & 2021 & Encoder \& Decoder & Yes & ImageNet ILSVRC 2012 & COCO2017, ADE20K, Oxford 102 Flowers, Oxford-IIIT Pets, iNaturalist 2019, CIFAR-10, CIFAR-100 \\
 \hline
SWIN \citep{Swin_Transformer} & Object detection and segmentation & 2021 & Encoder & Yes & ImageNet-22k & ImageNet-1k, COCO 2017, ADE20K \\
 \hline
DETR \citep{DETR} & Object detection \& prediction & 2020 & Encoder \& Decoder & Yes & ImageNet pretrained backbone ResNet-50 & COCO 2017, panoptic segmentation datasets \\
 \hline
HOTR \citep{HOTR} & Human-object interaction detection & 2021 & Encoder \& Decoder & Yes & MS-COCO & V-COCO HICO-DET \\
\hline
\caption{\label{tab:image recognition}Transformer models for natural image processing - image recognition \& object detection}
\end{longtable}
\par 

\begin{itemize}[leftmargin=.1in]

\item \textbf{ViT:}
The ViT (Vision Transformer) is one of the earliest transformer-based models that has been applied to computer vision. ViT views an image as a sequence of patches and processes it using only the encoder module of the Transformer. ViT performs very well for classification tasks and can also be applied to image recognition tasks. It demonstrates that a transformer-based model can serve as an alternative to convolutional neural networks \citep{Vit}.

\item \textbf{TNT: }
Transformer in Transformers (TNT) is a transformer-based computer vision model that uses a transformer model inside another transformer model to capture features inside local patches of an image \citep{TnT}. The image is divided into local patches, which are further divided into smaller patches to capture more detailed information through attention mechanisms. TNT shows promising results in visual recognition tasks and offers an alternative to convolutional neural networks for computer vision tasks.

\item \textbf{LoFTR:}
LoFTR, which stands for Local Feature Matching with Transformer, is a computer vision model that is capable of learning feature representations directly from raw images, as opposed to relying on hand-crafted feature detectors for feature matching. This model employs both the encoder and decoder modules of the transformer. The encoder takes features from the image, while the decoder works to create a feature map. By leveraging the transformer's ability to capture global context and long-range dependencies, LoFTR can achieve high performance in visual recognition tasks.

\item \textbf{DETR:}
The Detection Transformer (DETR) represents a new approach to object detection or recognition, which performs the object detection task as a direct set of prediction problems \citep{DETR}. In contrast, other models accomplish this task in two stages. DETR uses an encoder to generate object queries, a self-attention mechanism to capture the relationship between the queries and objects in the image, and creates an object detection scheme. This model has been shown to be effective for object detection and recognition tasks and represents a significant advancement in the field.

\item \textbf{HOTR:}
The HOTR model, which stands for Human-Object Interaction Transformer, is a Transformer-based model designed for predicting Human-Object Interaction. It is the first Transformer-based Human-Object Interaction (HOI) detection prediction model that employs both the encoder and decoder modules of the Transformer. Unlike conventional hand-crafted post-processing schemes, HOTR uses a prediction set to extract the semantic relationship of the image, making it one of the fastest human-object interaction detection models available \citep{HOTR}.\\

\textbf{CMT}, \textbf{Conformer} \& \textbf{SWIN Transformer} 
The CMT and SWIN Transformer model have already been described in the Image Segmentation and Image Classification sections, respectively. Both of these models are also capable of performing the task of Image Segmentation. Additionally, the Conformer model was described in the Image Classification section.
\end{itemize}

\subsubsection{ Image Segmentation} 
Segmentation is the process of partitioning an image based on objects and creating boundaries between them, requiring pixel-level information extraction. There are two popular types of image segmentation tasks in computer vision: (i) Semantic Segmentation, which aims to identify and color similar objects belonging to the same class among all other objects in an image, and (ii) Instance Segmentation, which aims to detect instances of objects and their boundaries \citep{image_segmentation1,image_segmentation2}. In this section, we will discuss some Transformer-based models that have shown exceptional performance in image segmentation tasks (refer to Table \ref{tab:Image segmentation} for more details).

\medskip
\setlength\LTleft{-0.87cm}
\setlength\LTright{\LTleft}
\begin{longtable}{| M{2cm} | M{2.5cm}| M{.7cm} |M{1.8cm} |M{1.6cm} |M{3cm} |M{4.4cm} |}

 \hline
\textbf{Transformer Models} & \textbf{Task Accomplished} & \textbf{Year} & \textbf{Architecture (Encoder/ Decoder)} & \textbf{Pre-trained (Yes/NO)} & \textbf{Pre-training Dataset} & \textbf{Dataset (Fine-tuning, Training, Testing)} \\
 \hline
 \endfirsthead
 
 \multicolumn{7}{c}%
{{\bfseries \tablename\ \thetable{} -- continued from previous page}} \\
 \hline
\textbf{Transformer Models} & \textbf{Task Accomplished} & \textbf{Year} & \textbf{Architecture (Encoder/ Decoder)} & \textbf{Pre-trained (Yes/NO)} & \textbf{Pre-training Dataset} & \textbf{Dataset(Fine-tuning, Training, Testing)} \\
 \hline
\endhead

\hline \multicolumn{7}{|r|}{{Continued on next page}} \\ \hline
\endfoot

\endlastfoot

SWIN \citep{Swin_Transformer} & Object detection and segmentation & 2021 & Encoder & Yes & ImageNet-22k & ImageNet-1k, COCO 2017, ADE20K \\
 \hline
CMT \citep{CMT} & Image recognition, detection \& segmentation & 2022 & Encoder & No & NA & ImageNet, CIFAR10, CIFAR100, Flowers, Standford cars, Oxford-IIIT pets, COCO val2017 \\
 \hline
SETR \citep{SETR} & Image segmentation & 2020 & Encoder \& Decoder & Yes & ImageNet-1k, pre-trained weights provided by ViT or DeiT & ADE20K, Pascal Context, CityScapes\\
 \hline
IBOT \citep{iBOT} & Image classification, segmentation, object detection \& recognition & 2022 & Encoder & Yes & ImageNet-1K, ViT-L/16 ImageNet-22K & COCO, ADE20K\\
 \hline
\caption{\label{tab:Image segmentation}Transformer models for natural image processing - image segmentation}
\end{longtable}
\begin{itemize}[leftmargin=.1in]

\item \textbf{SWIN Transformer:}
The SWIN Transformer \citep{Swin_Transformer}, short for Scaled WINdowed Transformer, is a transformer-based model that is capable of handling large images by dividing them into small patches, or windows, and processing them through its architecture. By using shifted windows, the model requires a smaller number of parameters and less computational power, making it useful for real-life image applications. SWIN Transformer can perform image classification, segmentation, and object detection tasks with exceptional accuracy and efficiency \citep{eswa/ZidanGA23,eswa/Yang23}.

\item \textbf{CMT: }
CNNs Meet Transformer is a model that combines both Convolutional Neural Networks (CNN) and Vision Transformer (ViT). CNNs are better suited to capturing local features, while Transformers excel at capturing global context. CMT takes advantage of the strengths of both these models and performs well in image classification tasks as well as object detection and recognition tasks. The integration of CNN and Transformer allows CMT to handle both spatial and sequential data effectively, making it a powerful tool for computer vision tasks \citep{CMT}.

\item \textbf{SETR:}
SETR stands for SEgmentation TRansformer, which is a transformer-based model used for image segmentation tasks. It uses sequence-to-sequence prediction methods and removes the dependency of fully convolutional network with vanilla Transformer architecture. Before feeding the image into the Transformer architecture, it divides the image into a sequence of patches and the flattened pixel of each patch. There are three variants of SETR models available with different model sizes and performance levels \citep{SETR}.

\textbf{IBOT} 
The IBOT model, described above in the Image Classification section, is also capable of performing the Image Classification task.

\end{itemize}

\subsubsection{ Image Generation}

Image generation is a challenging task in computer vision, and transformer-based models have shown promising results in this area due to their parallel computational capability. This task involves generating new images using existing image pixels as input. It can be used for object reconstruction and data augmentation \citep{image_generation1,image_generation2}. While several text-to-image generation models exist, we focus on image generation models that use image pixels without any other type of data. In Table \ref{tab:image generation}, we discuss some transformer-based models that have demonstrated exceptional performance in image generation tasks.

\medskip
\setlength\LTleft{-0.86cm}
\setlength\LTright{\LTleft}
\begin{longtable}{| M{3cm} | M{2.5cm}| M{.7cm} |M{1.8cm} |M{1.6cm} |M{2cm} |M{4.4cm} |}

 \hline
\textbf{Transformer Models} & \textbf{Task Accomplished} & \textbf{Year} & \textbf{Architecture (Encoder/ Decoder)} & \textbf{Pre-trained (Yes/NO)} & \textbf{Pre-training Dataset} & \textbf{Dataset (Fine-tuning, Training, Testing)} \\
 \hline
 \endfirsthead
 
 \multicolumn{7}{c}%
{{\bfseries \tablename\ \thetable{} -- continued from previous page}} \\
 \hline
\textbf{Transformer Models} & \textbf{Task Accomplished} & \textbf{Year} & \textbf{Architecture (Encoder/ Decoder)} & \textbf{Pre-trained (Yes/NO)} & \textbf{Pre-training Dataset} & \textbf{Dataset(Fine-tuning, Training, Testing)} \\
 \hline
\endhead

\hline \multicolumn{7}{|r|}{{Continued on next page}} \\ \hline
\endfoot

\endlastfoot

Image Transformer \citep{Image_transfomer} & Image Generation & 2018 & Encoder \& Decoder & Not mentioned & N/A & ImageNet, CIFAR-10, CelebA \\
 \hline
I-GPT \citep{I-GPT} & Image Generation & 2020 & Decoder & Yes & BooksCorpus dataset, 1B
Word Benchmark & SNLI, MultiNLI, Question NLI, RTE, SciTail, RACE, Story Cloze, MSR Paraphrase Corpus, Quora Question Pairs, STS Benchmark, Stanford Sentiment Treebank-2, CoLA \\
 \hline
VideoGPT \citep{VideoGPT} & Video Generation & 2021 & Decoder & No & NA & BAIR RobotNet, Moving MNIST, ViZDoom, UCF-101, Tumblr GIF\\
 \hline
\caption{\label{tab:image generation}Transformer models for natural image processing - image generation}
\end{longtable}

\begin{itemize}[leftmargin=.1in]

\item \textbf{Image Transformer:}
Image Transformer is an autoregressive sequence generative model that uses the self-attention mechanism for image generation. This model generates new pixels and increases the size of the image by utilizing the attention mechanism on local pixels. It uses both the encoder and decoder module of the transformer, but does not use masking in the encoder. The encoder layer is used less than the decoder for better performance on image generation. Image Transformer is a remarkable model in the field of image generation \citep{Image_transfomer}.

\item \textbf{I-GPT:}
I-GPT or Image GPT is an image generative model that utilizes the GPT-2 model for training to auto-regressively predict pixels by learning image representation, without using the 2D image. BERT motifs can also be used during pre-training. I-GPT has four variants based on the number of parameters: IGPT-S (76M parameters), IGPT-M (455M parameters), IGPT-L (1.4B parameters), and IGPT-XL (6.8M parameters), where models with higher parameters have more validation losses \citep{I-GPT}.

\item\textbf{VideoGPT:}
VideoGPT is a generative model that combines two classes of architecture: likelihood-based models and VAE (Vector Quantized Variational Autoencoder). The aim of this combination is to create a model that is easy to maintain and use, as well as resource-efficient, while also being able to encode spatio-temporal correlations in video frames. VideoGPT has shown remarkable results compared to other models, particularly in tests conducted on the ``BAIR Robot Pushing" dataset \citep{VideoGPT}.

\end{itemize}
\subsection{Medical Image processing}


The diagnosis of pathologies based on medical images is often criticized as complicated, time-consuming, error-prone, and subjective \citep{lopez2020medical}. To overcome these challenges, alternative solutions such as deep learning approaches have been explored. Deep learning has made great progress in many other applications, such as Natural Language Processing and Computer Vision. Although transformers have been successfully applied in various domains, their application to medical images is still relatively new. Other deep learning approaches such as Convolutional Neural Networks (CNN), Recurrent Neural Networks (RNN), and Generative Adversarial Networks (GAN) are commonly used. This survey aims to provide a comprehensive overview of the various transformer models developed for processing medical images.

\subsubsection{Medical Image Segmentation}


Image segmentation refers to the task of grouping parts of the image that belong to the same category. In general, encoder-decoder architectures are commonly used for image segmentation \citep{lopez2020medical}. In some cases, image segmentation is performed upstream of the classification task to improve the accuracy of the classification results \citep{wang2022net}. The most frequently used loss functions in image segmentation are Pixel-wise cross-entropy loss and Dice Loss \citep{lopez2020medical}. Common applications of medical image segmentation include detecting lesions, identifying cancer as benign or malignant, and predicting disease risk. This paper presents a comprehensive overview of relevant models used in medical image segmentation.  Table \ref{tab:segmentation} provides a summary of these models.

\medskip

\setlength\LTleft{-0.86cm}
\setlength\LTright{\LTleft}
\begin{longtable}{| M{2.2cm} | M{2.5cm}| M{.7cm} |M{2cm} |M{1.5cm} |M{2.5cm} |M{4.4cm} |}

 \hline
\textbf{Transformer Name} & \textbf{Field of application} & \textbf{Year} & \textbf{Fully Transformer Architecture} & \textbf{Image type} & \textbf{Transformer Task} & \textbf{Dataset} \\
 \hline
 \endfirsthead
 
 \multicolumn{7}{c}%
{{\bfseries \tablename\ \thetable{} -- continued from previous page}} \\
 \hline
\textbf{Transformer Name} & \textbf{Field of application} & \textbf{Year} & \textbf{Fully Transformer Architecture} & \textbf{Image type} & \textbf{Transformer Task} & \textbf{Dataset}  \\
 \hline
\endhead

\hline \multicolumn{7}{|r|}{{Continued on next page}} \\ \hline
\endfoot

\endlastfoot

FTN \citep{DBLP:journals/mia/HeTBZZL22} & Skin lesion & 2022 & YES  & 2D & Image segmentation / classification & ISIC 2018 dataset \\
 \hline
RAT-Net \citep{DBLP:conf/midl/ZhuHWYLOX22} & Oncology (breast cancer) & 2022 & NO & 3D ultrasound & Image segmentation & a dataset of 256 subjects(330 Automatic Breast Ultrasound images for each patient)\\
 \hline
nnFormer \citep{DBLP:journals/corr/abs-2109-03201} & Brain tumor multi-organ cardiac diagnosis & 2022 & YES & 3D & Image segmentation & Medical Segmentation Decathlon (MSD), Synapse multiorgan segmentation,
Automatic Cardiac Diagnosis Challenge (ACDC)\\
 \hline
 TransConver \citep{liang2022transconver} &
Brain tumor &
2022 &
NO &
2D/3D &
Image Segmentation &
MICCAI BraTS2019, 
MICCAI BraTS2018\\
 \hline
 SwinBTS \citep{jiang2022swinbts} &
Brain tumor &
2022 &
NO &
3D &
Image Segmentation &
BraTS 2019, BraTS 2020, BraTS 2021\\
 \hline
 MTPA\_Unet \citep{DBLP:journals/sensors/JiangLCLZD22} &
Retinal vessel &
2022 &
NO &
2D &
Image segmentation &
DRIVE, CHASE DB1, and STARE Datasets\\
 \hline
 Dilated Transformer \citep{shen2022dilated} &
Oncology (Breast Cancer) &
2022 & 
NO & 
2D ultrasound &
Image segmentation &
2 small breast ultrasound image datasets\\
 \hline
 TFNet \citep{DBLP:conf/acpr/WangLK21} &
Oncology (Breast lesion) &
2022 &
NO &
2D ultrasound &
Image Segmentation &
BUSI Dataset
DDTI Dataset\\
 \hline
 Chest L-Transformer \citep{gu2022chest} &
Chest radiograph / Thoracic diseases &
2022 &
NO &
2D &
Image Classification / Segmentation &
SIIM-ACR Pneumothorax Segmentation
dataset contains 12,047\\
 \hline
\caption{\label{tab:segmentation}Transformer models for medical image segmentation}
\end{longtable}

\medskip

\begin{itemize}[leftmargin=.1in]

\item \textbf{FTN: }
FTN is a transformer-based architecture developed specifically for segmenting and classifying 2D images of skin lesions. It comprises of 5 layers, where each layer has a tokenization module known as SWT "Sliding Window Tokenization" and a transformer module. The model is segregated into encoders and decoders for the segmentation task, while only an encoder is needed for classification tasks. To improve computational efficiency and storage optimization, MSPA "Multi-head Spatial Pyramid Attention" is utilized in the "transformer" module instead of the traditional multi-head attention (MHA). In comparison to CNN, FTN has demonstrated superior performance on 10,025 images extracted from the publicly available ISIC 2018 dataset \citep{DBLP:journals/mia/HeTBZZL22}.

\medskip
\item \textbf{RAT-Net: }
 The primary objective of RAT-Net (Region Aware Transformer Network) is to replace the laborious and time-consuming manual task of detecting lesion contours in 3D ABUS (Automatic Breast Ultrasound) images. Compared to other state-of-the-art models proposed for medical image segmentation, RAT-Net has shown excellent performance. It is based on the SegFormer Transformer model, which is used to encode input images and determine the regions that are more relevant for lesion segmentation \citep{DBLP:conf/midl/ZhuHWYLOX22}.

\medskip
\item \textbf{nnFormer: } 
 is a model that uses Transformer architecture for segmentation of 3D medical images. The experiments were performed on 484 brain tumor images, 30 multi-organ scans, and 100 cardiac diagnosis images. Instead of using the conventional attention mechanism, nnFormer introduced LV-MSA ``Volume-based Multi-head Self-attention" and GV-MSA "Global Volume-based Multi-head Self-attention" to reduce the computational complexity. Additionally, nnFormer employs multiple convolution layers with small kernels in the encoder instead of large convolution kernels as in other visual transformers \citep{DBLP:journals/corr/abs-2109-03201}.

\medskip
\item \textbf{TransConver: } 
It combines CNN and SWIN transformers in parallel to extract global and local features simultaneously is proposed. The transformer block employs a cross-attention mechanism to merge semantically different global and local features. The network is designed to process both 2D and 3D brain tumor images and is trained on 335 cases from the training dataset of MICCAI BraTS2019. It is evaluated on 66 cases from MICCAI BraTS2018 and 125 cases from MICCAI BraTS2019 \citep{liang2022transconver}.

\medskip
\item \textbf{SwinBTS: }
 is a recently developed model that addresses the segmentation of 3D medical images by combining the Swin Transformer with CNN. It adopts an encoder-decoder architecture that applies the Swin Transformer to both the encoder and decoder. In addition, SwinBTS incorporates an advanced feature extraction module called ETrans (Enhanced Transformer) that follows the transformer approach and leverages convolution techniques \citep{jiang2022swinbts}.

\medskip
\item \textbf{MTPA\_Unet: } 
 (Multi-scale Transformer-Position Attention Unet) is a model that has been evaluated on several publicly recognized retinal datasets to enhance the performance of retinal image segmentation tasks. This model combines CNN and transformer architectures sequentially to accurately capture local and global image information. To capture long-term dependencies between pixels as well as contextual information about each pixel location, this model employs TPA (Transformer Position Attention), which is a combination of MSA (Multi-headed Self-Attention) and "Position Attention Module". Additionally, to optimize the model's extraction ability, feature map inputs of different resolutions are implemented due to the detailed information contained in retinal images \citep{DBLP:journals/sensors/JiangLCLZD22}.

\medskip
\item \textbf{TFNet: } 
 TFNet, aims to segment 2D ultrasound images of breast lesions by combining CNN with a transformer architecture. To address the challenge of lesions with different scales and variable intensities, CNN is employed as a backbone to extract features from the images, resulting in 3 high-level features containing semantic information and 1 low-level feature. These high-level features are fused through a Transformer Fuse Module (TFM), while the low-level features are fused via skip connection. The transformer module includes two main parts: Vanilla Multi-Head Self-Attention to capture the long-range dependency between sequences and MultiHead Channel-Attention (MCA) to detect dependencies between channels. To enhance the model's performance, novel loss functions are introduced, resulting in superior segmentation performance compared to other models. This approach is evaluated on a range of ultrasound image datasets, demonstrating excellent segmentation results \citep{DBLP:conf/acpr/WangLK21}.

\medskip
\item \textbf{Dilated Transformer: } 
The  (DT) model has been developed for the segmentation of 2D ultrasound images from small datasets of breast cancer using transformer architecture. Standard transformer models require large pre-training datasets to generate high-quality segmentation results, but DT overcomes this challenge by implementing the "Residual Axial Attention" mechanism for segmenting images from small breast ultrasound datasets. This approach applies attention to a single axis, namely the height axis and the width axis, instead of the whole feature map, which saves time and enhances computation efficiency \citep{shen2022dilated}.

\medskip
\item \textbf{Chest L-Transformer: }
This model is developed for the segmentation and classification of chest radiograph images. It uses a combination of CNN and transformer architecture, where the CNN is used as a backbone to extract local features from the 2D images and the transformer block is applied to detect the location of lesions using attention mechanisms. By using transformers in chest radiograph images, the model focuses more on areas where the disease may be more likely to occur, as opposed to treating them similarly using CNN alone \citep{gu2022chest}.

\end{itemize}
\subsubsection{Medical Image Classification}


Image classification refers to the process of recognizing, extracting, and selecting different types of features from an image for classification using labels \citep{wang2020medical}. Features in an image can be categorized into three types: low-level features, mid-level features, and high-level features \citep{wang2020medical}. Deep learning networks are designed to extract high-level features. Common applications of medical image classification include the detection of lesions, the identification of cancers as benign or malignant, and the prediction of disease risk \citep{Anwar_Khan23,Jungiewicz23}.  Table~\ref{tab:classification}, provides an overview of several relevant examples of Transformers used in medical image classification.
\medskip
\setlength\LTleft{-0.86cm}
\setlength\LTright{\LTleft}
\begin{longtable}{| M{2.2cm} | M{2.5cm}| M{.7cm} |M{2cm} |M{1.5cm} |M{2.5cm} |M{4.4cm} |}

 \hline
\textbf{Transformer Name} & \textbf{Field of application} & \textbf{Year} & \textbf{Fully Transformer Architecture} & \textbf{Image type} & \textbf{Transformer Task} & \textbf{Dataset} \\
 \hline
 \endfirsthead
 
 \multicolumn{7}{c}%
{{\bfseries \tablename\ \thetable{} -- continued from previous page}} \\
 \hline
\textbf{Transformer Name} & \textbf{Field of application} & \textbf{Year} & \textbf{Fully Transformer Architecture} & \textbf{Image type} & \textbf{Transformer Task} & \textbf{Dataset}  \\
 \hline
\endhead

\endfoot

\endlastfoot
CCT-based Model \citep{DBLP:journals/sensors/IslamNGSAAH22} &
Malaria Disease &
2022 &
NO &
2D images &
Image Classification &
National Library of Medicine malaria dataset\\
 \hline
Chest L-Transformer \citep{gu2022chest} &
Chest radiograph / Thoracic diseases &
2022 &
NO &
2D &
Image Classification / Segmentation &
SIIM-ACR Pneumothorax Segmentation
dataset contains 12,047\\
 \hline
\caption{\label{tab:classification}Transformer models for medical image classification}
\end{longtable}

\begin{itemize}[leftmargin=.1in]

\item \textbf{CCT-based Model \citep{DBLP:journals/sensors/IslamNGSAAH22}:} 
The model presented in this work is designed for classifying red blood cell (RBC) images as containing malaria parasites or not, by using Compact Convolutional Transformers (CCTs). The model input consists of image patches generated through convolutional operations and preprocessed by reshaping them to a fixed size. Unlike other vision transformer models, this model performs classification using sequence pooling instead of class tokens.

Compared to other deep learning models such as CNN, this model shows good performance in classifying RBC images. This satisfactory result was achieved by implementing a transformer architecture, using GRAD-CAM techniques to validate the learning process, and fine-tuning hyperparameters.

\item \textbf{Chest L-Transformer \citep{gu2022chest}:} 
 is a model designed for segmenting and classifying chest radiograph images \citep{gu2022chest}. The model utilizes a CNN backbone to extract local features from 2D images and a transformer block to apply attention mechanisms for detecting lesion locations. By incorporating transformers into the chest radiograph image analysis, the model is better able to attend to areas where disease may be more likely to occur, as opposed to traditional CNNs which treat all areas similarly. 

\end{itemize}
\medskip
\subsubsection{Medical Image Translation}

The field of research that involves altering the context (or domain) of an image without changing its original content is gaining traction. One example of this involves applying cartoon-style effects to images to change their appearance \citep{DBLP:journals/tmm/PangLQC22}. Image-to-image translation is a promising technique that can be utilized to synthesize medical images from non-corrupted sources with less cost and time, and it is also helpful for preparing medical images for registration or segmentation. Some of the most popular deep learning models developed for this area include ``Pix2Pix" and ``cyclic-consistency generative adversarial network" (GAN) \citep{yan2022swin}.   Table~\ref{tab:translation} provides an overview of some relevant examples of "Transformers" designed for medical image-to-image translation.

\medskip
\setlength\LTleft{-0.86cm}
\setlength\LTright{\LTleft}
\begin{longtable}{| M{2cm} | M{2.5cm}| M{.7cm} |M{2cm} |M{1.5cm} |M{2.5cm} |M{4.4cm} |}

 \hline
\textbf{Transformer Name} & \textbf{Field of application} & \textbf{Year} & \textbf{Fully Transformer Architecture} & \textbf{Image type} & \textbf{Transformer Task} & \textbf{Dataset} \\
 \hline
 \endfirsthead
 
 \multicolumn{7}{c}%
{{\bfseries \tablename\ \thetable{} -- continued from previous page}} \\
 \hline
\textbf{Transformer Name} & \textbf{Field of application} & \textbf{Year} & \textbf{Fully Transformer Architecture} & \textbf{Image type} & \textbf{Transformer Task} & \textbf{Dataset}  \\
 \hline
\endhead

\hline \multicolumn{7}{|r|}{{Continued on next page}} \\ \hline
\endfoot

\endlastfoot
MMTrans \citep{yan2022swin} &
Magnetic resonance imaging (MRI) &
2022 &
NO &
2D &
Medical image-to-image translation &
BraTs2018,
fastMRI,
The clinical brain MRI dataset\\
 \hline
TransCBCT \citep{chen2022more} &
Oncology (prostate Cancer) &
2022 &
NO &
2D &
Image Translation &
91 patients with prostate cancer\\
 \hline
\caption{\label{tab:translation}Transformer models for medical image translation}
\end{longtable}

\medskip

\begin{itemize}[leftmargin=.1in]

\item \textbf{MMTrans \citep{yan2022swin}:} 
The MMtrans (Multi-Modal Medical Image Translation) model is proposed based on the GAN architecture and Swin transformer structure for performing medical image-to-image translation on Magnetic Resonance Imaging (MRI). Unlike other image-to-image translation frameworks, MMtrans utilizes the transformer to model long global dependencies to ensure accurate translation results. Moreover, MMtrans does not require images to be paired and pixel-aligned since it employs SWIN as a registration module adapted for paired and unpaired images, which makes it different from other architectures like Pix2Pix. The remaining modules of GAN use SwinIR as a generator module, and CNN as a discriminator module.

\medskip
\item \textbf{TransCBCT \citep{chen2022more}:} 
A new architecture called TransCBT is proposed for the purpose of performing accurate radiotherapy by improving the quality of 2D images, specifically cone-beam computed tomography (CBCT), and generating synthetic 2D images (sCT) without damaging their structures. TransCBT integrates pure-transformer modeling and convolution approaches to facilitate the extraction of global information and enhances performance by introducing the multi-head self-attention method (SW-MSA). Another model that can improve the quality of CT images reconstructed via sinograms is the CCTR \citep{shi2022dual}. In comparison to TransCBCT, CCTR experiments utilized a lung image database with 1010 patients, rather than the 91 patients used in TransCBCT.

\end{itemize}

\subsection{Multi-Modality}

The transformer has demonstrated its potential in multi-modality, which stems from the human ability to perceive and process information from various senses such as vision, hearing, and language. Multi-modality machine learning models are capable of processing and combining different types of data simultaneously. Natural language, vision, and speech are among the most common types of data handled by multi-modal models. Several popular tasks in multi-modality include visual question answering, classification and segmentation, visual captioning, commonsense reasoning, and text/image/video/speech generation. In this section, we present a selection of transformer-based multi-modal models for each of these tasks 
providing an overview of their key features and working methods.

\subsubsection{Visual Question Answering}

Visual question answering is a popular task that can be accomplished using multi-modal models. It involves combining NLP and computer vision to answer questions about an image or video. The goal is to understand the features of both textual and visual information and provide the correct answer. Typically, the models take an image or video and text as input and deliver text as output answers \citep{VQA,VQA2}. In this context, we have identified and discussed the significant transformer models for visual question-answering tasks in Table~\ref{tab:visual question answering}.

\medskip
 \setlength\LTleft{-0.86cm}
\setlength\LTright{\LTleft}
\begin{longtable}{| M{2cm}| M{1.5cm} | M{2.2cm}| M{.7cm} |M{1.8cm} |M{1.5cm} |M{2.5cm} |M{3.5cm} |}

 \hline
\textbf{Transformer Models} & \textbf{Processed Data type (i/o)} & \textbf{Task Accomplished} & \textbf{Year} & \textbf{Architecture (Encoder/ Decoder)} & \textbf{Pre-trained (Yes/NO)} & \textbf{Pre-training Dataset} & \textbf{Dataset (Fine-tuning, Training, Testing)} \\
 \hline
 \endfirsthead
 
 \multicolumn{8}{c}%
{{\bfseries \tablename\ \thetable{} -- continued from previous page}} \\
 \hline
\textbf{Transformer Models} & \textbf{Processed Data type (i/o)} & \textbf{Task Accomplished} & \textbf{Year} & \textbf{Architecture (Encoder/ Decoder)} & \textbf{Pre-trained (Yes/NO)} & \textbf{Pre-training Dataset} & \textbf{Dataset (Fine-tuning, Training, Testing)} \\
 \hline
\endhead

\hline \multicolumn{8}{|r|}{{Continued on next page}} \\ \hline
\endfoot
\endlastfoot
BERT-Verients \citep{Pixel-BERT,LXMERT,ViLBERT,VL-BERT,UNITER} & Text and Image & Question Answering, Common sense reasoning & 2019-2020 & Encoder & Yes & \textbf{Pixel-BERT:} MS-COCO, Visual Genome \ \textbf{LX-MERT:} MS COCO,Visual Genome,VQA v2.0,GQA,VG-QA \ \textbf{ViLBERT:} Visual Genome, COCO \ \textbf{VL-BERT:} Conceptual Captions, BooksCorpus, English Wikipedia \textbf{Uniter:} COCO, VG, CC, SBU & \textbf{Pixel-BERT:} VQA 2.0
NLVR2, Flickr30K MS-COCO \ \textbf{LX-MERT:} VQA,GQA,NLVR \ \textbf{ViLBERT:} Conceptual Captions, Flickr30k \ \textbf{VL-BERT:} VCR dataset, RefCOCO \textbf{Uniter:} COCO, Flickr30K, VG, CC, SBU  \\
\hline
VIOLET \citep{VIOLET} & Video and Text & Video Question Answering, Text-to-video retrieval, Visual-Text Matching & 2022 & Encoder & Yes & Conceptual Captions-3M, WebVid-2.5M, YT-Temporal-180M & MSRVTT, DiDeMo, YouCook2, LSMDC, TGIF-Action, TGITransition, TGIF-Frame, MSRVTT-MC, MSRVTT-QA, MSVD-QA, LSMDC-MC, LSMDC-FiB \\
\hline
GIT \citep{GIT} & Image and Text & Image Classification, Image/video captioning, Question answering & 2022 & Encoder \& Decoder & Yes & combination of COCO, SBU, CC3M, VG, GITL, ALT200M and CC12M & Karpathy split-COCO, Flickr30K, no caps, TextCaps, VizWiz-Captions, CUTE, TextOCR \\
\hline
SIMVLM \citep{SimVLM} & Image and Text & Visual Question answering, image captioning & 2022 & Encoder \& Decoder & Yes & ALIGN \& Colossal Clean Crawled Corpus (C4) datasets & SNLI-VE, SNLI, MNLI, Multi30k, 10\% ALIGN , CC-3M \\
 \hline
BLIP \citep{BLIP} & Image, Video and Text & Question Answering, Image Captioning, image-text retrieval & 2022 & Encoder \& Decoder & Yes & Bootstrapped dataset- COCO, VG, SBU, CC3M, CC12M, LAION & COCO, Flickr30K, NoCaps, MSRVTT \\
\hline
\caption{\label{tab:visual question answering}Transformer models for multi-modality - visual question answering task}
\end{longtable}
\begin{itemize}[leftmargin=.1in]

\item \textbf{BERT-Variants: } 
Following the successful application of BERT-based models in NLP and computer vision tasks, several BERT-based models have demonstrated significant improvements in multi-modal tasks, particularly in question answering and commonsense reasoning. Currently, there are two distinct types of BERT-based models available in the literature: (i) Single-Stream Models and (ii) Two-Stream Models.

Single-Stream Models, such as VL-BERT, Uniter, etc., encode both modalities (text and image) within the same module. In contrast, Two-Stream Models, such as VilBERT, LXMERT, etc., process text and image through separate modules. Both types of models have been shown to yield promising results in various multi-modal tasks.

\item \textbf{ViLBERT: } 
ViLBERT is a two-stream model that is trained on text-image pairs and then passed both of the modules through co-attention, which helps to detect the important features of both text and images \citep{ViLBERT}. VLBERT, on the other hand, is a single-stream model that is pre-trained and takes both the image and text embedding features as input, making this model simple yet powerful \citep{VL-BERT}. Uniter represents Universal Image-Text Representation, which is a large-scale pre-trained model completed through masking \citep{UNITER}. Pixel-BERT is built using a combination of convolutional neural network (CNN) to extract image pixels and an encoder to extract text tokens, while the BERT-based transformer works as the cross-modality module. To capture all the spatial information from the image, Pixel-BERT takes the whole image as input, whereas other models extract image features from the regions \citep{Pixel-BERT}. Finally, LXMERT stands for Learning Cross-Modality Encoder Representations from Transformers. It processes the image and text through two different modules and is built with three encoders. This pre-trained model follows masked modeling and cross-modality for pre-training, which captures better relationships between text and images \citep{LXMERT}.

\item \textbf{VIOLET: } 
VIOLET is a neural network based on the transformer architecture that is designed to associate video with text. It consists of three modules: (1) Video Swin Transformer (VT), (2) Language Embedder (LE), and (3) Cross-modal Transformer (CT). VIOLET differs from other models in that it processes the temporal-spatial information from the video, rather than just extracting static images. This model has been evaluated on 12 datasets and has shown outstanding performance in many downstream tasks, such as Text-To-Video Retrieval and Video Question Answering. Furthermore, the authors propose a new pre-training task called "Masked Visual Token Modeling," which is used to pre-train VIOLET. This approach is combined with two other pre-training approaches, Masked Language Modeling and Visual-Text Matching, to achieve state-of-the-art performance on various benchmarks \citep{VIOLET}.

\item \textbf{GIT: } 
The Generative Image-to-Text Transformer, or GIT, is a multi-modal model designed to generate textual descriptions from visual images. This model employs both the encoder and decoder modules of the transformer architecture, using the image for encoding and decoding the text. To train the model, a large dataset of images paired with textual descriptions is used, allowing GIT to generate textual descriptions for previously unseen images. This approach has shown promising results in generating high-quality textual descriptions of images \citep{GIT}.\\

\textbf{SimVLM} \& \textbf{BLIP: } 
Both SimVLM and BLIP are models that can perform the task of visual captioning, which involves generating textual descriptions of visual images. The highlights of these models can be found in that section.
\end{itemize}

\subsubsection{Classification \& Segmentation}

Multi-modal classification and segmentation are often considered related tasks that involve classifying or segmenting data based on multiple modalities, such as text, image/video, and speech. As segmentation often helps to classify the image, text, or speech. In multi-modal classification, the task is to classify data based on its similarity and features using multiple modalities. This can involve taking text, image/video, or speech as input and using all of these modalities to classify the data more accurately.
Similarly, in multi-modal segmentation, the task is to segment data based on its features and use multiple modalities to achieve a more accurate segmentation. Both of these tasks require a deep understanding of the different forms of data and how they can be used together to achieve better classification or segmentation performance \citep{multi-classification,multi-classification2,multi-segmentation,multi-segmentation2}.
In recent years, transformer models have shown promising results in multi-modal classification and segmentation tasks. In Table \ref{tab:multi-modal classification & segmentation}, we highlight some of the significant transformer models that have been developed for these tasks.

\medskip
 \setlength\LTleft{-0.86cm}
\setlength\LTright{\LTleft}
\begin{longtable}{| M{2cm}| M{1.5cm} | M{2.2cm}| M{.7cm} |M{1.8cm} |M{1.5cm} |M{2.5cm} |M{3.5cm} |}

 \hline
\textbf{Transformer Models} & \textbf{Processed Data type (i/o)} & \textbf{Task Accomplished} & \textbf{Year} & \textbf{Architecture (Encoder/ Decoder)} & \textbf{Pre-trained (Yes/NO)} & \textbf{Pre-training Dataset} & \textbf{Dataset (Fine-tuning, Training, Testing)} \\
 \hline
 \endfirsthead
 
 \multicolumn{8}{c}%
{{\bfseries \tablename\ \thetable{} -- continued from previous page}} \\
 \hline
\textbf{Transformer Models} & \textbf{Processed Data type (i/o)} & \textbf{Task Accomplished} & \textbf{Year} & \textbf{Architecture (Encoder/ Decoder)} & \textbf{Pre-trained (Yes/NO)} & \textbf{Pre-training Dataset} & \textbf{Dataset (Fine-tuning, Training, Testing)} \\
 \hline
\endhead

\hline \multicolumn{8}{|r|}{{Continued on next page}} \\ \hline
\endfoot

\endlastfoot
CLIP \citep{Clip} & Image and Text & image Classification & 2021 & Encoder & Yes & Pre-training dataset from internet for CLIP & ImageNet, ImageNet V2, ImageNet Rendition, ObjectNet, ImageNet Sketch, ImageNet Adversarial 30 datasets \\
\hline
VATT \citep{VATT} & Video, Audio and Text & Audio event classification, Image classification, Video action recognition, Text-To-Video retrieval & 2021 & Encoder & Yes & AudioSet, HowTo100M & UCF10,HMDB5, Kinetics-400, Kinetics-600,Moments in Time, ESC50, AudioSet, YouCook2, MSR-VTT, ImageNet \\
\hline
Unicoder-VL \citep{Unicoder-VL} & Image and Text & Object Classification, Visual-linguistic Matching, visual commonsense reasoning, image-text retrieval & 2020 & Encoder & Yes & Conceptual Captions-3M, SBU Captions & MSCOCO, Flickr30K \\
\hline
ViLT \citep{ViLT} & Image and Text & Visual Question Answering, Image text matching, Natural Language for Visual Reasoning & 2021 & Encoder & Yes & MS-COCO,Visual Genome, SBU Captions, Google Conceptual Captions & VQA 2.0, NLVR2, MSCOCO, Flickr30K \\
 \hline
MBT \citep{MBT} & Audio and Visual & Audio-visual classification & 2022 & Encoder & Yes & VGGSoun, Kinetics400 and AS-500K, VGGSound & Audioset-mini and VGGSound, Moments In Time, Kinetics \\
 \hline
ALIGN \citep{Align} & Image and Text & Visual Classification & 2021 & Encoder & Yes & ALIGN training data, 0\% randomly sampled ALIGN training data, and CC-3M & Conceptual
Captions-CC, Flickr30K, MSCOCO, ILSVRC-2012 \\
 \hline
Florence \citep{Florance} & Image and Text & Classification, image caption, visual action recognition, Text-visual \& visual-text retrieval & 2021 & Encoder & Yes & FLD-900M, ImageNet (Swin transformer \& CLIP) & Web-scale by data curation, UniCL, ImageNet, COCO, Kinetics-600, Flickr30k, MSCOCO, SR-VTT \\
 \hline
GIT \citep{GIT} & Image and Text & Image Classification, Image/video captioning, Question answering & 2022 & Encoder \& Decoder & Yes & combination of COCO, SBU, CC3M, VG, GITL, ALT200M and CC12M & Karpathy split-COCO, Flickr30K, no caps, TextCaps, VizWiz-Captions, CUTE, TextOCR \\
\hline
\caption{\label{tab:multi-modal classification & segmentation} Multi-modal Transformer models - classification \& segmentation tasks}
\end{longtable}

\begin{itemize}[leftmargin=.1in]

\item \textbf{CLIP:} 
CLIP (Constructive Language-Image Pre-Training) is a multi-modal model that is trained in a supervised way with text and image data. This model simultaneously trains both the text and image through an encoder and predicts proper batches of text and image. CLIP is capable of understanding the relationship between text and images, and can generate images based on input text as well as generate text based on input images. CLIP has been shown to perform well on several benchmark datasets and is considered a state-of-the-art model for multi-modal tasks involving text and images \citep{Clip}.
\item \textbf{VATT:} 
which stands for Video, Audio, Text Transformer, is a multi-modal model based on the traditional transformer architecture without convolution layers. It is inspired by BERT and ViT and is pre-trained on two datasets using the Drop Token approach to optimize the training process. VATT is evaluated on 10 datasets containing videos, audio data, and text speech for 4 downstream tasks: Video action recognition, audio event classification, image classification, and text-to-video retrieval \citep{VATT}.

\item \textbf{MBT: } 
Multimodal Bottleneck Transformer (MBT) is a transformer-based model designed for processing both audio and visual data. The MBT model utilizes a bottleneck structure to focus on essential information for processing through the transformer architecture, thereby reducing the amount of data processed and minimizing the risk of overfitting. The bottleneck structure reduces the size of the model, training time, and computational cost. MBT has shown promising results in various multimodal tasks, such as audio-visual speech recognition and audio-visual event detection \citep{MBT}.

\item \textbf{ALIGN: } 
it stands for Large-scale ImaGe and Noise-text embedding. This model is a large-scale model that uses vision-language representational learning with noisy text annotations. ALIGN is a pre-trained model which uses a dual-encoder and is trained on huge-sized noisy image-text pair datasets. The dataset scale is able to adjust for noise, eliminating the need for pre-processing. ALIGN uses the contrastive loss to train the model, considering both image-to-text and text-to-image classification losses \citep{Align}.

\item \textbf{Florence: } 
Florence is a visual-language representation model that is capable of handling multiple tasks. It is an encoder-based pre-trained model trained on web-scale image-text data, and it can handle high-resolution images. This model shows strong performance on classification tasks, as well as other tasks like object/action detection and question answering \citep{Florance}.

\textbf{Unicoder-VL}, \textbf{GIT} \& \textbf{ViLT: }
Unicoder-VL and ViLT models have been described in the Visual Commonsense Reasoning section. Both models can perform the Commonsense Reasoning task in addition to other tasks. However, the characteristics of GIT model can be found on the visual question-answering section.
\end{itemize}

\subsubsection{Visual Captioning}
Visual captioning is a multi-modal task that involves both computer vision and NLP. The task aims to generate a textual description of an image, which requires a deep understanding of the relationship between image features and text. The visual captioning process usually involves several steps, starting with image processing, followed by encoding the features into vectors that can be used by the NLP model. These encoded vectors are then decoded into text, typically through generative NLP models. Although it is a complex process, visual captioning has a wide range of applications  \citep{visual-captioning,visual-captioning2}. In this section, we discuss significant transformer models for visual captioning tasks  (see Table \ref{tab:visual captioning}).

\medskip
 \setlength\LTleft{-0.86cm}
\setlength\LTright{\LTleft}
\begin{longtable}{| M{2cm}| M{1.5cm} | M{2.2cm}| M{.7cm} |M{1.8cm} |M{1.5cm} |M{2.5cm} |M{3.5cm} |}

 \hline
\textbf{Transformer Models} & \textbf{Processed Data type (i/o)} & \textbf{Task Accomplished} & \textbf{Year} & \textbf{Architecture (Encoder/ Decoder)} & \textbf{Pre-trained (Yes/NO)} & \textbf{Pre-training Dataset} & \textbf{Dataset (Fine-tuning, Training, Testing)} \\
 \hline
 \endfirsthead
 
 \multicolumn{8}{c}%
{{\bfseries \tablename\ \thetable{} -- continued from previous page}} \\
 \hline
\textbf{Transformer Models} & \textbf{Processed Data type (i/o)} & \textbf{Task Accomplished} & \textbf{Year} & \textbf{Architecture (Encoder/ Decoder)} & \textbf{Pre-trained (Yes/NO)} & \textbf{Pre-training Dataset} & \textbf{Dataset (Fine-tuning, Training, Testing)} \\
 \hline
\endhead

\hline \multicolumn{8}{|r|}{{Continued on next page}} \\ \hline
\endfoot

\endlastfoot

BLIP \citep{BLIP} & Image, Video and Text & Image Captioning, Question Answering, image-text retrieval & 2022 & Encoder \& Decoder & Yes & Bootstrapped dataset- COCO, VG, SBU, CC3M, CC12M, LAION & COCO, Flickr30K, NoCaps, MSRVTT \\
\hline
SIMVLM \citep{SimVLM} & Image and Text & Image captioning, Visual Question answering & 2022 & Encoder \& Decoder & Yes & ALIGN \& Colossal Clean Crawled Corpus (C4) datasets & SNLI-VE, SNLI, MNLI, Multi30k, 10\% ALIGN , CC-3M \\
 \hline
Florence \citep{Florance} & Image and Text & Classification, image caption, visual action recognition, Text-visual \& visual-text retrieval & 2021 & Encoder & Yes & FLD-900M, ImageNet (Swin transformer \& CLIP) & Web-scale by data curation, UniCL, ImageNet, COCO, Kinetics-600, Flickr30k, MSCOCO, SR-VTT \\
 \hline
GIT \citep{GIT} & Image and Text & Image Classification, Image/video captioning, Question answering & 2022 & Encoder \& Decoder & Yes & combination of COCO, SBU, CC3M, VG, GITL, ALT200M and CC12M & Karpathy split-COCO, Flickr30K, no caps, TextCaps, VizWiz-Captions, CUTE, TextOCR \\
\hline
\caption{\label{tab:visual captioning}Multi-modal Transformer models - visual captioning task}
\end{longtable}

\begin{itemize}[leftmargin=.1in]

\item \textbf{BLIP:} 
Bootstrapping Language-Image Pre-training (BLIP) is a pre-trained model designed to enhance performance on various tasks through fine-tuning for specific tasks. This model utilizes a VLP (Vision and Language Pre-training) framework with an encoder-decoder module of the Transformer architecture, which uses noisy data with captions and is trained to remove noisy captions. BLIP is capable of performing a range of downstream tasks, including image captioning, question answering, image-text retrieval, and more \citep{BLIP}.

\item \textbf{SimVLM: } 
 short for SIMple Visual Language Model, SimVLM is a pre-trained model that uses weak supervision methods for training. This approach provides the model with greater flexibility and scalability. Instead of using pixel patch projection, this model uses the full image as patches and is trained with a language model. As a result of these methods, SimVLM is capable of performing various tasks, with question answering being one of its significant strengths \citep{SimVLM}.

\item \textbf{Florence: } 
Florence is a visual-language representation model that can perform multiple tasks. It is an encoder-based pre-trained model trained on web-scale image-text data, which enables it to handle high-resolution images. In addition to tasks such as object/action detection and question answering, Florence also shows strong performance in classification tasks \citep{Florance}.

\textbf{GIT: } 
The description of the model has already been provided in the Question Answering section above. These models can also perform the Question Answering task with high performance.

\end{itemize}
\subsubsection{Visual Commonsense Reasoning}

Visual commonsense reasoning is a challenging task that requires a model with a deep understanding of visualization and different images or videos containing objects and scenes, inspired by how humans see and visualize things. These models capture information from different sub-tasks like object recognition and feature extraction. This information is then transformed into a vector to be used for reasoning. The reasoning module understands the relationship between the objects in the image and the output of the inferencing step provides a prediction about the interaction and relationship between the objects. Visual commonsense reasoning helps to improve the performance of various tasks like classification, image captioning, and other deep understanding-related tasks \citep{visual-common,visual-common2}. In this section, we highlight and discuss significant transformer models for visual commonsense reasoning tasks that are summarized in Table \ref{tab:visual commonsense}.

\medskip
 \setlength\LTleft{-0.86cm}
\setlength\LTright{\LTleft}
\begin{longtable}{| M{2cm}| M{1.3cm} | M{2.0cm}| M{.7cm} |M{1.8cm} |M{1.2cm} |M{3.5cm} |M{3.3cm} |}

 \hline
\textbf{Transformer Models} & \textbf{Processed Data type(i/o)} & \textbf{Task Accomplished} & \textbf{Year} & \textbf{Architecture (Encoder/ Decoder)} & \textbf{Pre-trained (Yes/NO)} & \textbf{Pre-training Dataset} & \textbf{Dataset (Fine-tuning, Training, Testing)} \\
 \hline
 \endfirsthead
 
 \multicolumn{8}{c}%
{{\bfseries \tablename\ \thetable{} -- continued from previous page}} \\
 \hline
\textbf{Transformer Models} & \textbf{Processed Data type (i/o)} & \textbf{Task Accomplished} & \textbf{Year} & \textbf{Architecture (Encoder/ Decoder)} & \textbf{Pre-trained (Yes/NO)} & \textbf{Pre-training Dataset} & \textbf{Dataset(Fine-tuning, Training, Testing)} \\
 \hline
\endhead

\hline \multicolumn{8}{|r|}{{Continued on next page}} \\ \hline
\endfoot

\endlastfoot

BERT-Verients \citep{Pixel-BERT,LXMERT,ViLBERT,VL-BERT,UNITER} & Text and Image & Question Answering, Common sense reasoning & 2019-2020 & Encoder & Yes & \textbf{Pixel-BERT:} MS-COCO, Visual Genome \ \textbf{LX-MERT:} MS COCO,Visual Genome,VQA v2.0,GQA,VG-QA \ \textbf{ViLBERT:} Visual Genome, COCO \ \textbf{VL-BERT:} Conceptual Captions, BooksCorpus, English Wikipedia \textbf{Uniter:} COCO, VG, CC, SBU & \textbf{Pixel-BERT:} VQA 2.0
NLVR2, Flickr30K MS-COCO \ \textbf{LX-MERT:} VQA,GQA,NLVR \ \textbf{ViLBERT:} Conceptual Captions, Flickr30k \ \textbf{VL-BERT:} VCR dataset, RefCOCO \textbf{Uniter:} COCO, Flickr30K, VG, CC, SBU  \\
\hline
ViLT \citep{ViLT} & Image and Text & Visual Question Answering, Image text matching, Natural Language for Visual Reasoning & 2021 & Encoder & Yes & MS-COCO,Visual Genome, SBU Captions, Google Conceptual Captions & VQA 2.0, NLVR2, MSCOCO, Flickr30K \\
\hline
Unicode-VL \citep{Unicoder-VL} & Image and Text & Object Classification, Visual-linguistic Matching, visual commonsense reasoning, image-text retrieval & 2020 & Encoder & Yes & Conceptual Captions-3M, SBU Captions & MSCOCO, Flickr30K \\
\hline
\caption{\label{tab:visual commonsense} Multi-modal Transformer models - visual commonsense reasoning task}
\end{longtable}
\begin{itemize}[leftmargin=.1in]

\item \textbf{Unicoder-VL:} 
Unicoder-VL is a large-scale pre-trained encoder-based model that utilizes cross-modeling to build a strong understanding of the relationship between image and language. The model employs a masking scheme for pre-training on a large corpus of data. These methods enhance the model's performance on visual commonsense reasoning tasks in addition to visual classification tasks \citep{Unicoder-VL}

\item \textbf{ViLT (Vision-and-Language Transformer):} 
 is a multi-modal architecture based on the ViT (Vision Transformer) model, utilizing a free-convolution approach. Unlike other VLP (Vision-and-Language Pre-training) models, ViLT performs data augmentation during the execution of downstream tasks of classification and retrievals, which improves the model's performance. Inspired by Pixel-BERT, ViLT takes the entire image as input instead of just using selected regions. By omitting convolutional visual embedders, ViLT reduces the model size and achieves remarkable performance compared to other VLP models \citep{ViLT}.
 
\textbf{BERT-Variants: }  
The BERT-Variants models have been previously described in the Classification \& segmentation section. It should be noted that these models are also capable of performing the Classification \& segmentation task.
\end{itemize}
\subsubsection{Image/Video/Speech Generation}

Multi-modal generation tasks have gained a lot of attention in the field of artificial intelligence. These tasks involve generating images, text, or speech from inputs of different modalities of input. In recent times, several generative models have demonstrated outstanding performance, making this field of research even more attractive \citep{multi-generative}. In this section, we discuss some significant transformer models that have been used for multi-modal generation tasks. These models are summarized in Table \ref{tab: multi-modal generation}.


\medskip

 \setlength\LTleft{-0.86cm}
\setlength\LTright{\LTleft}
\begin{longtable}{| M{2cm}| M{1.3cm} | M{2.0cm}| M{.7cm} |M{1.8cm} |M{1.2cm} |M{2.2cm} |M{4.3cm} |}

 \hline
\textbf{Transformer Models} & \textbf{Processed Data type(i/o)} & \textbf{Task Accomplished} & \textbf{Year} & \textbf{Architecture (Encoder/ Decoder)} & \textbf{Pre-trained (Yes/NO)} & \textbf{Pre-training Dataset} & \textbf{Dataset (Fine-tuning, Training, Testing)} \\
 \hline
 \endfirsthead
 
 \multicolumn{8}{c}%
{{\bfseries \tablename\ \thetable{} -- continued from previous page}} \\
 \hline
\textbf{Transformer Models} & \textbf{Processed Data type (i/o)} & \textbf{Task Accomplished} & \textbf{Year} & \textbf{Architecture (Encoder/ Decoder)} & \textbf{Pre-trained (Yes/NO)} & \textbf{Pre-training Dataset} & \textbf{Dataset (Fine-tuning, Training, Testing)} \\
 \hline
\endhead

\hline \multicolumn{8}{|r|}{{Continued on next page}} \\ \hline
\endfoot

\endlastfoot

DALL-E \citep{DALL-E} & Image and Text & Image Generation from text & 2021 & Encoder \& Decoder & Yes & Not disclosed & Conceptual Captions (MS-COCO extension), Wikipedia (text-image pairs), and YFCC100M(filtered subset) \\
 \hline
GLIDE \citep{Glide} & Image and Text & Image generation \& edit from text & 2021 & Encoder \& Decoder & No & NA & MS-COCO, ViT-B CLIP (noised), CLIP and DALL-E filtered datasets \\
\hline
Chimera \citep{Chimera} & Audio and Text & Text generation from speech & 2021 & Encoder \& Decoder & Yes & ALIGN \& MT-Dataset & MuST-C, Augmented LibriSpeech Dataset (En-Fr), Machine Translation Datasets (WMT 2016, WMT 2014, OPUS100, OpenSubtitles) \\
 \hline
CogView \citep{Cogview} & Image and Text & Classification, Image generation from text & 2021 & Decoder & Yes & VQ-VAE & MS COCO, Wudao Corpora-extension dataset. \\
\hline
\caption{\label{tab: multi-modal generation} Multi-modal Transformer models  - image/video/speech generation task}
\end{longtable}

\begin{itemize}[leftmargin=.1in]

\item \textbf{DALL-E: } 
DALL-E \citep{DALL-E} is a popular transformer-based model for generating images from text. It is trained on a large and diverse dataset of both text and images, utilizing 12 billion parameters from the GPT-3 architecture. To reduce memory consumption, DALL-E compresses images without compromising their visual quality. An updated version of DALL-E, known as DALL-E 2, has been introduced with a higher number of parameters (175 billion) which allows for the generation of higher resolution images. Additionally, DALL-E 2 is capable of generating a wider range of images.

\item\textbf{Chimera: } 
The Chimera end-to-end architecture is designed for Speech-And-Text-Translation (ST). This architecture draws inspiration from Text-Machine-Translation and proposes a new module called the Shared Semantic Projection Module based on attention mechanisms. The objective of this module is to reduce latency and errors during the speech translation process by removing dissimilarities between speech and language modalities.

\item \textbf{CogView: } 
CogView \citep{Cogview} is an image generation model that generates images based on input text descriptions, making it a challenging task that requires a deep understanding of the contextual relationship between text and image. It utilizes a transformer-GPT-based architecture to encode the text into a vector and decode it into an image. This model outperforms DALL-E in some cases, which also generates images from text descriptions, but uses text-image pairs for training the model.

\item \textbf{GLIDE:} 
 short for Guided Language to Image Diffusion for Generation and Editing, GLIDE is a diffusion model that is distinct from conventional models in that it can both generate and edit images. Unlike other models, diffusion models are sequentially injected with random noise and trained to remove that noise to construct the original data. GLIDE takes textual information as input and generates an output image conditioned on that information. In some instances, the images generated by GLIDE are more impressive than those generated by DALL-E \citep{Glide}.
\end{itemize}
\subsubsection{Cloud computing}

Cloud computing is a crucial element of modern technology, particularly with regard to the Internet of Things (IoT). It encompasses a wide variety of cloud-based tasks, including server computing, task scheduling, storage, networking, and more. In wireless networks, cloud computing aims to improve scalability, flexibility, and adaptability, thereby providing seamless connectivity. To achieve this, data or information is retrieved from the network for computation, with various types of data being processed, including text, images, speech, and digits. Due to this multi-modal approach, cloud computing is classified in this category \citep{cloud-com,cloud-com1,cloud-com2}. In this article, we focus on the significant transformer models used in cloud computing tasks, which are presented in Table \ref{tab: cloud computing}.


\medskip

 \setlength\LTleft{-0.86cm}
\setlength\LTright{\LTleft}
\begin{longtable}{| M{2cm}| M{1.3cm} | M{2.0cm}| M{.7cm} |M{1.8cm} |M{1.2cm} |M{2.0cm} |M{4.5cm} |}

 \hline
\textbf{Transformer Models} & \textbf{Processed Data type(i/o)} & \textbf{Task Accomplished} & \textbf{Year} & \textbf{Architecture (Encoder/ Decoder)} & \textbf{Pre-trained (Yes/NO)} & \textbf{Pre-training Dataset} & \textbf{Dataset (Fine-tuning, Training, Testing)} \\
 \hline
 \endfirsthead
 
 \multicolumn{8}{c}%
{{\bfseries \tablename\ \thetable{} -- continued from previous page}} \\
 \hline
\textbf{Transformer Models} & \textbf{Processed Data type (i/o)} & \textbf{Task Accomplished} & \textbf{Year} & \textbf{Architecture (Encoder/ Decoder)} & \textbf{Pre-trained (Yes/NO)} & \textbf{Pre-training Dataset} & \textbf{Dataset (Fine-tuning, Training, Testing)} \\
 \hline
\endhead

\hline \multicolumn{8}{|r|}{{Continued on next page}} \\ \hline
\endfoot

\endlastfoot

VMD \& R-Transformer \citep{VMD-R-Transformer} & workload sequence & cloud workload forecasting & 2020 & Encoder \& Decoder & No & NA & Google cluster trace, Alibaba cluster trace \\
 \hline
ACT4JS \citep{ACT4JS} & Cloud jobs/task & Cloud computing resource job scheduling. & 2022 & Encoder & No & NA & Alibaba Cluster-V2018 \\
\hline
TEDGE-Catching \citep{TEDGE-Catching} & Sequential request pattern of the contents (Ex: video, image, websites, etc) & Predict the content popularity in proactive caching schemes. & 2021 & Encoder & No & NA & MovieLens \\
 \hline
SACCT \citep{SACCT} & Network status (Bandwidth, storage, and etc) & Optimize network based on network status. & 2021 & Encoder & No & NA & Not mentioned clearly \\
\hline
\caption{\label{tab: cloud computing}Multi-modal Transformer models - cloud computing task}
\end{longtable}

\begin{itemize}[leftmargin=.1in]

\item \textbf{VMD \& R-TRANSFORMER:} 
Workload forecasting is a critical task for the cloud, and previous research has focused on using recurrent neural networks (RNNs) for this purpose. However, due to the highly complex and dynamic nature of workloads, RNN-based models struggle to provide accurate forecasts because of the problem of vanishing gradients. In this context, the proposed Variational Mode Decomposition-VMD and R-Transformer model offers a more accurate solution by capturing long-term dependencies using multi-head attention and local non-linear relationships of workload sequences with local techniques \citep{VMD-R-Transformer}. Therefore, this model is capable of executing the workload forecasting task with greater precision than existing RNN-based models.

\item \textbf{TEDGE-Caching:} 
TEDGE is an acronym for Transformer-based Edge Caching, which is a critical component of the 6G wireless network as it provides a high-bandwidth, low-latency connection. Edge caching stores multimedia content to deliver it to users with low latency. To achieve this, it is essential to proactively predict popular content. However, conventional models are limited by long-term dependencies, computational complexity, and the inability to compute in parallel. In this context, the Transformer-based Edge (TEDGE) caching framework incorporates a vision transformer (ViT) to overcome these limitations, without requiring data pre-processing or additional contextual information to predict popular content at the Mobile Edge. This is the first model to apply a transformer-based approach to execute this task, resulting in superior performance \citep{TEDGE-Catching}.

\item \textbf{ACT4JS:} 
The Actor-Critic Transformer for Job Scheduling (ACT4JS) is a transformer-based model designed to allocate cloud computing resources to different tasks in cloud computing. The model consists of an Actor and Critic network, where the Actor-network selects the best action to take at each step, and the Critic network evaluates the action taken by the Actor-network and provides feedback to improve future steps. This approach allows for a better understanding of the complex relationship between cloud jobs and enables prediction or scheduling of jobs based on different features such as job priority, network conditions, resource availability, and more \citep{ACT4JS}.

\item \textbf{SACCT:} 
SACCT refers to the Soft Actor-Critic framework with Communication Transformer, which combines the transformer, reinforcement learning, and convex optimization techniques. This model introduces the Communication Transformer (CT), which works with reinforcement learning to adapt to different challenges, such as bandwidth limitations, storage constraints, and more, in the wireless edge network during live streaming. Adapting to changing network conditions is critical during live streaming, and SACCT provides a system that adjusts resources to improve the quality of service based on user demand and network conditions. The SACCT model's ability to adapt to changing conditions and optimize resources makes it an important contribution to the field of wireless edge network technology \citep{SACCT}.

\end{itemize}

\subsection{Audio \& Speech}

Audio and speech processing is one of the most essential tasks in the field of deep learning. Along with NLP, speech processing has also gained attention from researchers, leading to the application of deep neural network methods. As transformers have achieved great success in the field of NLP, researchers have also had significant success when applying transformer-based models to speech processing.

\subsubsection{Speech Recognition}

Speech Recognition is one of the most popular tasks in the field of artificial intelligence. It is the ability of a model to identify human speech and convert it into a textual or written format. This process is also known as speech-to-text, automatic speech recognition, or computer-assisted transcription. Speech recognition technology has advanced significantly in the last few years. It involves two types of models, namely the acoustic model and the language model. Several features contribute to the effectiveness of speech recognition models, including language weighting, speaker labeling, acoustics training, and profanity filtering. Despite significant advancements in speech recognition technology, there is still room for further improvement \citep{speech-recognition,speech-recognition1,speech-recognition2}. One promising development in this area is the use of transformer-based models, which have shown significant improvement in various stages of the speech recognition task. The majority of transformer-based audio/speech processing models have focused on speech recognition tasks, and among these, several models have made exceptional contributions, exhibited high levels of accuracy, introduced new effective ideas, or created buzz in the AI field. In Table \ref{tab:speech recognition}, we highlight  the most significant transformer models for speech recognition tasks.

\medskip
\setlength\LTleft{-.85cm}
\setlength\LTright{\LTleft}
\begin{longtable}{| M{2.6cm} | M{2.5cm}| M{.7cm} |M{1.8cm} |M{1.6cm} |M{3cm} |M{4cm} |}

 \hline
\textbf{Transformer Models} & \textbf{Task Accomplished} & \textbf{Year} & \textbf{Architecture (Encoder/ Decoder)} & \textbf{Pre-trained (Yes/NO)} & \textbf{Pre-training Dataset} & \textbf{Dataset (Fine-tuning, Training, Testing)} \\
 \hline
\endfirsthead
 


\hline
Conformer \citep{Conformer} & Speech Recognition & 2020 & Encoder \& Decoder & No & NA & Librispeech, test/testother. \\
\hline
Speech Transformer \citep{Speech-Transformer} & Speech Recognition & 2018 & Encoder \& Decoder & No & NA & Wall StreetJournal (WSJ), NVIDIA K80 G-PU \\
\hline
VQ-Wav2vec \citep{vq-wav2vec} & Speech Recognition & 2020 & Encoder & Yes & Librispeech & TIMIT, WSJ-Wall StreetJournal \\
 \hline
Wav2vec 2.0 \citep{wav2vec-2.0} & Speech Recognition & 2020 & Encoder & Yes & Librispeech, LibriVox & Librispeech, LibriVox, TIMIT \\
 \hline
HuBERT \citep{HuBERT} & Speech Recognition & 2021 & Encoder & Yes & Librispeech, Libri-light & Librispeech(train-clean), Libri-light(train-clean)\\
 \hline
BigSSL \citep{BigSSL} & Speech Recognition & 2022 & Encoder \& Decoder & Yes & wav2vec 2.0, YT-U, Libri-Light & YouTube, English(US) Voice search(VS), SpeechStew, LibriSpeech, CHiME6, Telephony\\
 \hline
Whisper \citep{Whisper} & Speech recognition, Translation, Language Identification & 2022 & Encoder \& Decoder & Yes & Not Mentioned & VoxLingua107, LibriSpeeech, CoVoST2, Fleurs, Kincaid46\\
 \hline
Transformer Transducer \citep{Transformer-Transducer} & Speech recognition & 2020 & Encoder & No & NA &  LibriSpeeech\\
 \hline
XLSR-Wav2Vec2 \citep{XLSR-Wav2Vec2} & Speech recognition & 2020 & Encoder & Yes & 53 languages datasets & CommonVoice, BABEL,Multilingual LibriSpeech(MLS)\\
 \hline
\caption{\label{tab:speech recognition}Transformer models for audio \& speech recognition task}
\end{longtable}

\begin{itemize}[leftmargin=.1in]

\item \textbf{Conformer:} 
The Conformer architecture is a model that combines the advantages of both the Transformer and convolutional neural network (CNN) for automatic speech recognition tasks. While the Transformer is proficient at capturing global features, and CNN excels at capturing local features, the Conformer architecture leverages the strengths of both to achieve superior performance. Recent studies have shown that the Conformer architecture outperforms both CNN and Transformer models individually, thereby setting a new state-of-the-art in automatic speech recognition performance \citep{r63}.

\item \textbf{Speech Transformer:} 
It is a speech recognition model published in 2018 which is one of the earliest Transformer inspired speech models. It eliminates the conventional Recurrent Neural Network (RNN)-based approach in speech processing and instead applies an attention mechanism. The introduction of the Transformer model into speech recognition has led to a number of benefits. For instance, the training time and memory usage become lower, allowing for better scalability. This is especially helpful for tasks that require a long-term dependency due to the elimination of recurrence sequence-to-sequence processing  \citep{Speech-Transformer}. 
\item \textbf{Wav2vec 2.0: } 
Wav2Vec 2.0 is a self-supervised model which uses discretization algorithms to capture the vocabulary from raw speech representation. The learned vocabulary is passed through an architecture consisting of multi-layer convolutional feature encoder. This encoder has multiple convolution layers, layer normalization and an activation function, with the audio representations being masked during the training process.  Wav2Vec 2.0 offers the advantage of performing well on speech recognition tasks with a small amount of supervised data \citep{wav2vec-2.0}.

\item \textbf{VQ-Wav2vec:} 
VQ-Wav2Vec is a Transformer based model that enables Vector Quantization-VQ tasks to be executed in a self-supervised way. It is built on the Wav2Vec model, which is an effective way of compressing continuous signals into discrete symbols. Unlike other models, VQ-Wav2Vec trains without the need of unlabeled data. Instead, corrupted speech is used, and the model learns by predicting the missing parts of the speech. This process of training has been proven to be highly effective, and the model is capable of achieving a higher accuracy when compared to other models \citep{vq-wav2vec}.  

\item \textbf{BigSSL: } 
BigSSL is a large-scale semi-supervised Transformer-based model designed specifically for speech recognition tasks. Obtaining labeled speech data is a challenging and time-consuming process, while the availability of unlabeled data is considerably vast. In this context, the BigSSL model proposes a novel approach to enhance the performance of speech recognition tasks. By leveraging a smaller portion of labeled data in conjunction with a substantial amount of unlabeled data, this model achieves improved performance. Furthermore, the utilization of a larger quantity of unlabeled data helps alleviate the overfitting issue, thereby further enhancing the overall performance of the BigSSL model \citep{BigSSL}.

\item \textbf{HuBERT: } 
HuBERT, short for Hidden-Unit BERT, is a self-supervised speech representation model. Its approach involves offline clustering for feature representation, with the loss calculation restricted to the masked regions. This emphasis allows the model to effectively learn a combination of acoustic and language models over the input data. HuBERT consists of a convolutional waveform encoder, a projection layer, a BERT encoder, and a code embedding layer. The CNN component generates feature representations, which are then subjected to random masking. These masked representations are subsequently passed through the BERT encoder, yielding another set of feature representations. HuBERT's functioning resembles that of a mask language model and has demonstrated notable performance in speech representation tasks, particularly speech recognition \citep{HuBERT}.

\item \textbf{Transformer Transducer: } 
It is a speech recognition model that capitalizes on the strengths of both the self-attention mechanism of the Transformer and the recurrent neural network (RNN). This model is constructed by integrating the encoder module of the transformer with the RNN-T loss function. The encoder module is responsible for extracting speech representations, while the RNN-T component utilizes this information to make real-time predictions of the transcript, facilitating a swift response—an essential requirement for speech recognition tasks (Transformer-Transducer \citep{Transformer-Transducer}.

\item \textbf{Whisper:} 
It is a noteworthy speech recognition model that emerged in late 2022, specifically designed to address the challenging task of recognizing speech with low volume. The uniqueness of this model lies in its dedicated efforts to improve low-volume speech recognition. Whisper adopts a training approach that incorporates a lower level of speech data and leverages weak supervision methods, enabling training on a larger corpus of data. This strategic approach has proven instrumental in enhancing the performance of the Whisper model, enabling it to effectively capture and comprehend low-level speech phenomena \citep{Whisper}.

\item \textbf{XLSR-Wav2Vec2:} 
This model demonstrates the capability to recognize speech across multiple languages, eliminating the need for extensive labeled data in each language for training. By learning the relationships and shared characteristics among different languages, this model surpasses the requirement of training on specific language-labeled speech data. Consequently, the XLSR-Wav2Vec2 model offers an efficient solution for multiple-language speech recognition, requiring significantly less data for training while adhering to the architectural principles of Wav2Vec2 \citep{XLSR-Wav2Vec2}.

\end{itemize}

\subsubsection{Speech Separation}
It poses a considerable challenge within the field of audio signal processing. It involves the task of separating the desired speech signal, which may include various sources such as different speakers or human voices, from additional sounds such as background noise or interfering sources. In the domain of speech separation, three commonly employed methods are followed: (i) Blind source separation, (ii) Beamforming, and (iii) Single-channel speech separation.
The significance of speech separation has grown with the increasing popularity of automatic speech recognition (ASR) systems. It is often employed as a preprocessing step for speech recognition tasks. The accurate distinction between the desired speech signal and unwanted noise is crucial to ensure precise speech recognition results. Failure to properly segregate the desired speech from interfering noise can lead to erroneous speech recognition outcomes  \citep{speech-separation,speech-separation1}. 
In this context, we present several Transformer-based models that have showcased noteworthy advancements in audio and speech separation tasks. The details of these models are presented in Table  \ref{tab:speech separation}.

\medskip
\setlength\LTleft{-.85cm}
\setlength\LTright{\LTleft}
\begin{longtable}{| M{2.4cm} | M{2.5cm}| M{.7cm} |M{1.8cm} |M{1.6cm} |M{3cm} |M{4cm} |}

 \hline
\textbf{Transformer Models} & \textbf{Task Accomplished} & \textbf{Year} & \textbf{Architecture (Encoder/ Decoder)} & \textbf{Pre-trained (Yes/NO)} & \textbf{Pre-training Dataset} & \textbf{Dataset (Fine-tuning, Training, Testing)} \\
 \hline
 \endfirsthead
 
 \multicolumn{7}{c}%
{{\bfseries \tablename\ \thetable{} -- continued from previous page}} \\
 \hline
\textbf{Transformer Models} & \textbf{Task Accomplished} & \textbf{Year} & \textbf{Architecture (Encoder/ Decoder)} & \textbf{Pre-trained (Yes/NO)} & \textbf{Pre-training Dataset} & \textbf{Dataset (Fine-tuning, Training, Testing)} \\
 \hline
\endhead

\hline \multicolumn{7}{|r|}{{Continued on next page}} \\ \hline
\endfoot

\endlastfoot
\hline
DPTNeT \citep{DPTNeT} & Speech Separation & 2020 & Encoder \& Decoder & No & NA & WSJ0-2mix, LS-2mix \\
\hline
Sepformer \citep{Sepformer} & Speech Separation & 2021 & Encoder \& Decoder & No & NA & WSJ0-2mix, WSJ0-3mix \\
\hline
WavLM \citep{WavLM} & Speech separation, speech denoising, speech prediction, Speaker Verification, Speech recognition & 2022 & Encoder & Yes & GigaSpeech, VoxPopuli & VoxCeleb1, VoxCeleb2, Switchboard-2 CALLHOME, LIBRICSS, LibriSpeech \\
 \hline
\caption{\label{tab:speech separation}Transformer models for audio \& speech separation task}
\end{longtable}

\begin{itemize}[leftmargin=.1in]

\item \textbf{Sepformer: }
The sepformer model was published in a paper titled “Attention is all you need in speech separation” which uses an attention mechanism to separate speeches that are overlapped. This model does not contain any kind of recurrence scheme and it follows the self-attention mechanisms. Sepformer uses a binary mask prediction scheme for training while this masking network captures both short and long-term dependencies and provide higher accuracy in performance  \citep{Sepformer}.

\item \textbf{DPTNeT: }
DPTNeT stands for Dual-Path Transformer Network for monaural speech separation tasks. This model trained directly minimizes the error between estimated and target value which is called end-to-end processing. This model uses dual-path architecture, replacing positional encoding with the RNN in the transformer architecture which helps to capture complex features of the signal and improves the performance of the speech separation from the overlapped speech  \citep{DPTNeT}.

\item \textbf{WavLM:}
WavLM is a large-scale pre-trained model that can execute a range of tasks for speech. WavLM follows BERT inspired speech processing model-HuBERT, whereas, with the help of mask speech prediction, the model predicts the actual speech by removing the noise from the corrupted speech. By this way, this model is trained for a variety of tasks besides automatic speech recognition-ASR task \citep{WavLM}.
\end{itemize}

\subsubsection{Speech Classification}

The speech classification task refers to the ability to categorize input speech or audio into distinct categories based on various features, including speaker, words, phrases, language, and more. There exist several speech classifiers, such as voice activity detection (binary/multi-class), speech detection (multi-class), language identification, speech enhancement, and speaker identification. Speech classification plays a crucial role in identifying important speech signals, enabling the extraction of relevant information from large speech datasets \citep{speech-classification,speech-classification1}. In this context, we present a compilation of transformer-based models, which have demonstrated superior accuracy in speech classification tasks compared to conventional models. The details of these models are depicted in Table \ref{tab:speech classification}.

\medskip
\setlength\LTleft{-.85cm}
\setlength\LTright{\LTleft}
\begin{longtable}{| M{2.4cm} | M{2.5cm}| M{.7cm} |M{1.8cm} |M{1.6cm} |M{3cm} |M{4cm} |}

 \hline
\textbf{Transformer Models} & \textbf{Task Accomplished} & \textbf{Year} & \textbf{Architecture (Encoder/ Decoder)} & \textbf{Pre-trained (Yes/NO)} & \textbf{Pre-training Dataset} & \textbf{Dataset (Fine-tuning, Training, Testing)} \\
 \hline
 \endfirsthead
 
 \multicolumn{7}{c}%
{{\bfseries \tablename\ \thetable{} -- continued from previous page}} \\
 \hline
\textbf{Transformer Models} & \textbf{Task Accomplished} & \textbf{Year} & \textbf{Architecture (Encoder/ Decoder)} & \textbf{Pre-trained (Yes/NO)} & \textbf{Pre-training Dataset} & \textbf{Dataset(Fine-tuning, Training, Testing)} \\
 \hline
\endhead

\hline \multicolumn{7}{|r|}{{Continued on next page}} \\ \hline
\endfoot

\endlastfoot

AST \citep{AST} & Audio Classification & 2021 & Encoder & Yes & ImageNeT & AudioSet, ESC-50, Speech Commands \\
\hline
Mockingjay \citep{Mockingjay} & Speech Classification \& recognition & 2020 & Encoder & Yes & LibriSpeech & LibriSpeech test clean, MOSEI \\
\hline
XLS-R \citep{XLS-R} & Speech Classification, Speech Translation, Speech Recognition & 2021 & Encoder \& Decoder & Yes & VoxPopuli (VP-400K), Multilingual Librispeech (MLS), CommonVoice, VoxLingua107, BABEL & VoxPopuli, Multilingual Librispeech (MLS), CommonVoice, BABEL \\
\hline
UniSpeech-SAT \citep{Unispeech-Sat} & Speech Classification \& Recognition & 2022 & Encoder & Yes & LibriVox, Librispeech GigaSpeech, VoxPopuli & SUPERB \\
 \hline
\caption{\label{tab:speech classification}Transformer models for audio \& speech classification task}
\end{longtable}

\begin{itemize}[leftmargin=.1in]

\item \textbf{AST: } 
AST - Audio Spectrogram Transformer is a transformer-based model which is applied to an audio spectrogram. AST is the first audio classification model where the convolution was not used and it is capable of capturing long-range frames context. It used a transformer encoder to capture the features in the audio spectrogram, a linear projection layer and sigmoid activation function to capture the audio spectrogram representation for audio classification. As the attention mechanism is renowned for capturing global features so it shows significant performance in audio/speech classification tasks \citep{AST}.

\item \textbf{Mockingjay: }
Mockingjay is an unsupervised speech representation model that uses multiple layers of bidirectional Transformer pre-trained encoders. It uses both past and future features for speech representation rather than only past information, which helps it to gather more information about the speech context. Mockingjay also can improve the performance of the supervised learning tasks as well where the amount of labeled data is low. Capturing more information helped to improve several speech representational tasks like speech classification and recognition \citep{Mockingjay}.

\item \textbf{XLS-R: }
XLS-R is a transformer-based self-supervised large-scale speech representation model which is trained with a large amount of data. It is built on the wav2vec discretization algorithm, whereas it uses Wav2Vec 2.0 model that is pretrained with multiple languages. The architecture contains multiple convolution encoders to map raw speech and the output from this stage is transferred to the transformer model(encoder module) as input which provides better audio representation finally. A large amount of training data is crucial for this model where a range of public speech is used and it performed well for multiple downstream multilingual speech tasks  \citep{XLS-R}.

\item \textbf{UniSpeech:}
UniSpeech is a semi-supervised unified pre-trained model for speech representation. This model follows the Wav2Vec 2.0 architecture where it contains convolutional feature encoders that converts the raw audio to a higher-dimensional representation and this output is fed into the Transformer. This model is capable of learning to multitask while a quantizer is used in its architecture which helps to capture specific speech recognition information \citep{Unispeech-Sat}.
\end{itemize}

\subsection{Signal Processing}

With the growing recognition of the usability of Transformer-based models across various sectors, researchers have started exploring their application in signal processing. This recent development of utilizing Transformer-based models in signal processing represents a novel approach that outperforms conventional methods in terms of performance.
Signal processing involves the manipulation and analysis of various types of data, including signal status, information, frequency, amplitude, and more. While audio and speech are considered forms of signals, we have segregated the audio and speech sections to highlight the specific applications of Transformer-based models in those domains. Within the signal processing domain, we have focused on two distinct areas: wireless network signal processing and medical signal processing. These two fields exhibit distinct processing methods and functionalities due to their inherent differences. 
Here, we delve into both of these tasks and provide an overview of significant Transformer-based models that have demonstrated effectiveness in these specific domains.

\subsubsection{Wireless Network \& Signal Processing}


In the current era of the 21st century, wireless network communication has emerged as a prominent technology. However, the application of transformers in wireless network signal processing has not received substantial attention thus far. Consequently, the number of Transformer-inspired models developed for this field remains limited. 
Wireless network signal processing encompasses various tasks, including signal denoising, signal interface detection, wireless signal channel estimation, interface identification, signal classification, and more. Deep neural networks offer great potential for tackling these tasks effectively, and Transformer-based models have introduced significant advancements in this domain  \citep{wireless1,wireless2,wireless3,wireless4}. In this section, we present several models that have made notable contributions to the enhancements in wireless communication networks and signal processing. The details of these models are provided in Table \ref{tab:wireless network}.


\medskip
\setlength\LTleft{-.86cm}
\setlength\LTright{\LTleft}
\begin{longtable}{| M{2cm} | M{3cm}| M{.7cm} |M{1.8cm} |M{1.6cm} |M{3cm} |M{4cm} |}

 \hline
\textbf{Transformer Models} & \textbf{Task Accomplished} & \textbf{Year} & \textbf{Architecture (Encoder/ Decoder)} & \textbf{Pre-trained (Yes/NO)} & \textbf{Pre-training Dataset} & \textbf{Dataset (Fine-tuning, Training, Testing)} \\
 \hline
 \endfirsthead
 
 \multicolumn{7}{c}%
{{\bfseries \tablename\ \thetable{} -- continued from previous page}} \\
 \hline
\textbf{Transformer Models} & \textbf{Task Accomplished} & \textbf{Year} & \textbf{Architecture (Encoder/ Decoder)} & \textbf{Pre-trained (Yes/NO)} & \textbf{Pre-training Dataset} & \textbf{Dataset (Fine-tuning, Training, Testing)} \\
 \hline
\endhead

\hline \multicolumn{7}{|r|}{{Continued on next page}} \\ \hline
\endfoot

\endlastfoot

SigT \citep{SigT} & Signal detection, channel estimation, interference suppression, and data decoding in MIMO-OFDM & 2022 & Encoder & No & NA & Peng Cheng laboratory(PCL), local area data \\
\hline
TSDN \citep{TSDN} & Remove Interference and nose from wireless signal & 2022 & Encoder \& Decoder & No & NA & Wall NLoS, Foil NLOS, UWB dataset \\
 \hline
ACNNT \citep{ACNNT} & Wireless interface identification & 2021 & Encoder & No & NA & ST, BPSK, AM, NAM, SFM, LFM, 4FSK, 2FSK signal dataset \\
 \hline
MCformer \citep{MCformer} & Automatic modulation classification complex raw radio signals & 2021 & Encoder & No & NA & RadioML2016.10b \\
 \hline
Quan-Transformer \citep{Quan-Transformer} & Compress \& recover channel state information & 2022 & Encoder \& Decoder & No & NA & CsiNet, CLNet and CRNet \\
 \hline
\caption{\label{tab:wireless network}Transformer models for wireless network \& signal processing}
\end{longtable}

\begin{itemize}[leftmargin=.1in]
\item \textbf{SigT:}
SigT is a wireless communication network signal receiver designed with a transformer architecture, capable of handling Multiple-input Multiple-output (MIMO-OFDM) signals. Leveraging the transformer's encoder module, this innovative framework enables parallel data processing and performs essential tasks such as signal detection, channel estimation, and data decoding. Unlike traditional receivers that rely on distinct modules for each task, SigT seamlessly integrates these functions, providing a unified solution \citep{SigT}.

\item \textbf{TSDN:}
TSDN is an abbreviation for Transformer-based Signal Denoising Network. It refers to a signal denoising model based on transformers that aims to estimate the Angle-of-Arrival (AoA) of signals transmitted by users within a wireless communication network. This transformer-based model significantly enhances the accuracy of AoA estimation, especially in challenging non-line-of-sight (NLoS) environments where conventional methods often fall short in delivering the desired precision \citep{TSDN}.

\item \textbf{ACNNT:}
The Augmented Convolution Neural Network with Transformer (ACNNT) is an architectural framework specifically designed for identifying interference within wireless networks. This model combines the power of Convolutional Neural Networks (CNN) and transformer architectures. The multiple CNN layers in ACNNT extract localized features from the input signal, while the transformer component captures global relationships between various elements of the input sequence. By exploiting the strengths of both CNN and Transformer, this model has demonstrated superior accuracy in the identification of wireless interference compared to conventional approaches \citep{ACNNT}.

\item \textbf{MCformer:}
MCformer, short for Modulation Classification Transformer, refers to a model architecture based on transformers that performs feature extraction from input signals and subsequently classifies them based on modulation. This architectural design combines Convolutional Neural Network (CNN) and self-attention layers, enabling the processing of intricate features within the signal and achieving superior accuracy in comparison to conventional approaches. The introduction of this model has brought about noteworthy advancements in a wireless network and communication signals, particularly in the realm of Automatic Modulation Classification, thereby enhancing system security and performance \citep{MCformer}.

\item \textbf{Quan-Transformer:}
It refers to a Transformer-based model specifically designed to perform quantization in wireless communication systems. Quantization is the vital process of converting a continuous signal into a discrete signal, and it plays a crucial role in network channel feedback processing. Channel feedback processing is essential for estimating channel state information, which in turn aids in adjusting signal transmission parameters. This feedback mechanism holds particular significance in the context of Reconfigurable Intelligent Surface (RIS)-aided wireless networks, a critical component of the 6th-generation communication system \citep{Quan-Transformer}.

\end{itemize}
\subsubsection{Medical Signal processing}
The rise of healthcare data has resulted in the rapid growth of deep learning applications, enabling the automatic detection of pathologies, enhanced medical diagnosis, and improved healthcare services. These data can be categorized into three distinct forms: relational data ( symptoms, examinations, and laboratory tests), medical images, and biomedical signals (consisting of raw electronic and sound signals). While the application of deep learning models, particularly transformers, in the context of medical images has gained considerable attention and yielded promising results, the application of transformers to biomedical signals is still in its early stages. The majority of relevant studies have been published between the years 2021 and 2022, with a particular focus on the task of signal classification. We have summarized our findings regarding the application of transformers to biomedical signals in Table \ref{tab:signal}.

\medskip
\begin{itemize} 

\item \textbf{Epilepsy disease case: }  

Epilepsy is a serious debilitating condition for those it affects. Typically, its symptoms are detected through the use of electrical signals, such as electroencephalograms (EEGs) and magnetoencephalograms (MEGs). With the rise of deep learning, it is now possible to detect and predict epilepsy cases using its architecture. Utilizing transformers as their underlying framework, the following models have been developed to analyze electric signals for  predicting and classifying epilepsy.

\begin{itemize}

\medskip
\item \textbf{Three-tower transformer network \citep{yan2022seizure}: } The purpose of this model is to predict epileptic seizures from EEG signals. A transformer-based model is used to perform binary classification of EEG signals based on three EEG features: time, frequency, and channel. This model processes EEG signals as a whole using a model that is based on the classic transformer, which contains three encoders: a time encoder, a frequency encoder, and a channel encoder. Compared to other models, such as CNN, the model shows better performance results in predicting epilepsy attacks.

\medskip
\item \textbf{TransHFO \citep{guo2022transformer}:} (Transformer-based HFO) is a deep learning model based on BERT for classifying High-Frequency Oscillation (HFO) from normal control (NC). A transformer is used to detect the presence of HFO with high accuracy in one-dimensional magnetoencephalography (MEG) data in order to identify epileptic areas more precisely. Signal classification is performed under k-fold cross-validation and through various algorithms, including logistic regression, SMO, and the ResDen model. Due to the small dataset available, the authors propose to use the data augmentation technique ``ADASYN". Nevertheless, even with the addition of data, the dataset remains small, which makes the shallow transformer more efficient than the deep transformer with more layers.
\end{itemize}
\end{itemize}
\medskip

\begin{table}[H]
\footnotesize
 \begin{adjustbox}{width=1.1\textwidth,center=\textwidth}
\begin{tabular} {| M{3cm} | M{2cm}| M{1cm} |M{2.5cm} |M{1cm} |M{2cm} |M{3cm} |}
 \hline
\textbf{Transformer Name} & \textbf{Field of application} & \textbf{Year} & \textbf{Fully Transformer Architecture} & \textbf{Signal type} & \textbf{Transformer Task} & \textbf{Dataset} \\\hline
Three-tower transformer network \citep{yan2022seizure} &
Epilepsy &
2022 &
YES &
EEG &
Classification of EEG signals &
CHB-MIT datasets\\
 \hline
TransHFO \citep{guo2022transformer} &
Epilepsy &
2022 &
YES &
MEG &
Classification of MEG signals &
20 clinical patients\\
 \hline
TCN and Transformer-based model \citep{DBLP:journals/jocs/CasalPS22} &
Sleep pathologies &
2022 &
No (using TCN which is based on CNN) &
Cardiac signals ( Heart Rate) &
Classification of sleep stages &
Sleep Heart Health Study dataset\\
 \hline
Constrained transformer network \citep{DBLP:journals/midm/CheZZQJ21} &
Heart disease &
2021 &
No (Using CNN) & 
ECG &
Classification of ECG signals & 
6877 patients\\
 \hline
CRT-NET \citep{DBLP:journals/asc/LiuLFHYLXZW22} &
Heart Disease &
2022 &
No (Using CNN and Bi-directional GRU) &
ECG &
Classification and recognition of ECG signals & 
MIT-BIH, CPSC arrhythmia clinical private data\\
 \hline
CAT \citep{DBLP:journals/cbm/YangLP22} &
Atrial Fibrillation &
2022 &
No (Using MLP) &
ECG &
Classification of ECG signals &
Shaoxing database (more than 10000 patients)\\
 \hline
\end{tabular}
\end{adjustbox}
\caption{\label{tab:signal}Transformer models for medical signal processing}
\end{table}

\medskip
\begin{itemize}

\item \textbf{Cardiac diseases cases:} heart diseases are among the areas in which researchers are interested in applying transformers. Using ECG signals, a transformer model can detect long-range dependencies and identify heart disease types based on their characteristics.
\begin{itemize}
    
\medskip
\item \textbf{Constrained transformer network  \citep{DBLP:journals/midm/CheZZQJ21}:} Using CNN and transformer architecture, this model classifies heart arrhythmia disease based on temporal information in ECG (electrocardiogram) signals. There are also other models that use transformer encoders for classifying heart diseases (such as atrial fibrillation) using ECG signals, such as CRT-NET \citep{DBLP:journals/asc/LiuLFHYLXZW22} and CAT \citep{DBLP:journals/cbm/YangLP22} “Component-Aware Transformer”.

A major advantage of CAT's model \citep{DBLP:journals/cbm/YangLP22} is the use of a large database containing data from over ten thousand patients for experiments. In contrast, the strength of CRT-NET \citep{DBLP:journals/asc/LiuLFHYLXZW22} is the ability to extract different ECG features like waveforms, morphological characteristics, and time domain data, in order to identify many cardiovascular diseases.

\medskip
\item \textbf{TCN and Transformer-based model \citep{DBLP:journals/jocs/CasalPS22}:} An automatic sleep stage classification system based on 1-dimensional cardiac signals (Heart Rate). The classification is conducted in two steps: extracting features from signals using TCN "Temporal Convolution Network"[], and modeling signal sequence dependencies using the standard Transformer architecture consisting of two stacks of encoders and a simplified decoder module.
Based on a dataset of 5000 different participants, this study demonstrated that this new model outperforms other networks, such as CNN and RNN, which consume more memory and reduce process efficiency.
\end{itemize}
\end{itemize}

\section{Future Prospects and Challenges} \label{future_work}
One of the primary objectives of this survey is to identify and highlight potential research directions for transformer applications, with the goal of expanding their range of applications beyond the currently popular fields of NLP and computer vision. Despite considerable research attention in these areas, there are still a number of areas that remain relatively unexplored with the potential for significant improvements in the future.
In order to expand the application areas of Transformers, we have identified several potential directions for future research. These directions include but are not limited to the exploration of transformer-based models for speech recognition, recommendation systems, and natural language generation. In addition, further exploration of transformer-based approaches for multimodal tasks, such as combining audio and visual inputs, would be an interesting direction for future research. By pursuing these research directions, we hope to continue the advancement of transformer-based models and their utility in a broader range of applications.

\subsection{Transformers in Wireless Network and Cloud Computing}
While the majority of transformer applications have been in NLP and Computer Vision, there is an exciting potential for transformers in the wireless communication and cloud computing domains. Although there have been relatively fewer studies in this area, the ones that exist have demonstrated the enormous potential of transformers for improving various aspects of wireless signal communication and cloud workload computing.
In this section, we discuss some of the transformer models that have been developed for wireless signal communication and cloud computing. These models have shown promising results in areas such as wireless interference recognition, wireless signal communication mitigation, and cloud workload forecasting.
Moving forward, there are several potential directions for future research in both the wireless network and cloud domains. Some of the possible areas of focus for wireless communication include improving network security, enhancing the efficiency of wireless communication, and developing more accurate interference recognition models. On the other hand, for cloud computing, future work could focus on improving resource allocation and workload management, optimizing cloud performance, and enhancing data privacy and security.\\
\par
\textbf {Scope of future work for the wireless Signal communication:}
\begin{itemize}
  \item Detection of Wireless Interference: The global attention capabilities of Transformers present an exciting avenue for future research in the field of wireless signal communication. By leveraging the power of Transformers, researchers can explore and develop more widely used applications for detecting wireless interference in communication systems. This can involve experimenting with various successive Transformer models to reduce complexity and enhance the efficiency of wireless interference recognition. This research can lead to improved communication systems that are more resilient to interference and provide better overall performance.
  \item Enhancing 5G \& 6G Networks: As 5G and 6G networks gain popularity, there is significant potential for Transformers to contribute to this field. Advanced networking architectures, such as Reconfigurable Intelligent Surfaces (RIS) and Multiple-Output and Orthogonal Frequency-Division Multiplexing (MIMO-OFDM), play a crucial role in these networks. Transformers have shown promise in improving performance in these areas. Additionally, signal state feedback, which is essential for adjusting and updating networks based on signal state changes, can benefit from the parallel computational capability of Transformers. The ability of Transformers to handle multiple tasks simultaneously, including signal detection, channel estimation, and data decoding, makes them an effective alternative to conventional methods.
  \item  Integration of Transformers with Advanced Communication Technologies: Transformers can be integrated with other advanced communication technologies to further improve wireless signal communication. For example, combining Transformers with technologies like Massive MIMO, millimeter-wave communication, and cognitive radio can enhance the performance, capacity, and spectrum efficiency of wireless networks. Future research can focus on exploring these synergies and developing innovative solutions that leverage the unique capabilities of Transformers in conjunction with other cutting-edge communication technologies.\\
\end{itemize}
\textbf {Future possibilities for the Cloud:}
\begin{itemize}
  \item Advancements in Cloud Computing: With the increasing application of the Internet of Things (IoT), the cloud plays a crucial role in supporting and managing IoT devices. Transformers offer exciting possibilities for advancing cloud capabilities in various tasks, such as early attack and anomaly detection. By leveraging different Transformer approaches, the cloud can learn and adapt to its behavior, bringing more stability and security. Additionally, Transformers can be applied to cloud computing tasks like task scheduling and memory allocation. The multi-head attention and long-range attention features of the Transformers model make it well-suited for optimizing resource allocation and improving overall performance in cloud environments.
  \item Transformation in Mobile Edge Computing (MEC) and Mobile Edge Caching (MEC): In the context of advanced 6G networking systems, Mobile Edge Computing (MEC) and Mobile Edge Caching (MEC) play vital roles in reducing communication latency. Transformers have demonstrated significant potential in enhancing MEC and MEC through their parallel computational capabilities. Transformers can be applied to predict popular content, improve content management, optimize resource allocation, and enhance data transmission in MEC systems. By leveraging Transformers, the mobile cloud can respond and process user requests faster, resulting in reduced network response times and faster data transmission.
  \item Intelligent Resource Management in the Cloud: Transformers offer opportunities for intelligent resource management in cloud environments. By applying Transformers to tasks like workload prediction, resource allocation, and load balancing, cloud systems can optimize resource utilization and enhance performance. Transformers' ability to capture long-range dependencies and handle complex patterns makes them well-suited for efficiently managing cloud resources and improving overall system efficiency.
  \item Security and Privacy in the Cloud: Transformers can contribute to enhancing security and privacy in the cloud by enabling advanced threat detection, anomaly detection, and data privacy protection mechanisms. Transformers can analyze large volumes of data, identify patterns, and detect potential security breaches or anomalies in real-time. Additionally, Transformers can be utilized for data anonymization and privacy-preserving computations, ensuring that sensitive information remains protected in cloud-based systems.
\end{itemize}

\subsection{Medical Image \& Signal Processing}

Two types of medical data are discussed in this paper: images and signals. According to our literature review, segmentation, and classification are the most transformer-based medical applications, followed by image translation \citep{yan2022swin, chen2022more}.
In the context of medical images, we commonly see the reuse of existing transformers, such as BERT, ViT, and SWIN, regardless of how the original model was modified. Further, we observe that various types of medical images are used to conduct transformer-based medical applications, such as 2D images \citep{DBLP:journals/mia/HeTBZZL22, gu2022chest}, 3D images \citep{jiang2022swinbts, liang2022transconver,  DBLP:journals/corr/abs-2109-03201, DBLP:conf/midl/ZhuHWYLOX22} and multi-mode images \citep{sun2021hybridctrm}.

The selected research papers for this survey span the years 2021 to 2022, indicating that the use of transformer architecture in the medical field is still in its nascent stages. Despite its early adoption, there has been a remarkable influx of excellent publications exploring transformer applications in the analysis of medical images within this relatively short period. However, several challenges persist in applying transformers to medical images that need to be addressed and overcome:

\begin{itemize}
  \item \textbf{Limited focus on 3D images}: There is a scarcity of studies that specifically address the application of transformers to 3D medical images. Most research has been concentrated on 2D images, indicating the need for further exploration and development in this area.

  \item \textbf{Small and private medical image databases}: Medical image databases are often small and privately owned due to legal and ethical concerns regarding patient data privacy \citep{lopez2020medical}. This limits the availability of large-scale datasets necessary for training transformer models effectively.
  \item \textbf{Computational complexity in high-resolution imaging}: Transformer-based architectures encounter computational challenges when dealing with high-resolution medical images. The self-attention mechanism, which is integral to transformers, becomes computationally demanding for large images. However, some models, like DI-UNET \citep{DBLP:journals/bspc/WuWWWL22}, have introduced enhanced self-attention mechanisms to handle higher resolution images effectively.
    \item \textbf{Limited number of fully developed transformer-based models}: The development of transformer-based models for processing medical images is still relatively nascent. Due to the computational complexity and parameter requirements of transformers, existing architectures often combine deep learning techniques like CNNs and GANs with transformers \citep{DBLP:journals/kbs/MaXSL22}. Knowledge distillation techniques may offer a viable solution for training transformer models with limited computational and storage resources \citep{DBLP:journals/bspc/LengLGD22}.
\end{itemize}
  
Moreover, the application of transformers to bio-signals is relatively limited compared to medical images. There are two main challenges that transformers face in the domain of biomedical signals:

\begin{itemize}
\item Small bio-signal databases: Bio-signal databases often have limited sizes, which poses challenges for training and validating transformer models effectively. For instance, in a study mentioned by \citep{guo2022transformer}, only 20 patients were included, which is considered insufficient to establish the effectiveness of a model. To mitigate the limitations of small databases, some studies have proposed the use of virtual sample generation techniques like ADASYN \citep{DBLP:conf/ijcnn/HeBGL08} to augment the dataset.

\item Limited availability of transformer-based models: Currently, there is a scarcity of models that are exclusively based on transformers for processing biomedical signals. The application of transformers in this context is still relatively unexplored, and more research is needed to develop dedicated transformer architectures for bio-signal analysis and processing.
\end{itemize}

\subsection{Reinforcement Learning}
The integration of transformers with deep reinforcement learning (RL) methods has emerged as a promising approach for enhancing sequential decision-making processes. Within this domain, two main research categories can be identified: architecture enhancement and trajectory optimization.

In the "architecture enhancement" category, transformers are applied to RL problems based on traditional RL paradigms. This involves leveraging the capabilities of transformers to improve the representation and processing of RL states and actions. On the other hand, the "trajectory optimization" approach treats RL problems as sequence modeling tasks. It involves training a joint state-action model over entire trajectories, utilizing transformers to learn policies from static datasets, and leveraging the transformers' ability to model long sequences.

Deep RL heavily relies on interactions with the environment to collect data dynamically \cite{rjoub2019deep,rjoub2021deep}. However, in certain scenarios such as expensive environments like robotic applications or autonomous vehicles, collecting sufficient training data through real-time interaction may be challenging. To address this, offline RL techniques have been developed, which leverage deep networks to learn optimal policies from static datasets without direct environment interaction. In deep RL settings, transformers are often used to replace traditional components like convolutional neural networks (CNNs) or long short-term memory (LSTM) networks \cite{rjoub2022trust}, providing memory-awareness and improved modeling capabilities to the agent network. However, standard transformer structures applied directly to decision-making tasks may suffer from stability issues. To overcome this limitation, researchers have proposed modified transformer architectures, such as GtrX \citep{stab_transformer_rl}, as an alternative solution.

In summary, approaches like Decision Transformer and Trajectory Transformer have addressed RL problems as sequence modeling tasks, harnessing the power of transformer architectures to model sequential trajectories \citep{dec_transformer, traj_transformer}. While these methods show promise in RL tasks, there is still significant room for improvement. Treating RL as sequence modeling simplifies certain limitations of traditional RL algorithms but may also overlook their advantages. Therefore, an interesting direction for further exploration is the integration of traditional RL algorithms with sequence modeling using transformers, combining the strengths of both approaches.

\subsection{Other Prospects}
The successful application of transformers in the field of NLP has sparked interest and exploration in various other domains. Researchers have been inspired to apply transformer models to diverse areas, leading to promising developments. For instance, the transformer model BERT has been utilized to model proteins, which, similar to natural language, can be considered as sequential data \citep{DBLP:conf/iclr/VigMVXSR21}. Additionally, the transformer model GPT-2 has been employed to automatically fix JavaScript software bugs and generate patches without human intervention \citep{DBLP:conf/icse-apr/LajkoCV22}.

Beyond its impact in traditional machine learning and deep learning domains, transformers have found applications in industrial studies as well. They have demonstrated impressive performance in various tasks, ranging from predicting the state-of-charge of lithium batteries \citep{shen2022state} to classifying vibration signals in mechanical structures \citep{DBLP:journals/eswa/JinC21}. Notably, transformers have showcased superior capabilities compared to Graph Neural Networks (GNNs) in constructing meta-paths from different types of edges in heterogeneous graphs \citep{DBLP:conf/nips/YunJKKK19}. This highlights the potential of transformers in handling complex and diverse data structures.

Another intriguing future application of transformers lies in the field of "Generative Art," where intelligent systems are leveraged for automated artistic creation, including images, music, and poetry. While image generation is a well-explored application area for transformers, often focused on natural or medical images, the domain of artistic image generation remains relatively unexplored. However, there have been some initial models based on transformers, such as "AffectGAN," which generates images based on semantic text and emotional expressions using transformer models \citep{DBLP:conf/acii/GalanosLY21}. The exploration of transformers in generative art has significant untapped potential for further advancements and creative outputs.

Overall, the application of transformers extends beyond NLP and showcases immense potential in various domains, ranging from scientific research to industrial applications and artistic creativity. Continued exploration and innovation in these areas will further expand the possibilities and impact of transformers in the future.

\section{Conclusion} \label{conclusion}
The transformer, as a deep neural network, has demonstrated superior performance compared to traditional recurrence-based models in processing sequential data. Its ability to capture long-term dependencies and leverage parallel computation has made it a dominant force in various fields such as NLP, computer vision, and more. In this survey, we conducted a comprehensive overview of transformer models' applications in different deep learning tasks and proposed a new taxonomy based on the top five fields and respective tasks: NLP, Computer Vision, Multi-Modality, Audio \& Speech, and Signal Processing.
By examining the advancements in each field, we provided insights into the current research focus and progress of transformer models. This survey serves as a valuable reference for researchers seeking a deeper understanding of transformer applications and aims to inspire further exploration of transformers across various tasks. Additionally, we plan to extend our investigation to emerging fields like wireless networks, cloud computing, reinforcement learning, and others, to uncover new possibilities for transformer utilization.
The rapid expansion of transformer applications in diverse domains showcases its versatility and potential for continued growth. With ongoing advancements and novel use cases, transformers are poised to shape the future of deep learning and contribute to advancements in fields beyond the traditional realms of NLP and computer vision.

\bibliographystyle{model5-names}
{\footnotesize
\bibliography{ref.bib}}

\end{document}